\documentclass[10pt,twocolumn,letterpaper]{article}

\usepackage{cvpr}              

\usepackage{graphicx}
\usepackage{amsmath}
\usepackage{amssymb}
\usepackage{booktabs}
\usepackage{multirow}
\usepackage{colortbl}
\usepackage{tabularray}
\usepackage{float}
\usepackage{bm}
\definecolor{yzybest}{rgb}{0.96, 0.57, 0.58}
\definecolor{yzysecond}{rgb}{0.98, 0.78, 0.57}
\definecolor{yzythird}{rgb}{1.0, 1.0, 0.56}

\definecolor{gscolor}{rgb}{1.0,0.6,0.0} %

\usepackage[pagebackref,breaklinks,colorlinks]{hyperref}

\usepackage[capitalize]{cleveref}
\crefname{section}{Sec.}{Secs.}
\Crefname{section}{Section}{Sections}
\Crefname{table}{Table}{Tables}
\crefname{table}{Tab.}{Tabs.}

\def\paperID{3223} 
\def\confName{CVPR}
\def\confYear{2024}

\begin{document}

\title{Deformable 3D Gaussians for High-Fidelity Monocular Dynamic Scene Reconstruction}



\author{
    Ziyi Yang$^{1,2}$ 
    \quad Xinyu Gao$^{1}$ 
    \quad Wen Zhou$^{2}$ 
    \quad Shaohui Jiao$^{2}$ 
    \quad Yuqing Zhang$^{1}$
    \quad Xiaogang Jin$^{1\dagger}$ 
    \vspace{1em}
    \\
    $^1$State Key Laboratory of CAD\&CG, Zhejiang University \quad 
    $^2$ByteDance Inc. \quad
}


\twocolumn[{%
\renewcommand\twocolumn[1][]{#1}%
\maketitle
\begin{center}
    \centering
    \captionsetup{type=figure}
    \includegraphics[width=1.0\textwidth]{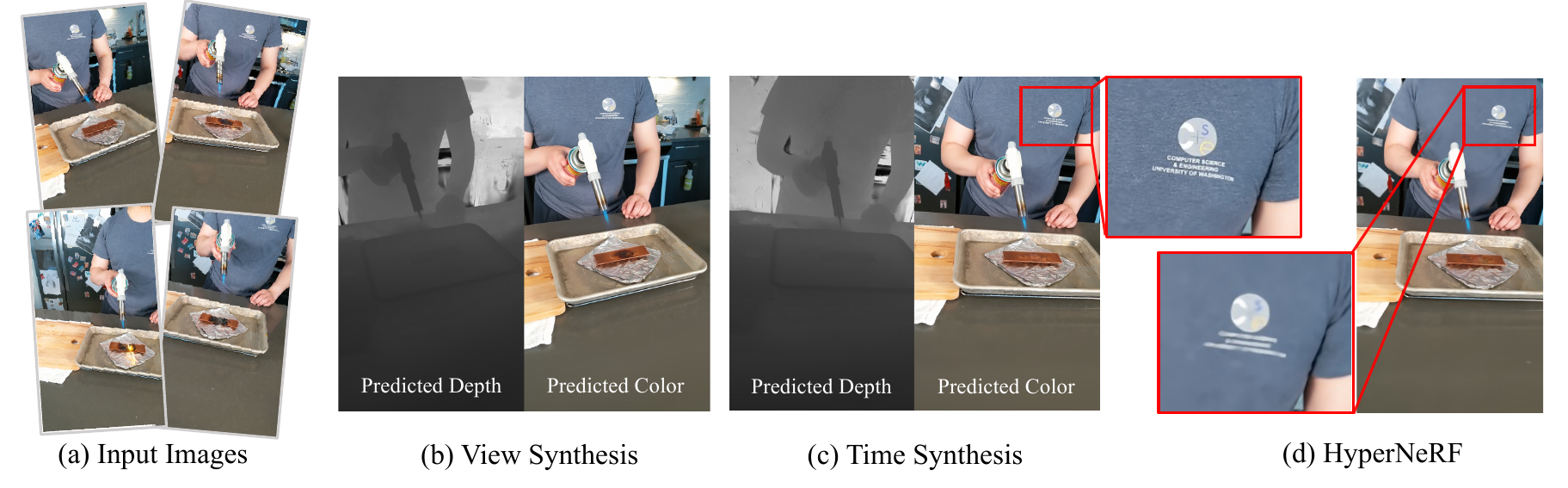}
    \captionof{figure}{Given a set of monocular multi-view images and camera poses (a), our proposed method can reconstruct accurate dynamic scene geometry and render high-quality images in both the novel-view synthesis (b) and time interpolation (c) tasks. In real-world datasets with intricate details, our method outperforms \emph{HyperNeRF} \cite{park2021hypernerf} (d) in terms of rendering quality and time performance.}
\label{fig:teaser}
\end{center}%
}]

\begin{abstract}
Implicit neural representation has paved the way for new approaches to dynamic scene reconstruction and rendering. Nonetheless, cutting-edge dynamic neural rendering methods rely heavily on these implicit representations, which frequently struggle to capture the intricate details of objects in the scene. Furthermore, implicit methods have difficulty achieving real-time rendering in general dynamic scenes, limiting their use in a variety of tasks.  
To address the issues, we propose a deformable 3D Gaussians Splatting method that reconstructs scenes using 3D Gaussians and learns them in canonical space with a deformation field to model monocular dynamic scenes. We also introduce an annealing smoothing training mechanism with no extra overhead, which can mitigate the impact of inaccurate poses on the smoothness of time interpolation tasks in real-world datasets. 
Through a differential Gaussian rasterizer, the deformable 3D Gaussians not only achieve higher rendering quality but also real-time rendering speed. Experiments show that our method outperforms existing methods significantly in terms of both rendering quality and speed, making it well-suited for tasks such as novel-view synthesis, time interpolation, and real-time rendering. 
Our code is available at \href{https://github.com/ingra14m/Deformable-3D-Gaussians}{https://github.com/ingra14m/Deformable-3D-Gaussians}.
\end{abstract}
\section{Introduction}
\label{sec:intro}

\par 
High-quality reconstruction and photorealistic rendering of \textit{dynamic scenes} from a set of input images is critical for a variety of applications, including augmented reality/virtual reality (AR/VR), 3D content production, and entertainment. 
Previously used methods for modeling these dynamic scenes relied heavily on mesh-based representations, as demonstrated by methods described in \cite{collet2015high, kanade1997virtualized, li2012temporally, starck2007surface}.   However, these strategies frequently face inherent limitations, such as a lack of detail and realism, a lack of semantic information, and difficulties in accommodating topological changes. With the introduction of neural rendering techniques, this paradigm has undergone a significant shift. Implicit scene representations, particularly as implemented by NeRF \cite{mildenhall2020nerf}, have demonstrated commendable efficacy in tasks such as novel-view synthesis, scene reconstruction, and light decomposition. 

\par 
To improve inference efficiency in NeRF-based static scenes, researchers have developed a variety of acceleration methods, including grid-based structures \cite{Chen2022ECCV, wang2023f2nerf} and pre-computation strategies \cite{yu2021plenoctrees, Wang2022fourier_plenOctrees}. Notably, by incorporating hash encoding, Instant-NGP \cite{mueller2022instant} has achieved rapid training. In terms of quality improvement, mipNeRF \cite{barron2021mipnerf} pioneered an effective anti-aliasing method, which was later incorporated into the grid-based approach by zipNeRF \cite{barron2023zipnerf}. 3D-GS \cite{kerbl3Dgaussians} recently extended the point-based rendering to efficient CUDA implementation with 3D Gaussians, which has enabled a real-time rendering while matches or even exceeds the quality of Mip-NeRF \cite{barron2021mipnerf}. However, this method is designed for representing static scenes, and its highly customized CUDA rasterization pipeline diminishes its scalability.

\par Implicit representations have been increasingly harnessed for modeling dynamic scenes. To handle the motion part in a dynamic scene, entangled methods \cite{wang2021neural, xian2021space} conditioned NeRF on a time variable. 
Conversely, disentangled methods \cite{pumarola2021d, park2021nerfies, park2021hypernerf, liu2022devrf, nerfplayer} employ a deformation field to model a scene in canonical space by mapping point coordinates at a given time to this space. 
This decoupled modeling approach can effectively represent scenes with non-dramatic action variations. 
However, irrespective of the categorization, adopting an implicit representation for dynamic scenes often proves both inefficient and ineffective, manifesting slow convergence rates coupled with a marked susceptibility to overfitting. Drawing inspiration from seminal NeRF acceleration research, numerous studies on dynamic scene modeling have integrated discrete structures, such as voxel-grids \cite{shao2023tensor4d, TiNeuVox}, or planes \cite{kplanes_2023, Cao2023HexPlane}. This integration amplifies both training speed and modeling accuracy. However, challenges remain. Techniques leveraging discrete structures still grapple with the dual constraints of achieving real-time rendering speeds and producing high-quality outputs with adequate detail. Multiple facets underpin these challenges: Firstly, ray-casting, as a rendering modality, frequently becomes inefficient, especially when scaled to higher resolutions. Secondly, grid-based methods rely on a low-rank assumption. Dynamic scenes, in comparison to static ones, exhibit a higher rank, which hampers the upper limit of quality achievable by such approaches.

\par In this paper, to address the aforementioned challenges, we extend the static 3D-GS and propose a deformable 3D Gaussian framework for modeling dynamic scenes. To enhance the applicability of the model, we specifically focus on the modeling of monocular dynamic scenes. Rather than reconstructing the scene frame by frame \cite{luiten2023dynamic}, we condition the 3D Gaussians on time and jointly train a purely implicit deformation field with the learnable 3D Gaussians in canonical space. The gradients for these two components are derived from a customized differential Gaussian rasterization pipeline. Furthermore, to solve the jitter in temporal sequences during the reconstruction process caused by inaccurate poses, we incorporate an annealing smoothing training (AST) mechanism. This strategy not only improves the smoothness between frames in the time interpolation task but also allows for greater detail to be rendered.

\par In summary, the major contributions of our work are:
\begin{itemize}
    \item A deformable 3D-GS framework for modeling monocular dynamic scenes that can achieve real-time rendering and high-fidelity scene reconstruction.
    \item A novel annealing smoothing training mechanism that ensures temporal smoothness while preserving dynamic details without increasing computational complexity.
    \item The first framework to extend 3D-GS for dynamic scenes through a deformation field, enabling the learning of 3D Gaussians in canonical space.
\end{itemize}


\begin{figure*}[ht] 
  \centering
  \centering
  \includegraphics[width=0.95\textwidth]{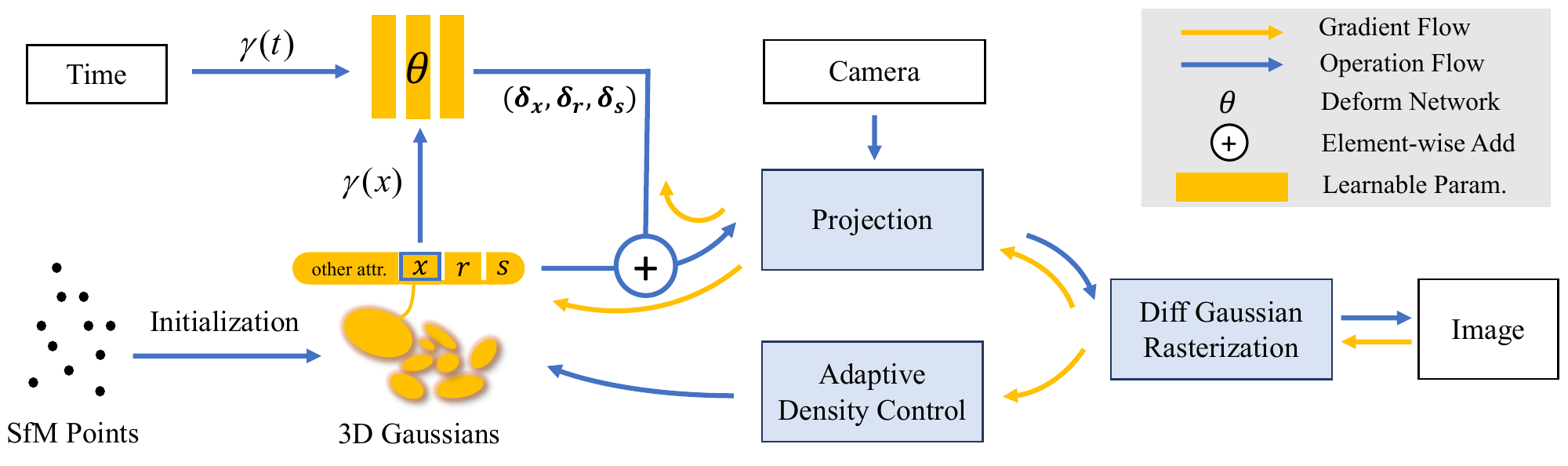}
  \caption{\textbf{Overview of our pipeline.} 
  The optimization process begins with Structure from Motion (SfM) points derived from COLMAP or generated randomly, which serve as the initial state for the 3D Gaussians. We use the position (detached) of 3D Gaussians $\gamma(\operatorname{sg}(\boldsymbol{x}))$ and time $\gamma(t)$ with positional encoding as input to a deformation MLP network to obtain the offset $(\delta \boldsymbol{x}, \delta \boldsymbol{r}, \delta \boldsymbol{s})$ of dynamic 3D Gaussians in canonical space. We use a warm-up phase for the 3D Gaussians during the first 3k iterations without optimizing the deformation field. Following that, we use the fast differential Gaussian rasterization pipeline to perform joint optimization of the deformation field and the 3D Gaussians, as well as to adaptively control the density of the set of Gaussians.
  } \label{fig:pipeline}
\end{figure*}
\section{Related Work}
\label{sec:related}

\subsection{Neural Rendering for Dynamic Scenes}
\par Neural rendering, due to its unparalleled capability to generate photorealistic images, has seen an uptick in scholarly interest. Recently, NeRF \cite{mildenhall2020nerf} facilitates photorealistic novel view synthesis through the use of MLPs. Subsequent research has expanded the utility of NeRF to various applications, encompassing tasks such as mesh reconstruction from a collection of images \cite{wang2021neus, li2023neuralangelo}, inverse rendering \cite{boss2021nerd, zhang2021nerfactor, liu2023nero}, optimization of camera parameters \cite{lin2021barf, wang2021nerfmm, wang2023badnerf}, and few-shot learning \cite{kangle2021dsnerf, Yang2023FreeNeRF}.

\par 
Constructing radiance fields for dynamic scenes is a critical branch in the advancement of NeRF, with significant implications for real-world applications. A cardinal challenge in rendering these dynamic scenes lies in the encoding and effective utilization of temporal information, especially when addressing the reconstruction of monocular dynamic scenes, a task inherently involves sparse reconstruction from a single viewpoint. One class of dynamic NeRF approaches models scene deformation by adding time $t$ as an additional input to the radiance field. 
However, this strategy couples the positional variations induced by temporal changes with the radiance field, lacking the geometric prior information regarding the influence of time on the scene. Consequently, substantial regularization is required to ensure temporal consistency in the rendering results. Another category of methods \cite{pumarola2021d, park2021nerfies, park2021hypernerf} introduces a deformation field to decouple time and the radiance field, mapping point coordinates to the canonical space corresponding to time $t$ through the deformation field. This decoupled approach is conducive to the learning of pronounced rigid motions and is versatile enough to cater to scenes undergoing topological shifts. Other methods seek to enhance the quality of dynamic neural rendering from various aspects, including segmenting static and dynamic objects in the scene \cite{nerfplayer, tretschk2020nonrigid}, incorporating depth information \cite{attal2021torf} to introduce geometric prior, introducing 2D CNN to encode scene priors \cite{lin2022efficient, peng2023representing}, and leveraging the redundant information in multi-view videos \cite{li2022neural} to set up keyframe compression storage, thereby accelerating the rendering speed.


\par However, the rendering quality of existing dynamic scene modeling based on MLP (Multilayer Perceptron) remains unsatisfactory. In this work, we will focus on the reconstruction of monocular dynamic scenes. We continue to decouple the deformation field and the radiance field. To enhance the editability and rendering quality of intermediate states in dynamic scenes, we have adapted this modeling approach to fit within the framework of differentiable point-based rendering.


\subsection{Acceleration of Neural Rendering}
\par Real-time rendering has long been a pivotal objective in the field of computer graphics, a goal that is also pursued in the domain of neural rendering. Numerous studies dedicated to NeRF acceleration have meticulously navigated the trade-off between spatial and temporal efficiency.

\par Pre-computed methods \cite{garbin2021fastnerf, reiser2021kilonerf} utilize spatial acceleration structures such as spherical harmonics coefficients \cite{yu2021plenoctrees} and feature vectors \cite{hedman2021baking}, cached or distilled from implicit neural representation, as opposed to directly employing the neural representations themselves. A prominent technique \cite{chen2022mobilenerf} in this category transforms NeRF scenes into an amalgamation of coarse meshes and feature textures, thereby enhancing rendering velocity in contemporary mobile graphics pipelines. However, this pre-computed approach may necessitate significant storage capacities for individual scenes. While it offers advantages in terms of inference speed, it demands protracted training durations and exhibits considerable overhead.

\par Hybrid methods \cite{liu2020neural, martel2021acorn, sun2022direct, Chen2022ECCV, wang2023f2nerf, barron2023zipnerf} incorporate a neural component within the explicit grid. The hybrid approaches confer the dual benefits of expediting both training and inference phases while producing outcomes on par with advanced frameworks \cite{barron2021mipnerf, barron2022mipnerf360}. This is primarily attributed to the robust representational capabilities of the grid.  This grid or plane-based strategy has been extended to the acceleration \cite{TiNeuVox} or representation of time-conditioned 4D feature \cite{shao2023tensor4d, kplanes_2023, Cao2023HexPlane} in dynamic scene modeling and time-conditioned compact 4D dynamic scene modeling.

\par Recently, several studies \cite{keselman2022fuzzy, zhang2022differentiable} have evolved the continuous radiance field from implicit representations to differentiable point-based radiance fields, markedly enhancing the rendering speed. 3D-GS \cite{kerbl3Dgaussians} further innovates point-based rendering by introducing a customized CUDA-based differentiable Gaussian rasterization pipeline. This approach not only achieves superior outcomes in tasks like novel-view synthesis and scene modeling, but also facilitates rapid training times on the order of minutes, and supports real-time rendering surpassing 100 FPS. However, the method employs a customized differential Gaussian rasterization pipeline, which complicates its direct extension to dynamic scenes. Inspired by this, our work will leverage the point-based rendering framework, 3D-GS, to expedite both the training and rendering speeds for dynamic scene modeling.
\section{Method}
\par The overview of our method is illustrated in Fig. \ref{fig:pipeline}. The input to our method is a set of images of a monocular dynamic scene, together with the time label and the corresponding camera poses calibrated by SfM \cite{schonberger2016structure} which also produces a sparse point cloud. From these points, we create a set of 3D Gaussians $G(\boldsymbol{x}, \boldsymbol{r}, \boldsymbol{s}, \sigma)$ defined by a center position $\boldsymbol{x}$, opacity $\sigma$, and 3D covariance matrix $\Sigma$ obtained from quaternion $\boldsymbol{r}$ and scaling $\boldsymbol{s}$. The view-dependent appearance of each 3D Gaussian is represented via spherical harmonics (SH). To model the dynamic 3D Gaussians that vary over time, we decouple the 3D Gaussians and the deformation field. The deformation field takes the positions of the 3D Gaussians and the current time $t$ as inputs, outputting $\delta \boldsymbol{x}$, $\delta \boldsymbol{r}$, and $\delta \boldsymbol{s}$. Subsequently, we put the deformed 3D Gaussians $G(\boldsymbol{x} + \delta \boldsymbol{x}, \boldsymbol{r} + \delta \boldsymbol{r}, \boldsymbol{s} + \delta \boldsymbol{s}, \sigma)$ into the efficient differential Gaussian rasterization pipeline, which is a tile-based rasterizer that allows $\alpha$-blending of anisotropic splats. The 3D Gaussians and deformation network are optimized jointly through the fast backward pass by tracking accumulated $\alpha$ values, together with the adaptive control of the Gaussian density. 
Experimental results show that after 30k training iterations, the shape of the 3D Gaussians stabilizes, as does the canonical space, which indirectly proves the efficacy of our design.


\begin {table*}[ht]
\centering
\resizebox{\textwidth}{!}{
\begin{tabular}{ccccccccccccccccc}
\hline \multirow[b]{2}{*}{ Method } & \multicolumn{3}{c}{ Hell Warrior } & \multicolumn{3}{c}{ Mutant } & \multicolumn{3}{c}{ Hook } & \multicolumn{3}{c}{ Bouncing Balls } \\
& PSNR$\uparrow$ & SSIM$\uparrow$ & LPIPS$\downarrow$ & PSNR$\uparrow$ & SSIM$\uparrow$ & LPIPS$\downarrow$ & PSNR$\uparrow$ & SSIM$\uparrow$ & LPIPS$\downarrow$ & PSNR$\uparrow$ & SSIM $\uparrow$ & LPIPS$\downarrow$ \\

\hline 3D-GS & \cellcolor{yzythird}29.89 & 0.9155 & 0.1056 & 24.53 & 0.9336 & 0.0580 & 21.71 & 0.8876 & 0.1034 & 23.20 & 0.9591 & 0.0600 \\

D-NeRF & 24.06 & 0.9440 & \cellcolor{yzysecond}0.0707 & 30.31 & \cellcolor{yzythird}0.9672 & \cellcolor{yzythird}0.0392 & \cellcolor{yzythird}29.02 & \cellcolor{yzythird}0.9595 & \cellcolor{yzysecond}0.0546 & 38.17 & 0.9891 & \cellcolor{yzythird}0.0323 \\

TiNeuVox & 27.10 & \cellcolor{yzysecond}0.9638 & 0.0768 & \cellcolor{yzythird}31.87 & 0.9607 & 0.0474 & \cellcolor{yzysecond}30.61 & \cellcolor{yzysecond}0.9599 & \cellcolor{yzythird}0.0592 & \cellcolor{yzysecond}40.23 & \cellcolor{yzythird}0.9926 & 0.0416 \\

Tensor4D & \cellcolor{yzysecond}31.26 & 0.9254 & \cellcolor{yzythird}0.0735 & 29.11 & 0.9451 & 0.0601 & 28.63 & 0.9433 & 0.0636 & 24.47 & 0.9622 & 0.0437 \\

K-Planes & 24.58 & \cellcolor{yzythird}0.9520 & 0.0824 & \cellcolor{yzysecond}32.50 & \cellcolor{yzysecond}0.9713 & \cellcolor{yzysecond}0.0362 & 28.12 & 0.9489 & 0.0662 & \cellcolor{yzythird}40.05 & \cellcolor{yzysecond}0.9934 & \cellcolor{yzysecond}0.0322 \\

Ours & \cellcolor{yzybest}41.54 & \cellcolor{yzybest}0.9873 & \cellcolor{yzybest}0.0234 & \cellcolor{yzybest}42.63 & \cellcolor{yzybest}0.9951 & \cellcolor{yzybest}0.0052 & \cellcolor{yzybest}37.42 & \cellcolor{yzybest}0.9867 & \cellcolor{yzybest}0.0144 & \cellcolor{yzybest}41.01 & \cellcolor{yzybest}0.9953 & \cellcolor{yzybest}0.0093 \\

\hline & \multicolumn{3}{c}{ Lego } & \multicolumn{3}{c}{ T-Rex } & \multicolumn{3}{c}{ Stand Up } & \multicolumn{3}{c}{ Jumping Jacks } \\

Method & PSNR$\uparrow$ & SSIM$\uparrow$ & LPIPS$\downarrow$ & PSNR$\uparrow$ & SSIM$\uparrow$ & LPIPS$\downarrow$ & PSNR$\uparrow$ & SSIM$\uparrow$ & LPIPS$\downarrow$ & PSNR$\uparrow$ & SSIM $\uparrow$ & LPIPS$\downarrow$ \\

\hline 3D-GS & 22.10 & \cellcolor{yzythird}0.9384 & \cellcolor{yzythird}0.0607 & 21.93 & 0.9539 & 0.0487 & 21.91 & 0.9301 & 0.0785 & 20.64 & 0.9297 & 0.0828 \\

D-NeRF & 25.56 & 0.9363 & 0.0821 & \cellcolor{yzythird}30.61 & \cellcolor{yzythird}0.9671 & 0.0535 & \cellcolor{yzythird}33.13 & 0.9781 & 0.0355 & \cellcolor{yzythird}32.70 & \cellcolor{yzysecond}0.9779 & \cellcolor{yzysecond}0.0388 \\

TiNeuVox & \cellcolor{yzythird}26.64 & 0.9258 & 0.0877 & \cellcolor{yzysecond}31.25 & 0.9666 & \cellcolor{yzythird}0.0478 & \cellcolor{yzysecond}34.61 & \cellcolor{yzysecond}0.9797 & \cellcolor{yzythird}0.0326 & \cellcolor{yzysecond}33.49 & \cellcolor{yzythird}0.9771 & \cellcolor{yzythird}0.0408 \\

Tensor4D & 23.24 & 0.9183 & 0.0721 & 23.86 & 0.9351 & 0.0544 & 30.56 & 0.9581 & 0.0363 & 24.20 & 0.9253 & 0.0667 \\

K-Planes & \cellcolor{yzysecond}28.91 & \cellcolor{yzysecond}0.9695 & \cellcolor{yzysecond}0.0331 & 30.43 & \cellcolor{yzysecond}0.9737 & \cellcolor{yzysecond}0.0343 & 33.10 & \cellcolor{yzythird}0.9793 & \cellcolor{yzysecond}0.0310 & 31.11 & 0.9708 & 0.0468 \\

Ours & \cellcolor{yzybest}33.07 & \cellcolor{yzybest}0.9794 & \cellcolor{yzybest}0.0183 & \cellcolor{yzybest}38.10 & \cellcolor{yzybest}0.9933 & \cellcolor{yzybest}0.0098 & \cellcolor{yzybest}44.62 & \cellcolor{yzybest}0.9951 & \cellcolor{yzybest}0.0063 & \cellcolor{yzybest}37.72 & \cellcolor{yzybest}0.9897 & \cellcolor{yzybest}0.0126 \\

\hline
\end{tabular}}
\vspace{-5pt}
\caption{\textbf{Quantitative comparison on synthetic dataset}. We compare our method to several previous approaches: 3D-GS \cite{kerbl3Dgaussians}, D-NeRF 
\cite{pumarola2021d}, TiNeuVox \cite{TiNeuVox}, Tensor4D \cite{shao2023tensor4d} and K-Planes \cite{kplanes_2023} on full resolution (800x800) test images. This may cause some methods to perform worse than the original paper because they downsample images by default. 
We report PSNR, SSIM, LPIPS(VGG) and color each cell as \colorbox{yzybest}{best}, \colorbox{yzysecond}{second best} and \colorbox{yzythird}{third best}. It is worth noting that we observed a discrepancy in the scenarios presented in the training and test sets of the Lego in D-NeRF dataset. This can be substantiated by examining the flip angles of the Lego shovels. To ensure a meaningful comparison, we opted to utilize the validation set of Lego as the test set in our experiments. See more in supplementary materials.}
\label{exp:dnerf}
\end{table*}

\subsection{Differentiable Rendering Through 3D Gaussians Splatting in Canonical Space}

\par To optimize the parameters of 3D Gaussians in canonical space, it is imperative to differentially render 2D images from these 3D Gaussians. In this work, we employ the differential Gaussian rasterization pipeline proposed by \cite{kerbl3Dgaussians}. Following \cite{zwicker2001ewa}, the 3D Gaussians can be projected to 2D and rendered for each pixel using the following 2D covariance matrix $\Sigma^{\prime}$:
\begin{equation}
    \Sigma^{\prime} = JV\Sigma V^TJ^T,
\end{equation}
where $J$ is the Jacobian of the affine approximation of the projective transformation, $V$ symbolizes the view matrix, transitioning from world to camera coordinates, and $\Sigma$ denotes the 3D covariance matrix.

\par To make learning the 3D Gaussians easier, $\Sigma$ is divided into two learnable components: the quaternion $\boldsymbol{r}$ represents rotation, and the 3D-vector $\boldsymbol{s}$ represents scaling. These components are then transformed into the corresponding rotation and scaling matrices $R$ and $S$. The resulting $\Sigma$ can be expressed as:
\begin{equation}
    \Sigma = RSS^TR^T.
\end{equation}

\par The color of the pixel on the image plane, denoted by \textbf{p}, is rendered sequentially with point-based volume rendering technique:
\begin{equation}
\begin{aligned}
    C (\textbf{p}) &= \sum_{i \in N} T_i \alpha_i c_i, \\
    \alpha_i &= \sigma_i e^{-\frac{1}{2} (\textbf{p} - \mu_i)^T \sum^{\prime} (\textbf{p} - \mu_i) },
\end{aligned}
\end{equation}
where $T_i$ is the transmittance defined by $\Pi_{j=1}^{i - 1}(1 - \alpha_{j}) $, $c_i$ signifies the color of the Gaussians along the ray, and $\mu_i$ represents the $uv$ coordinates of the 3D Gaussians projected onto the 2D image plane. 

\par During the optimization, adaptive density control emerges as a pivotal component, enabling the rendering of 3D Gaussians to achieve desirable outcomes. This control serves a dual purpose: firstly, it mandates the pruning of transparent Gaussians based on $\sigma$. Secondly, it necessitates the densification of Gaussian distribution. This densification fills regions void of geometric intricacies, while simultaneously subdividing areas where Gaussians are large and exhibit significant overlap. Notably, such areas tend to display pronounced positional gradients. Following \cite{kerbl3Dgaussians}, we discern the 3D Gaussians that demand adjustments using a threshold given by $t_{pos}=0.0002$. For diminutive Gaussians inadequate for capturing geometric details, we clone the Gaussians and move them a certain distance in the direction of the positional gradients. Conversely, for those that are conspicuously large and overlapping, we split them and divide their scale by a hyperparameter $\xi=1.6$.

\par It is clear that 3D Gaussians are only appropriate for representing static scenes. 
Applying a time-conditioned learnable parameter for each 3D Gaussian not only contradicts the original intent of the differentiable Gaussian rasterization pipeline, but also results in the loss of spatiotemporal continuity of motion.
To enable 3D Gaussians to represent dynamic scenes while retaining the practical physical meaning of their individual learnable components, we decided to learn 3D Gaussians in canonical space and use an additional deformation field to learn the position and shape variations of the 3D Gaussians. 


\subsection{Deformable 3D Gaussians}
\par An intuitive solution to model dynamic scenes using 3D Gaussians is to separately train 3D-GS set in each time-dependent view collection and then perform interpolation between these sets as a post-processing step. While such an approach is feasible for Multi-View Stereo (MVS) captures at discrete time, it falls short for continuous monocular captures within a temporal sequence. To deal with the latter, a more general case, we jointly learn a deformation field along with 3D Gaussians.

\par We decouple the motion and geometry structure by leveraging a deformation network alongside 3D Gaussians, converting the learning process into a canonical space to obtain time-independent 3D Gaussians. This decoupling approach introduces geometric priors of the scene, associating the changes in the positions of the 3D Gaussians with both time and coordinates.
The core of the deformation network is an MLP. In our study, we did not employ the grid/plane structures applied in static NeRF that can accelerate rendering and enhance its quality. This is because such methods operate on a \textbf{low-rank assumption}, whereas dynamic scenes possess a higher rank. Explicit point-based rendering further elevates the rank of the scene. 

\par Given time $t$ and center position $\boldsymbol{x}$ of 3D Gaussians as inputs, the deformation MLP produces offsets, which subsequently transform the canonical 3D Gaussians to the deformed space:
\begin{equation}
    (\delta \boldsymbol{x}, \delta \boldsymbol{r}, \delta \boldsymbol{s}) = \mathcal{F_\theta}(\gamma(\operatorname{sg}(\boldsymbol{x})), \gamma(t)),
\end{equation} 
where $\operatorname{sg}(\cdot)$ indicates a stop-gradient operation, $\gamma$ denotes the positional encoding:
\begin{equation}
    \gamma(p) = (sin(2^k\pi p), cos(2^k \pi p))_{k=0}^{L-1},
\end{equation}
where $L=10$ for $x$ and $L=6$ for $t$ in synthetic scenes, while $L=10$ for both $x$ and $t$ in real scenes. We set the depth of the deformation network $D=8$ and the dimension of the hidden layer $W=256$. Experiments demonstrate that applying positional encoding to the inputs of the deformation network can enhance the details in rendering results.

\subsection{Annealing Smooth Training}
\par A prevalent challenge with numerous real-world datasets is the inaccuracies in pose estimation, a phenomenon markedly evident in dynamic scenes. Training under imprecise poses can lead to overfitting on the training data. Moreover, as also mentioned in HyperNeRF \cite{park2021hypernerf}, the imprecise poses from colmap for real datasets can cause spatial jitter between each frame w.r.t. the test or train set, resulting in a noticeable deviation when rendering the test image compared to the ground truth. Previous methods that used implicit representations benefited from the MLP's inherent smoothness, making the impact of such minor offsets on the final rendering results relatively inconspicuous.
However, explicit point-based rendering tends to amplify this effect. For monocular dynamic scenes, novel-view rendering at a fixed time remains unaffected. However, for the task involving interpolated time, this kind of inconsistent scene at different times can lead to irregular rendering jitters.

\par To address this issue, we propose a novel annealing smooth training (AST) mechanism specifically designed for real-world monocular dynamic scenes:
\begin{equation} \label{eq:transformation}
\begin{aligned}
    \boldsymbol{\Delta} &= \mathcal{F}_{\boldsymbol{\theta}} \left(\bm{\gamma}(\operatorname{sg}(\boldsymbol{x})), \bm{\gamma}(t) + \mathcal{X}(i)\right), \\
    \mathcal{X}(i) &= \mathbb{N}(0, 1) \cdot \beta \cdot \Delta t \cdot \left(1 - i/{\tau}\right),
\end{aligned}
\end{equation}
where \(\mathcal{X}(i)\) represents the linearly decaying Gaussian noise at the \(i\)-th training iteration, \(\mathbb{N}(0, 1)\) denotes the standard Gaussian distribution, \(\beta\) is an empirically determined scaling factor with a value of \(0.1\), \(\Delta t\) represents the mean time interval, and \(\tau\) is the threshold iteration for annealing smooth training (empirically set to \(20k\)).


\par Compared to the smooth loss introduced by methods of \cite{shao2023tensor4d, pumarola2021d}, our approach does not incur additional computational overhead. It can enhance the model's temporal generalization in the early stages of training, as well as prevent excessive smoothing in the later stages, thus preserving the details of objects in dynamic scenes. Concurrently, it reduces the jitter observed in real datasets during time interpolation tasks.

\begin{figure*}
    \centering
    \addtolength{\tabcolsep}{-6.5pt}
    \footnotesize{
        \setlength{\tabcolsep}{1pt} 
        \begin{tabular}{p{8.2pt}cccccc}
            & GT & Ours & TiNeuVox & K-Planes & Tensor4D & D-NeRF  \\
        \raisebox{35pt}{\rotatebox[origin=c]{90}{T-Rex}}&
             \includegraphics[width=0.145\textwidth]{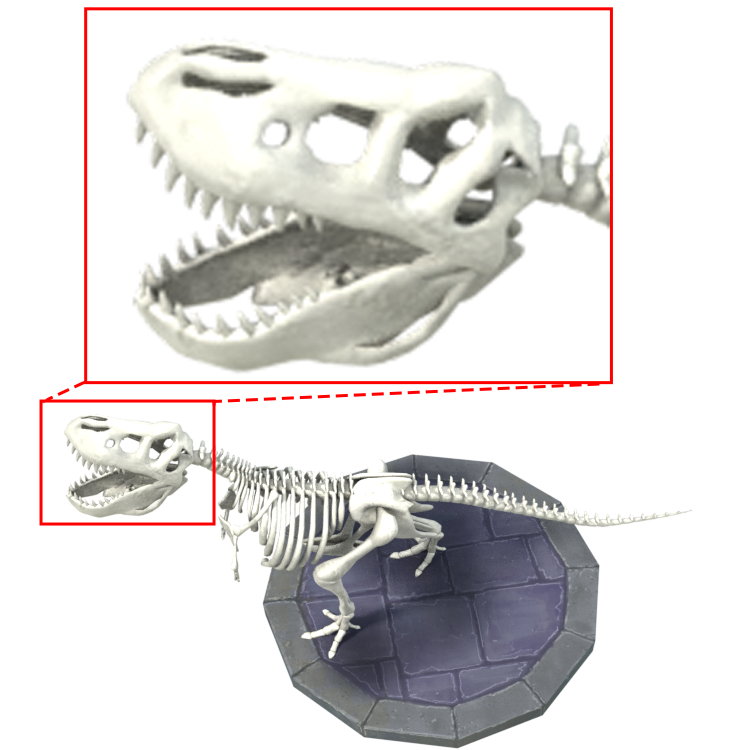} &
            \includegraphics[width=0.145\textwidth]{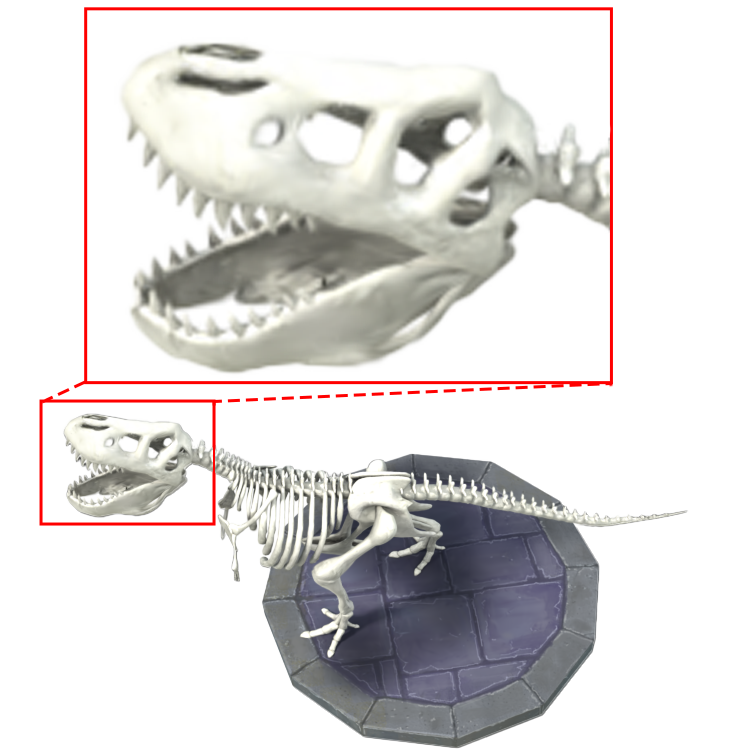} &
            \includegraphics[width=0.145\textwidth]{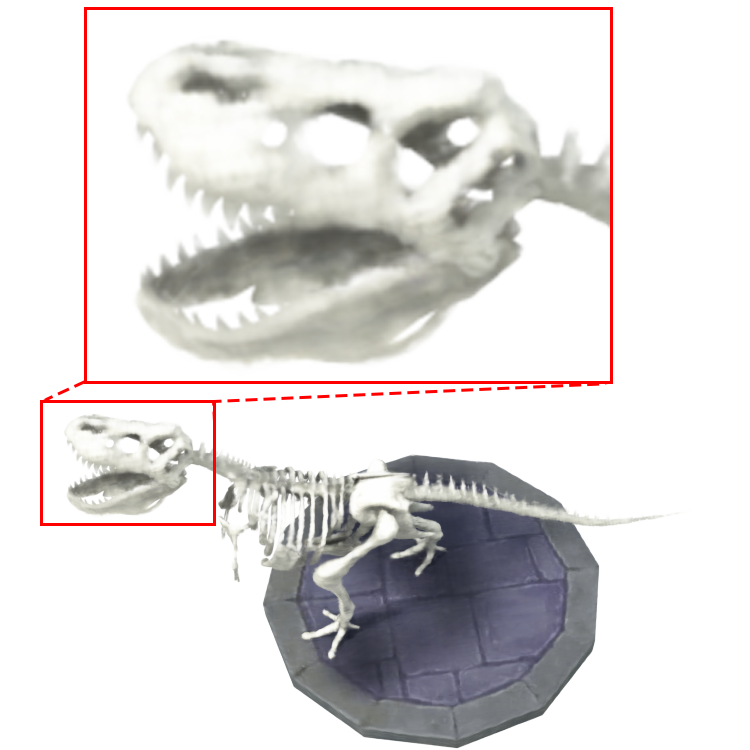} &
            \includegraphics[width=0.145\textwidth]{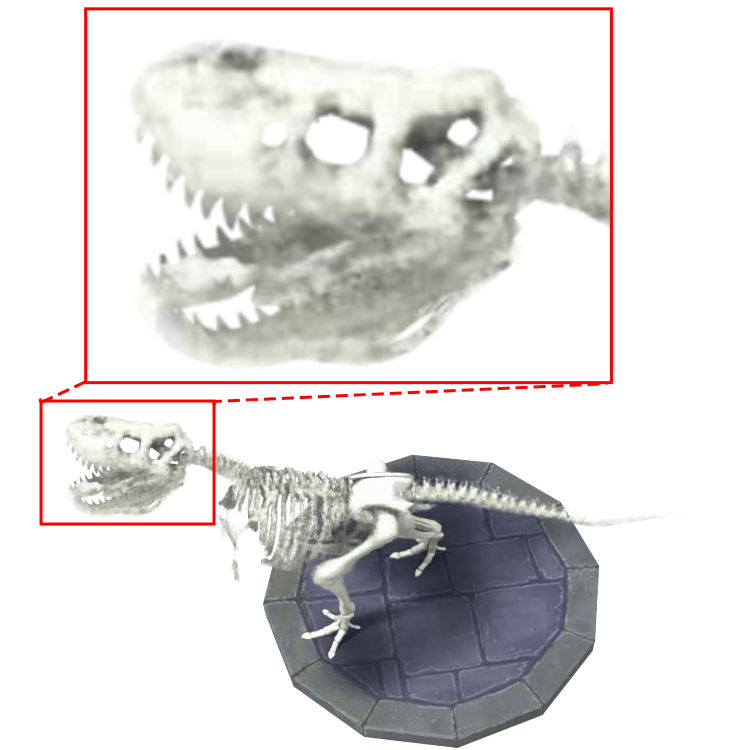} &
            \includegraphics[width=0.145\textwidth]{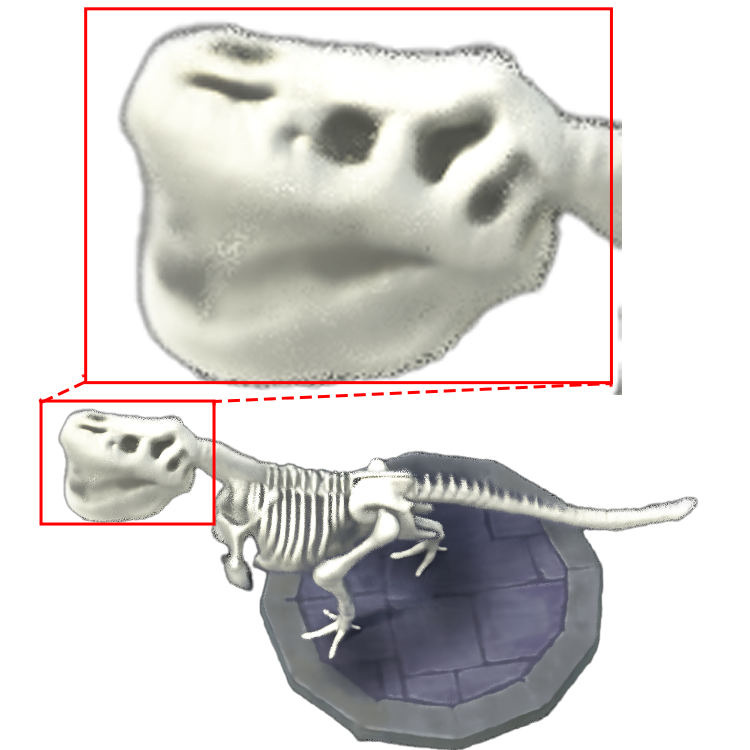} &
            \includegraphics[width=0.145\textwidth]{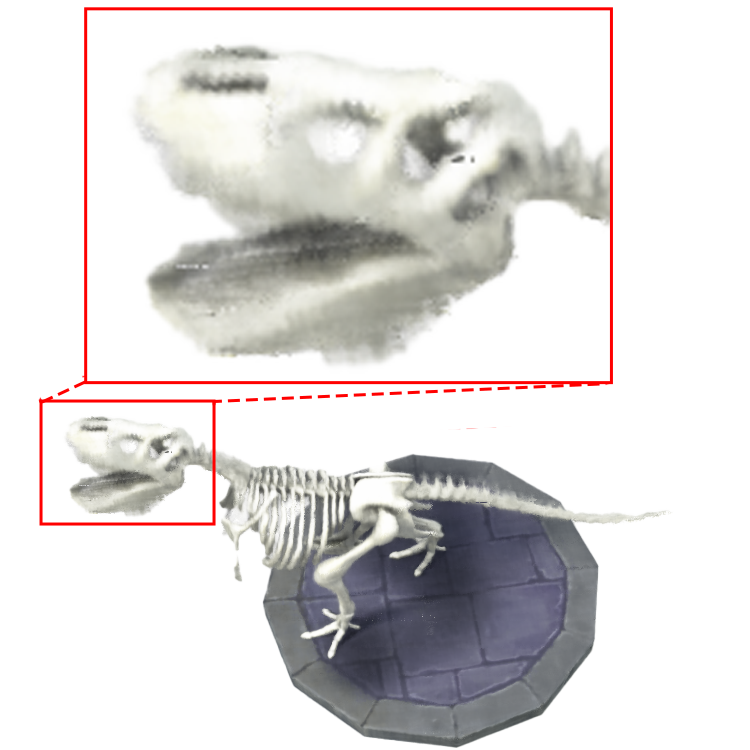}
             \\
        \raisebox{35pt}{\rotatebox[origin=c]{90}{StandUp}}&
             \includegraphics[width=0.145\textwidth]{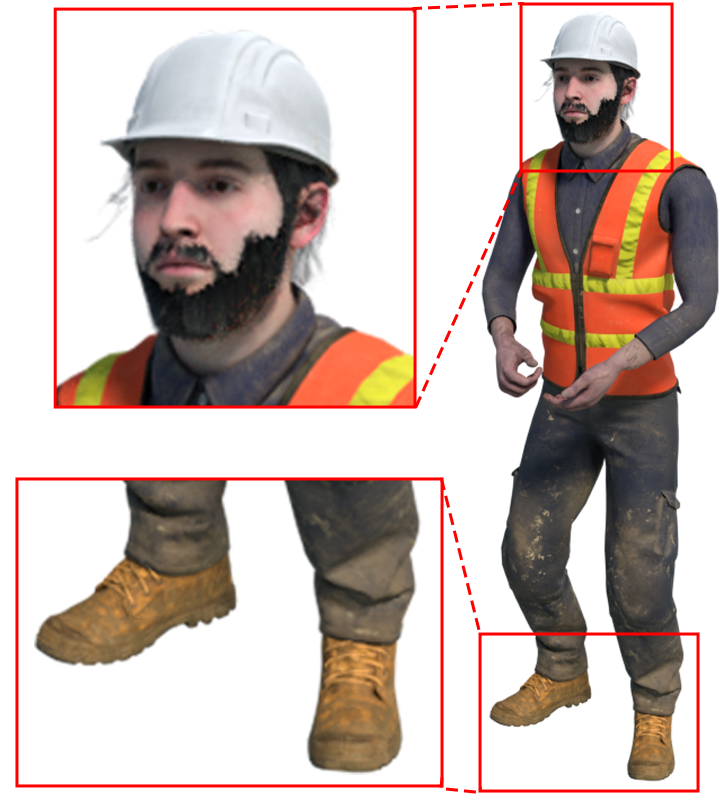} &
            \includegraphics[width=0.145\textwidth]{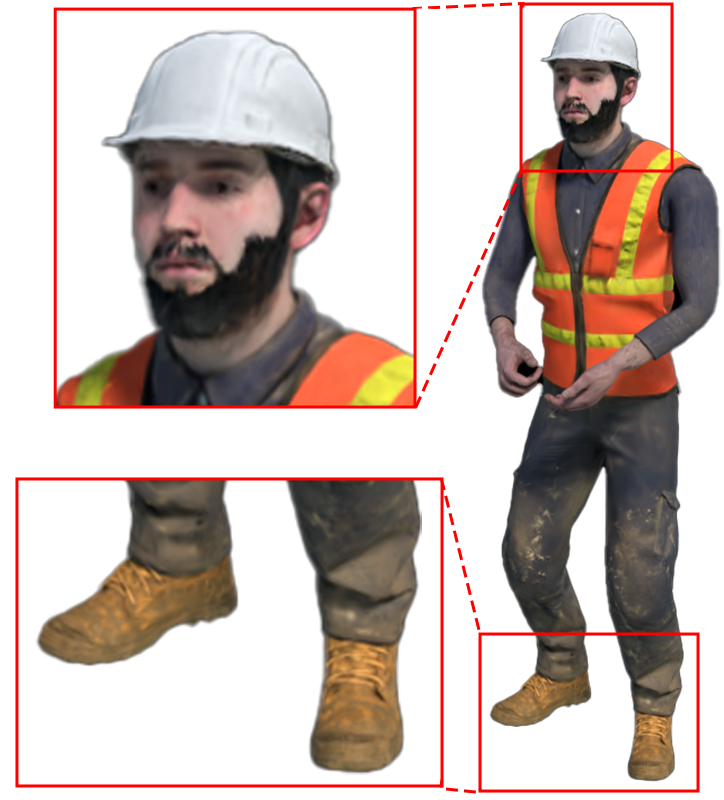} &
            \includegraphics[width=0.145\textwidth]{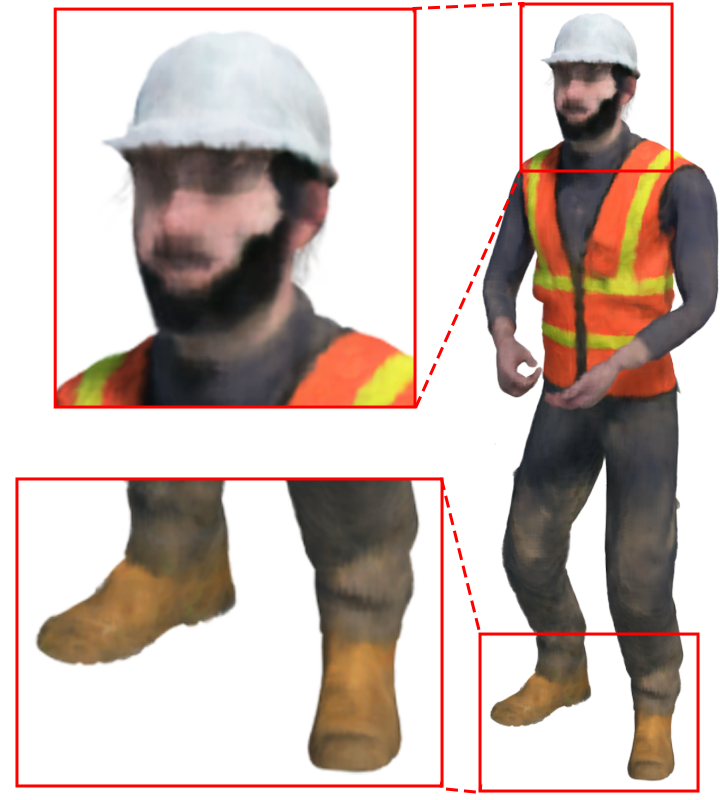} &
            \includegraphics[width=0.145\textwidth]{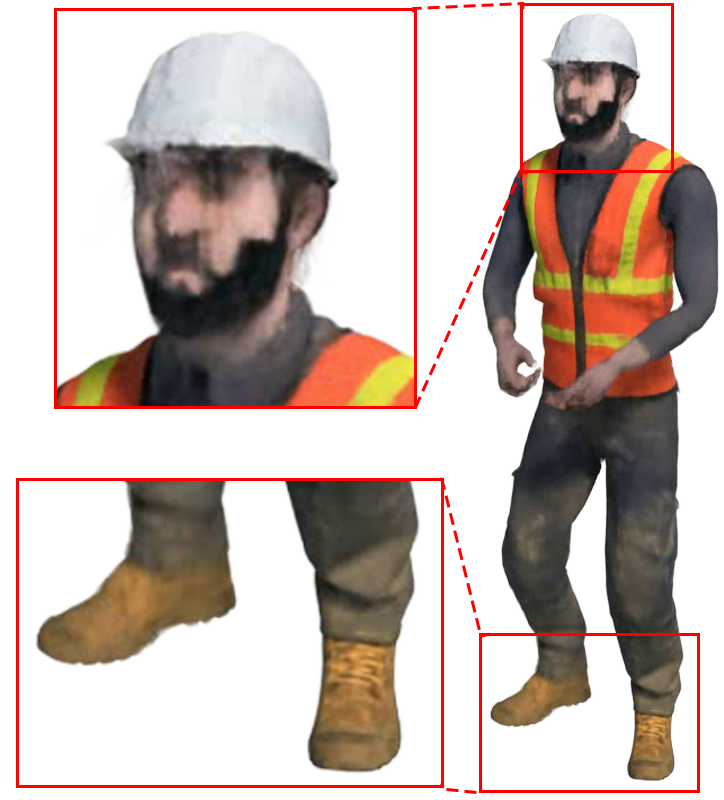} &
            \includegraphics[width=0.145\textwidth]{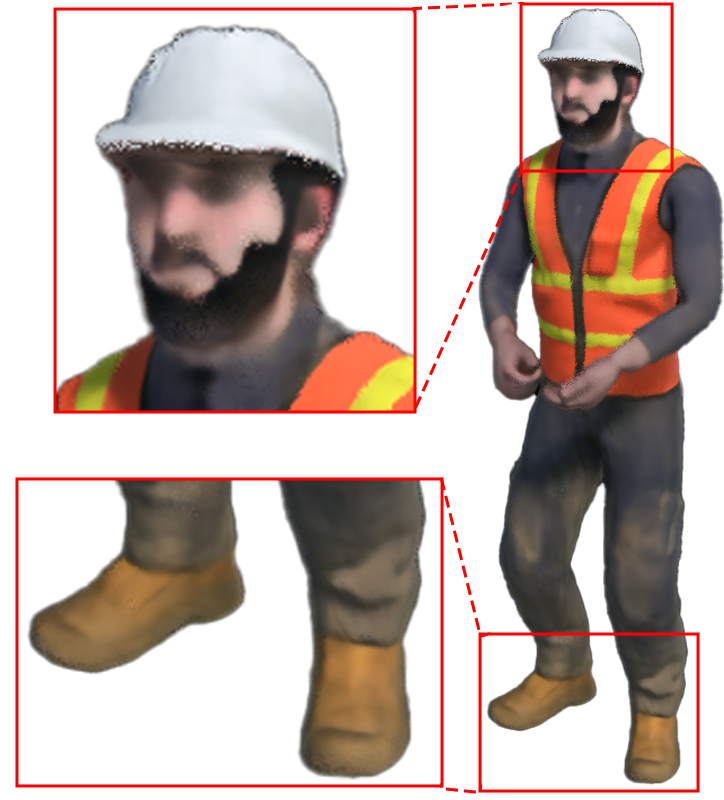} &
            \includegraphics[width=0.145\textwidth]{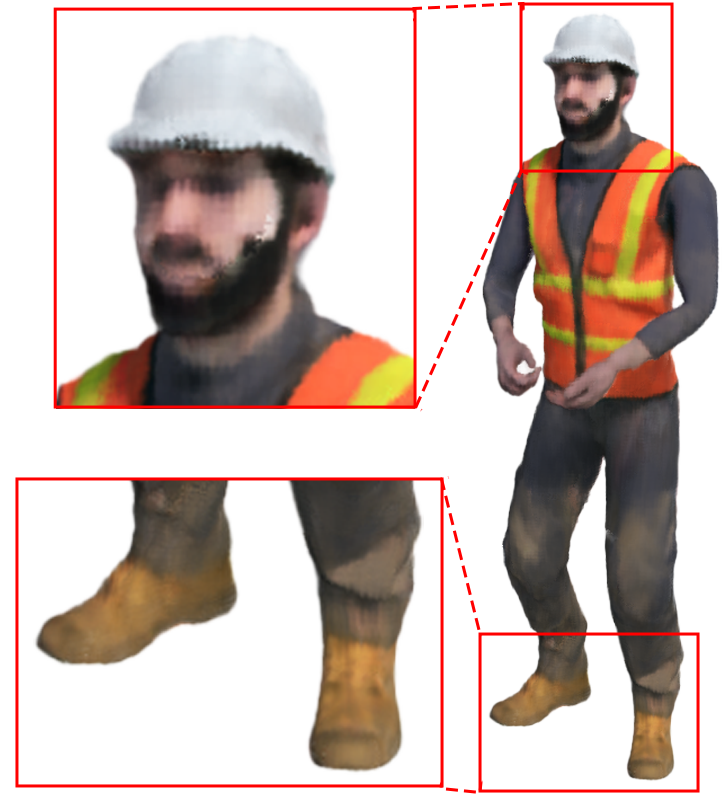}
            \\
            \raisebox{35pt}{\rotatebox[origin=c]{90}{Mutant}}&
             \includegraphics[width=0.145\textwidth]{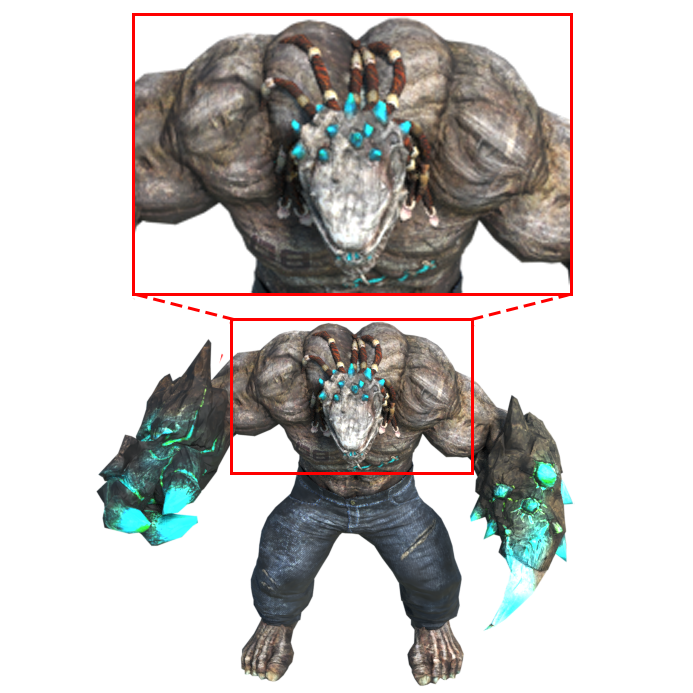} &
            \includegraphics[width=0.145\textwidth]{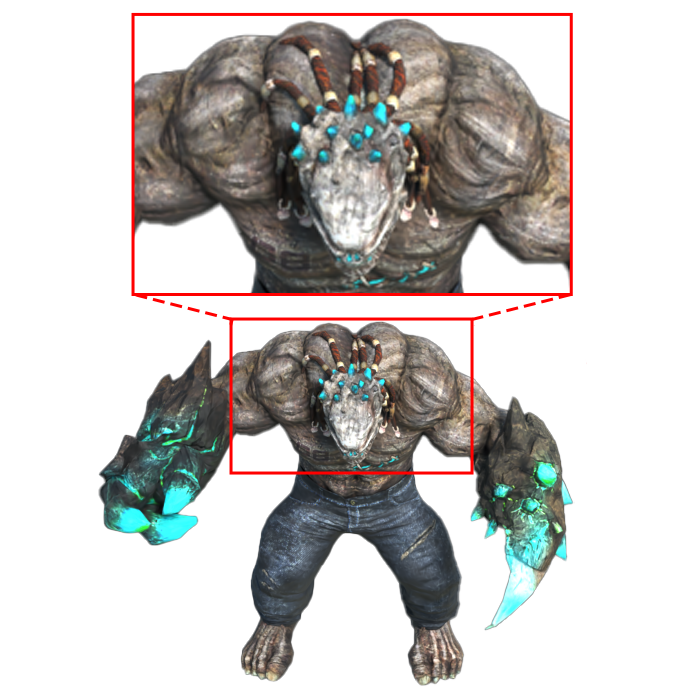} & 
            \includegraphics[width=0.145\textwidth]{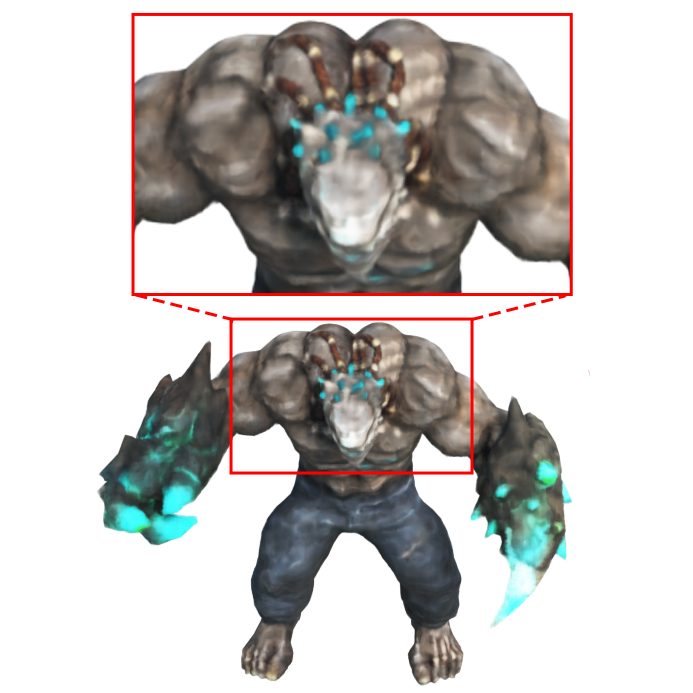} &
            \includegraphics[width=0.145\textwidth]{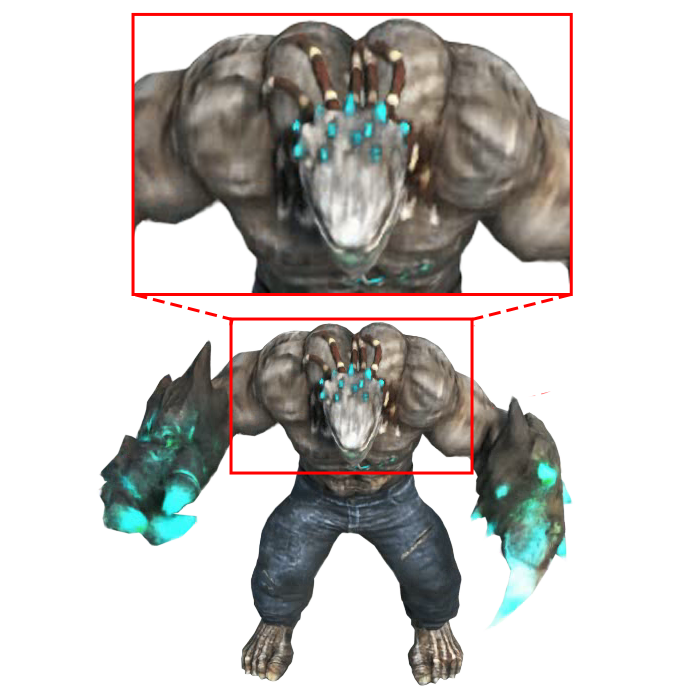} &
            \includegraphics[width=0.145\textwidth]{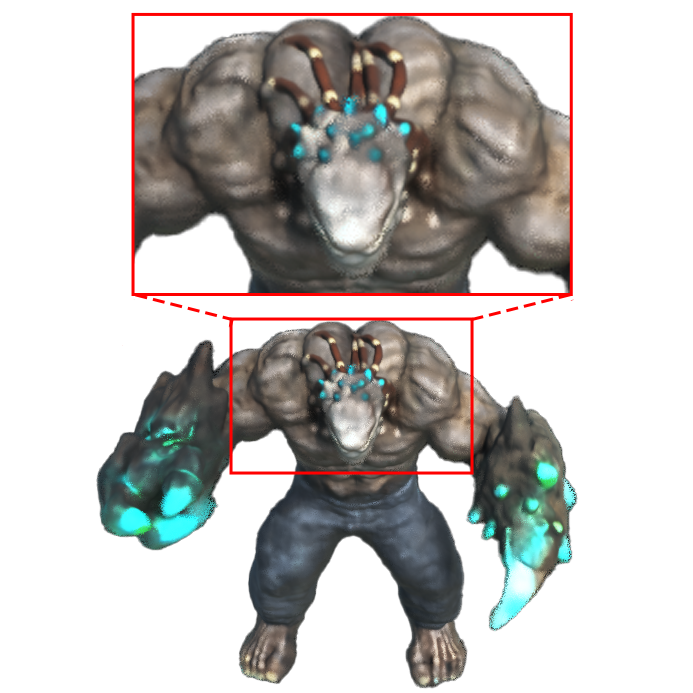} &
            \includegraphics[width=0.145\textwidth]{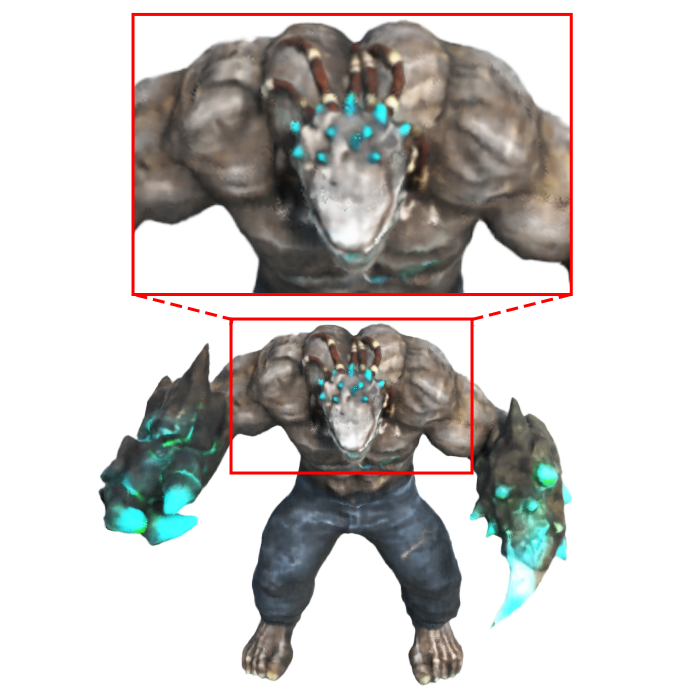}

            \\
            \raisebox{35pt}{\rotatebox[origin=c]{90}{Lego}}&
             \includegraphics[width=0.145\textwidth]{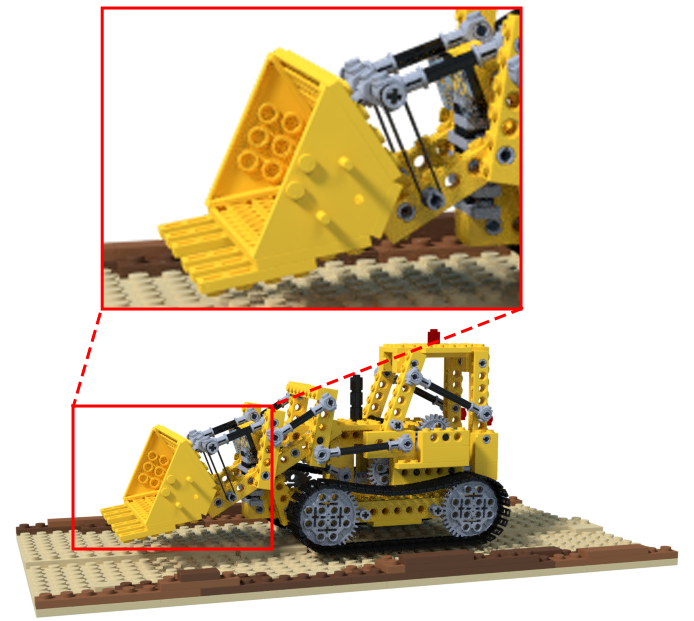} &
            \includegraphics[width=0.145\textwidth]{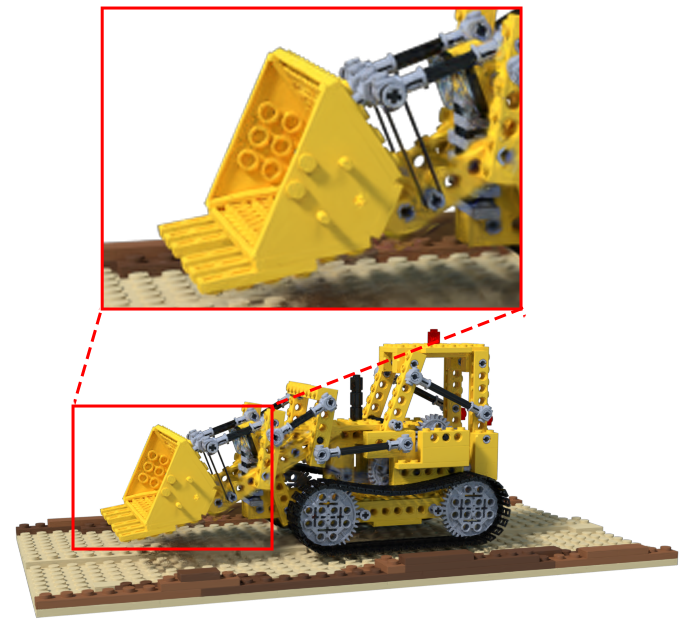} & 
            \includegraphics[width=0.145\textwidth]{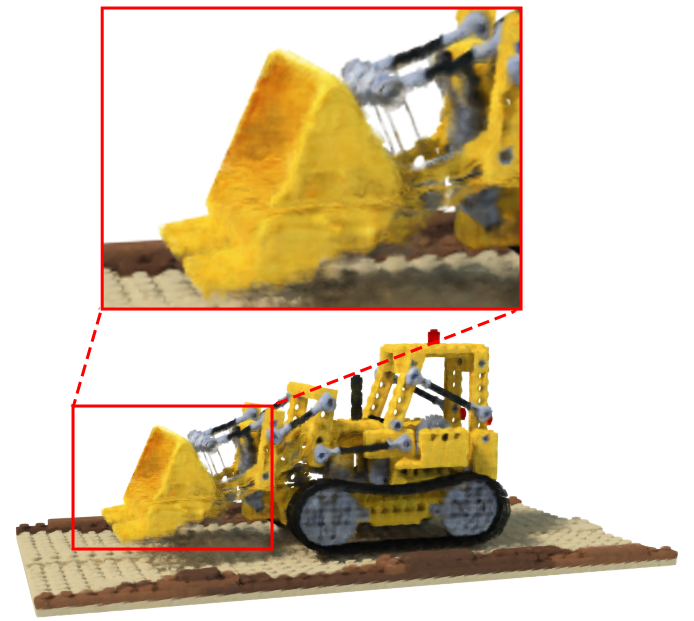} &
            \includegraphics[width=0.145\textwidth]{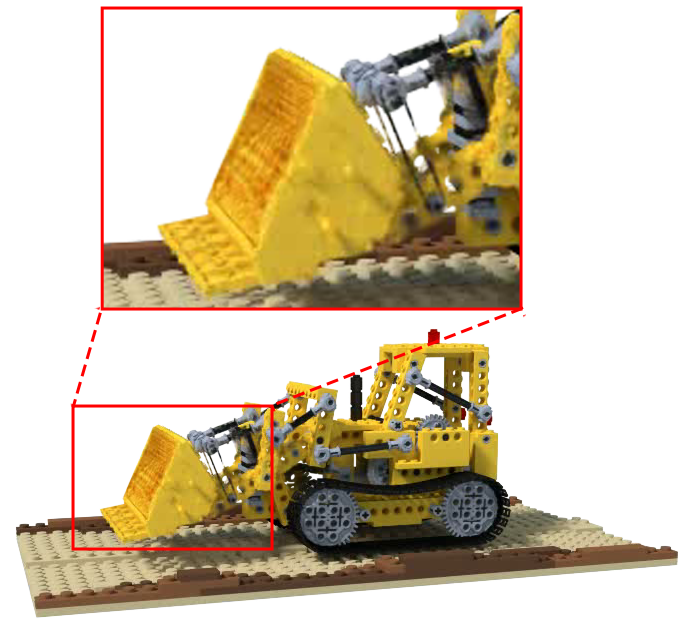} &
            \includegraphics[width=0.145\textwidth]{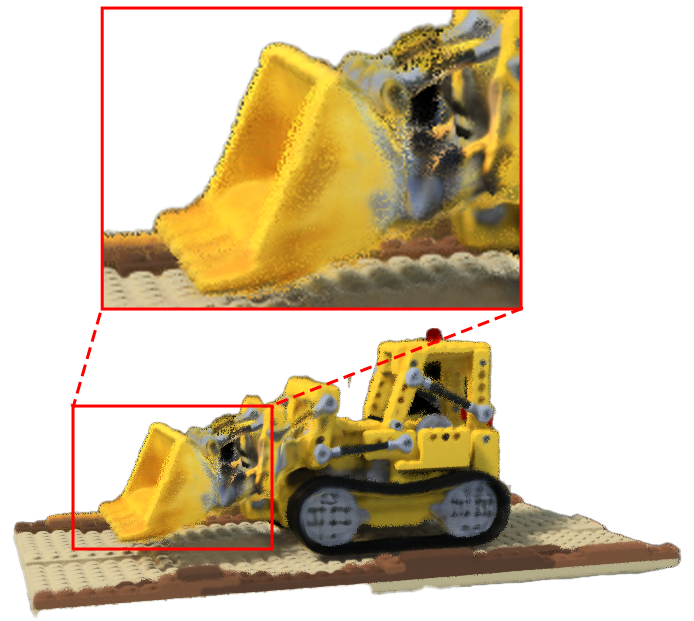} &
            \includegraphics[width=0.145\textwidth]{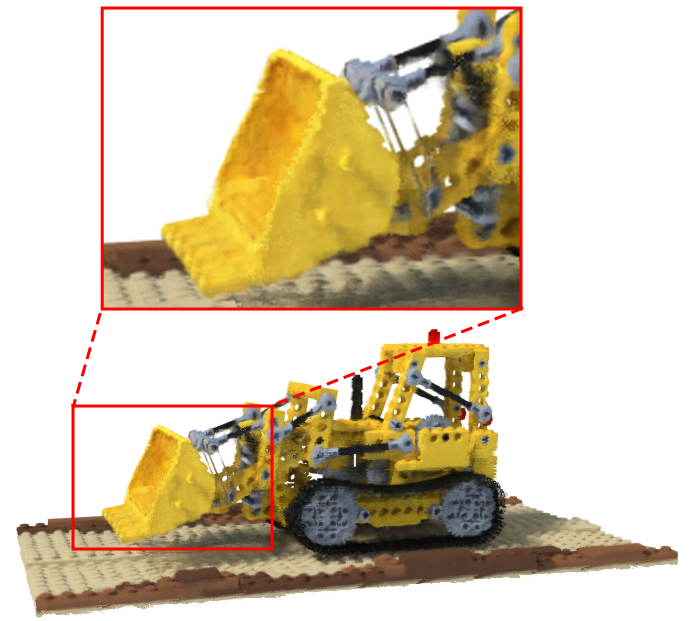}

            \\
            \raisebox{35pt}{\rotatebox[origin=c]{90}{Jumpingjacks}}&
             \includegraphics[width=0.145\textwidth]{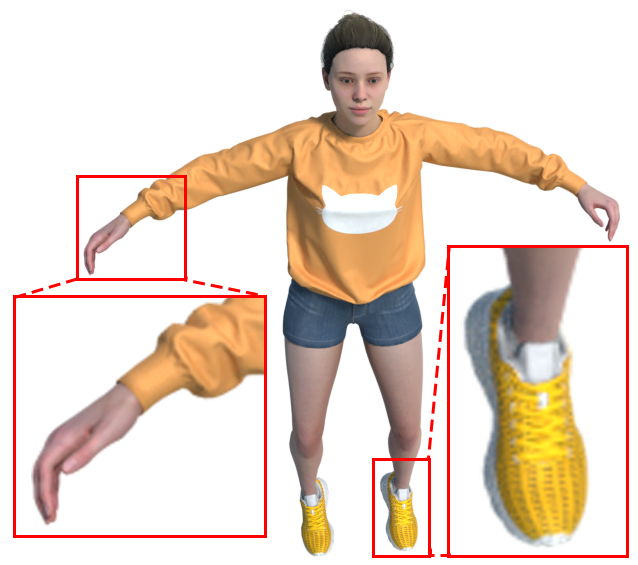} &
            \includegraphics[width=0.145\textwidth]{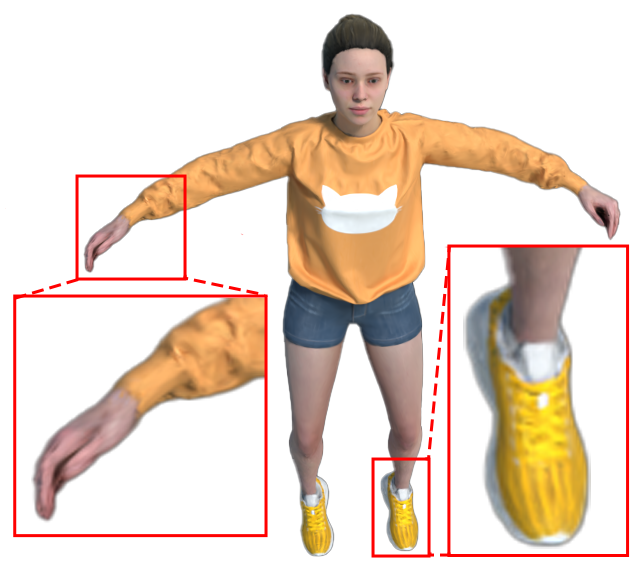} & 
            \includegraphics[width=0.145\textwidth]{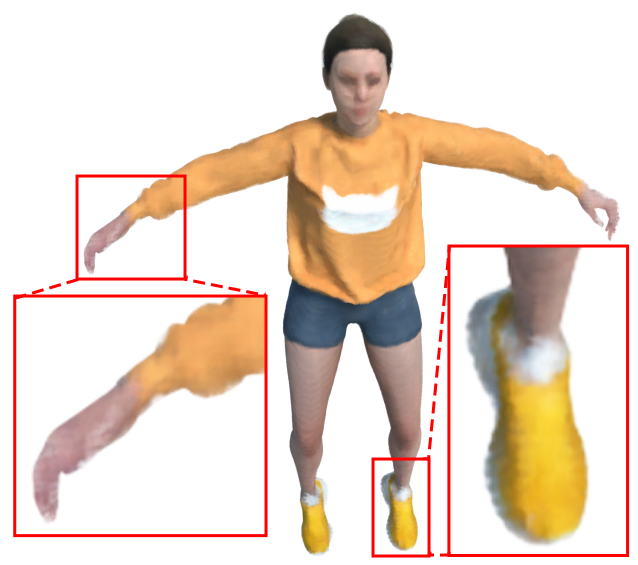} &
            \includegraphics[width=0.145\textwidth]{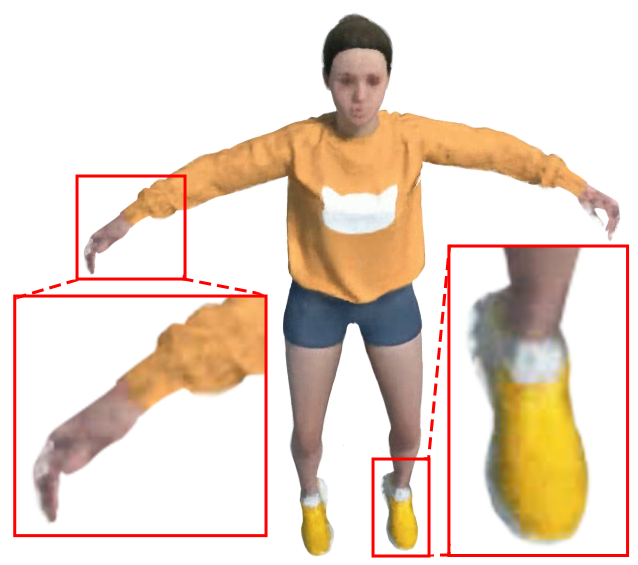} &
            \includegraphics[width=0.145\textwidth]{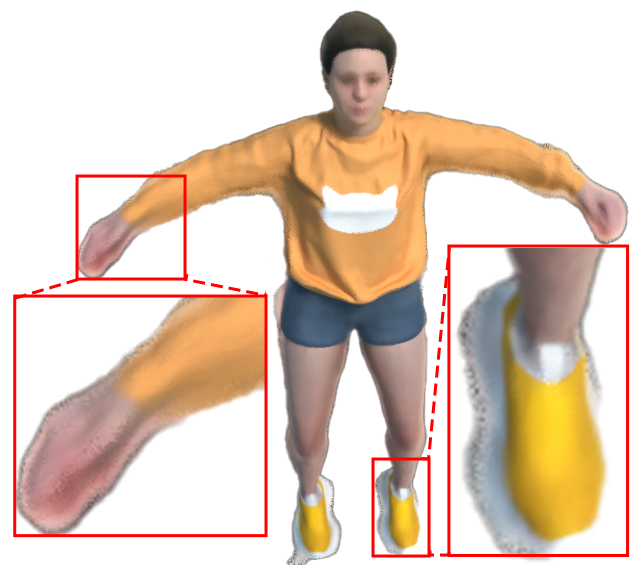} &
            \includegraphics[width=0.145\textwidth]{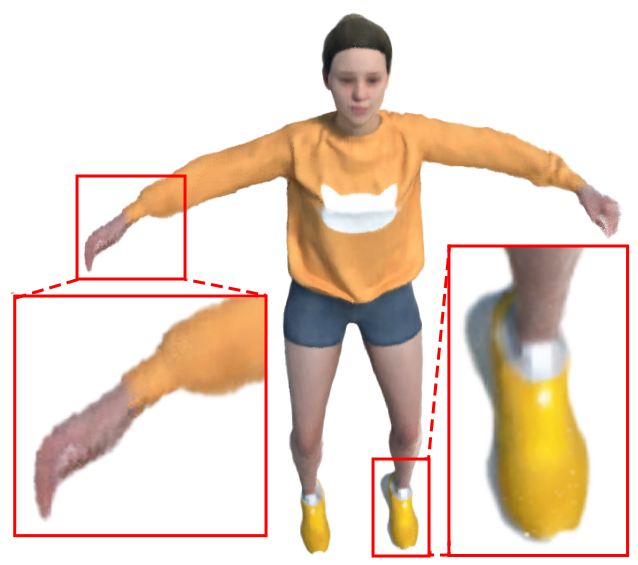}

            \\
            \raisebox{35pt}{\rotatebox[origin=c]{90}{hook}}&
             \includegraphics[width=0.145\textwidth]{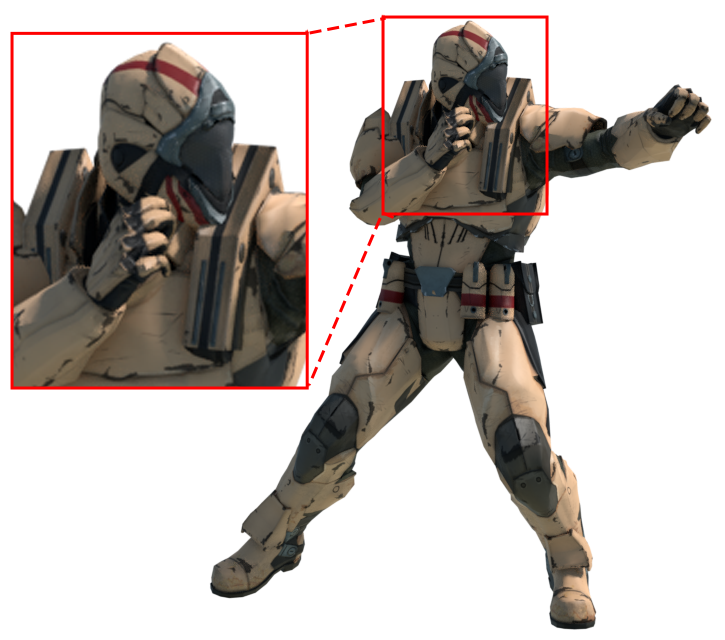} &
            \includegraphics[width=0.145\textwidth]{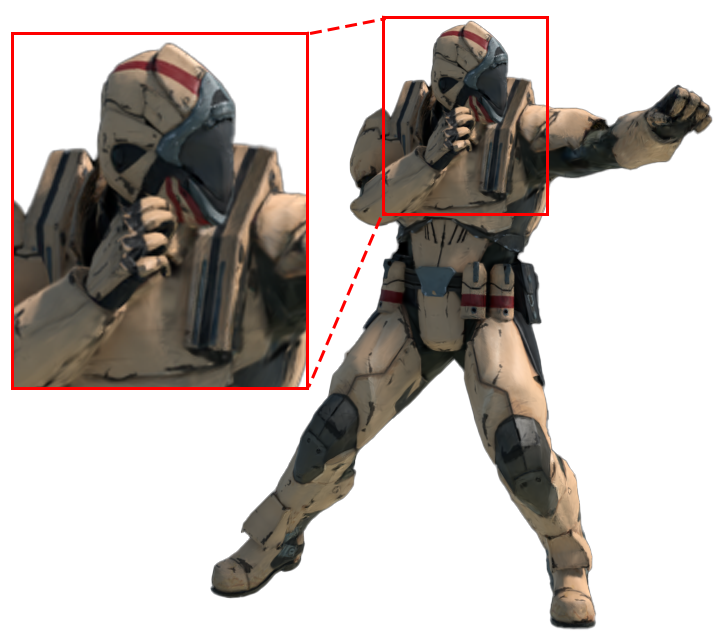} & 
            \includegraphics[width=0.145\textwidth]{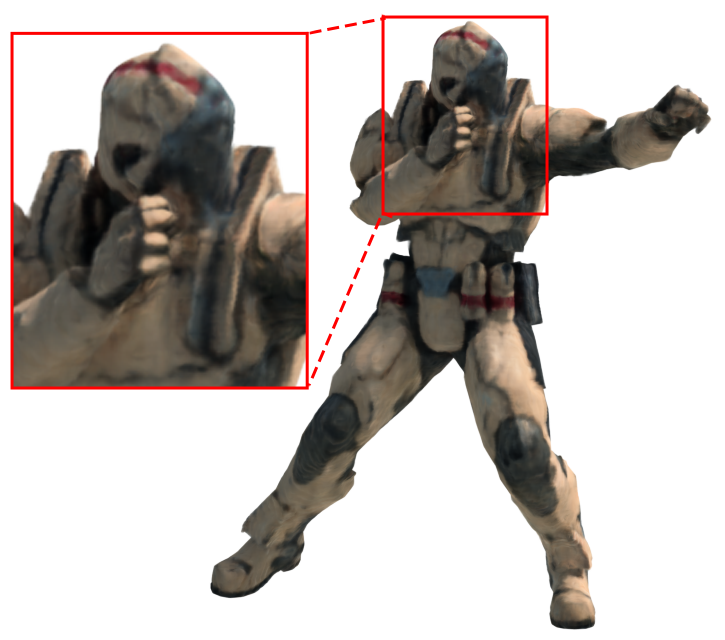} &
            \includegraphics[width=0.145\textwidth]{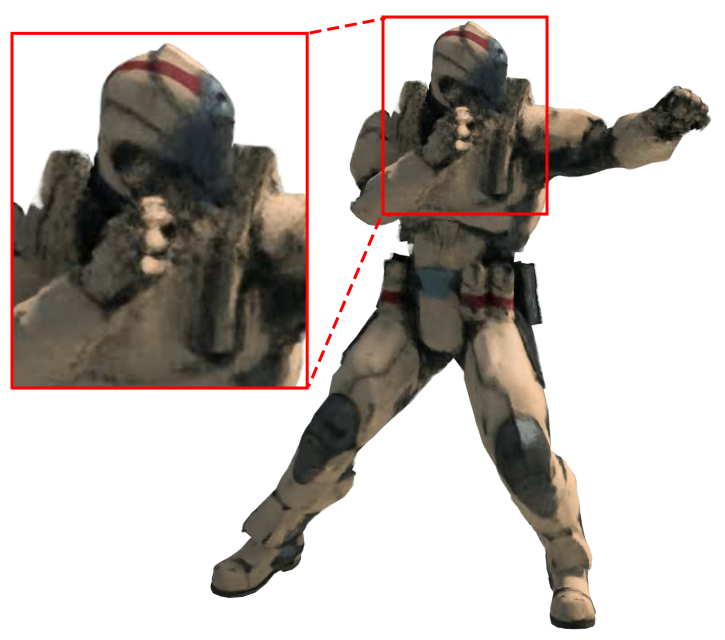} &
            \includegraphics[width=0.145\textwidth]{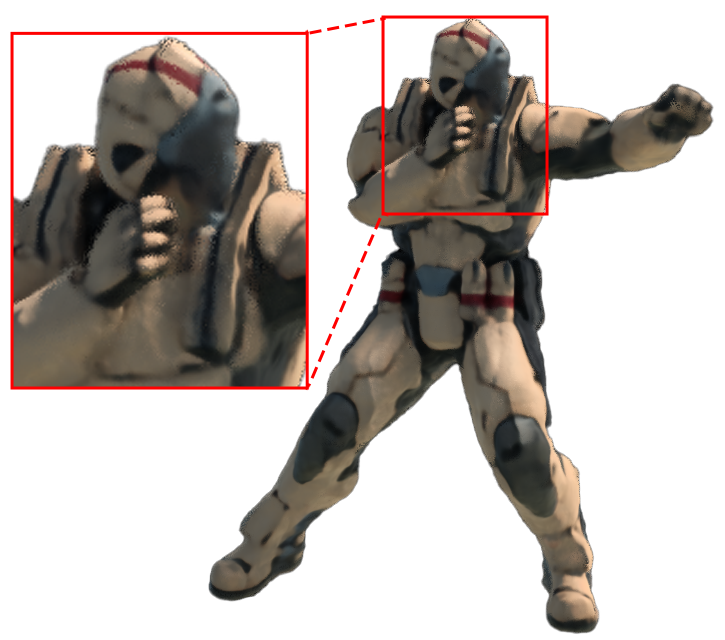} &
            \includegraphics[width=0.145\textwidth]{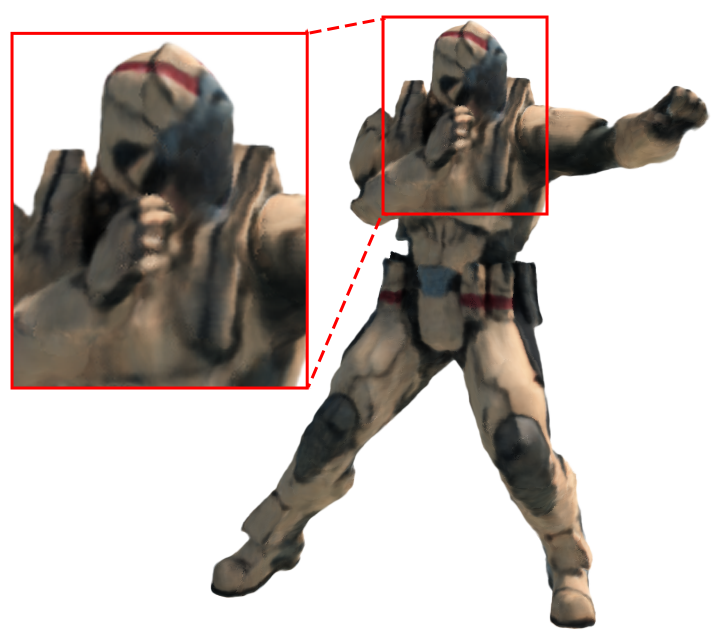}

            \\
            \raisebox{35pt}{\rotatebox[origin=c]{90}{HellWarrior}}&
             \includegraphics[width=0.145\textwidth]{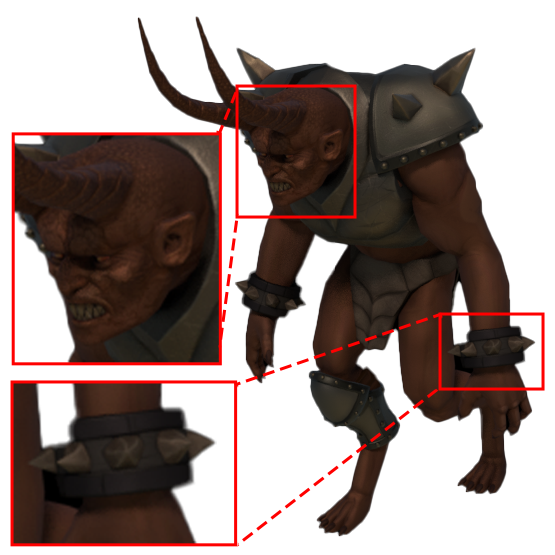} &
            \includegraphics[width=0.145\textwidth]{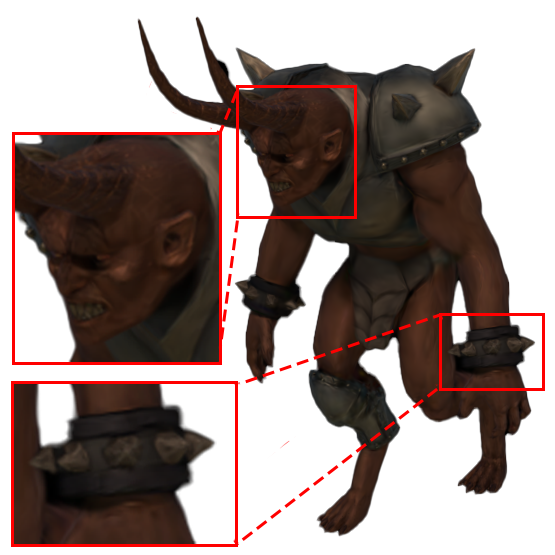} & 
            \includegraphics[width=0.145\textwidth]{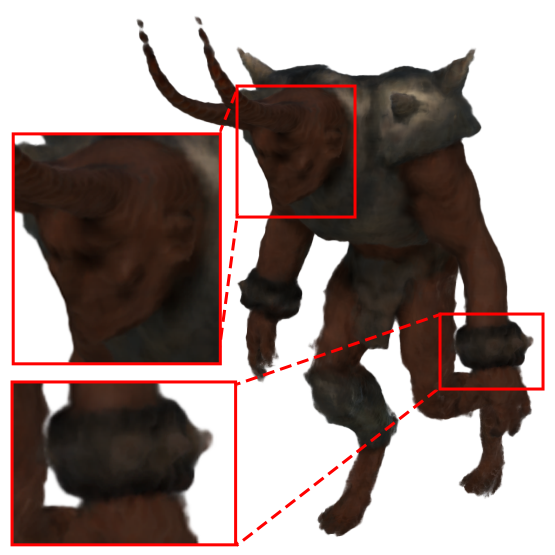} &
            \includegraphics[width=0.145\textwidth]{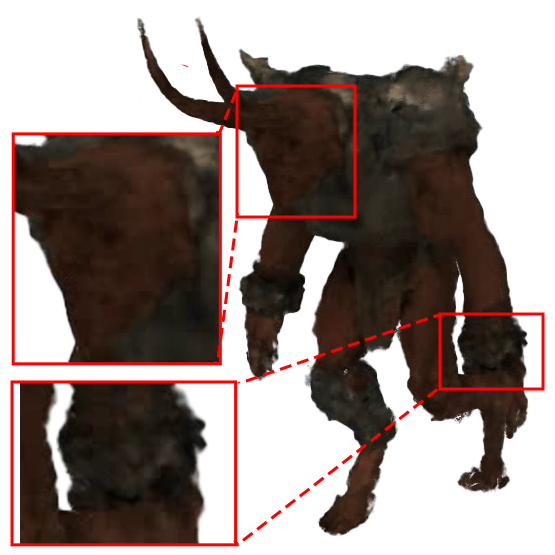} &
            \includegraphics[width=0.145\textwidth]{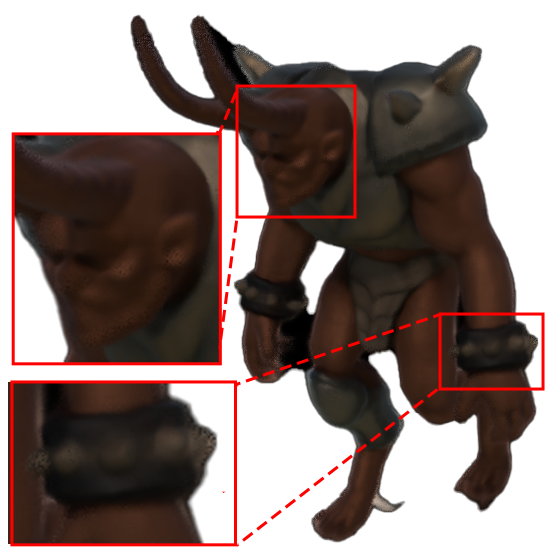} &
            \includegraphics[width=0.145\textwidth]{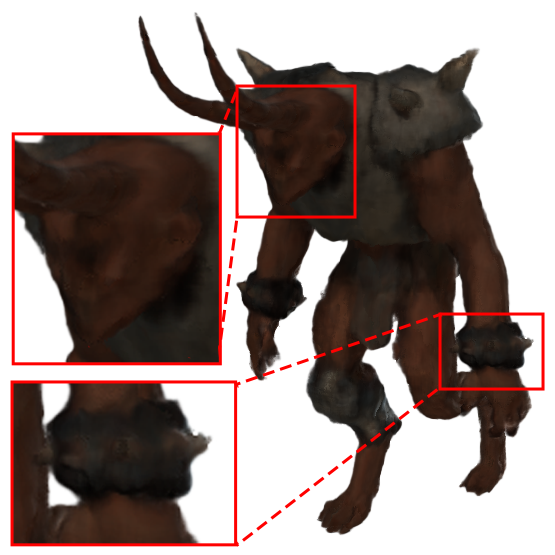}

            \\
            \raisebox{35pt}{\rotatebox[origin=c]{90}{BouncingBalls}}&
             \includegraphics[width=0.145\textwidth]{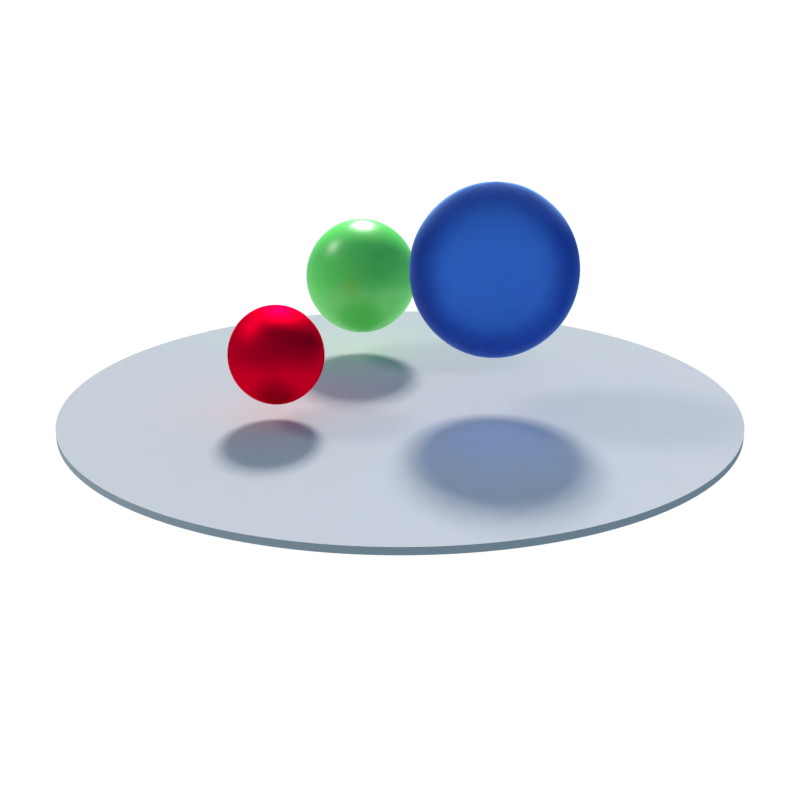} &
            \includegraphics[width=0.145\textwidth]{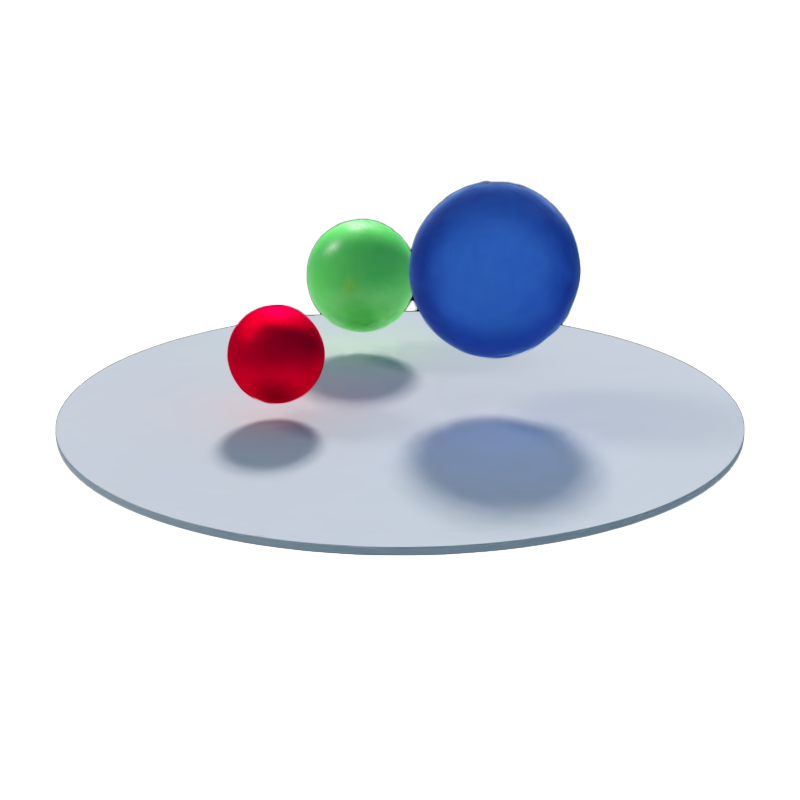} & 
            \includegraphics[width=0.145\textwidth]{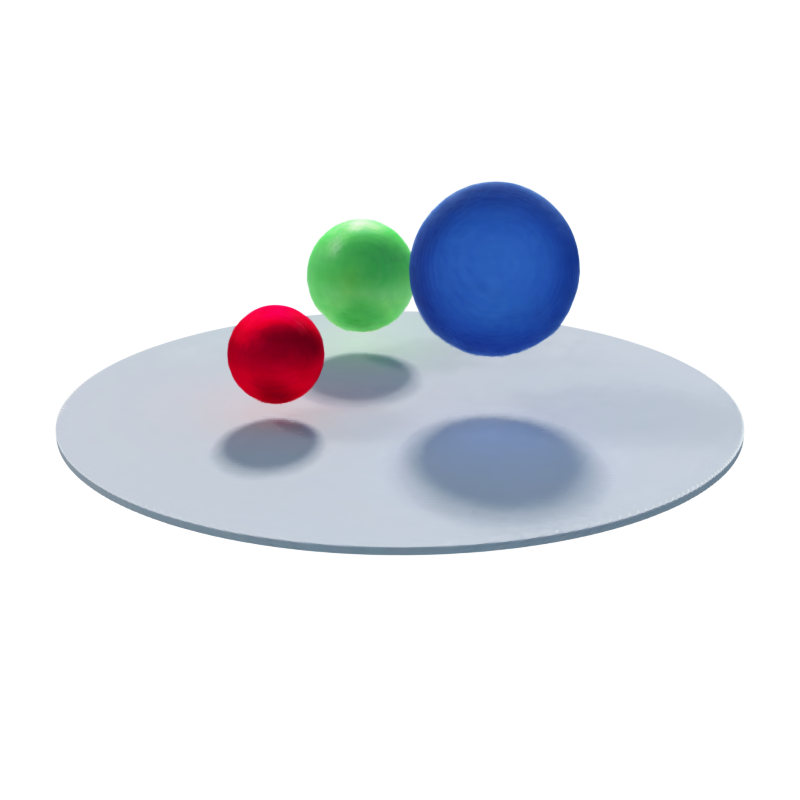} &
            \includegraphics[width=0.145\textwidth]{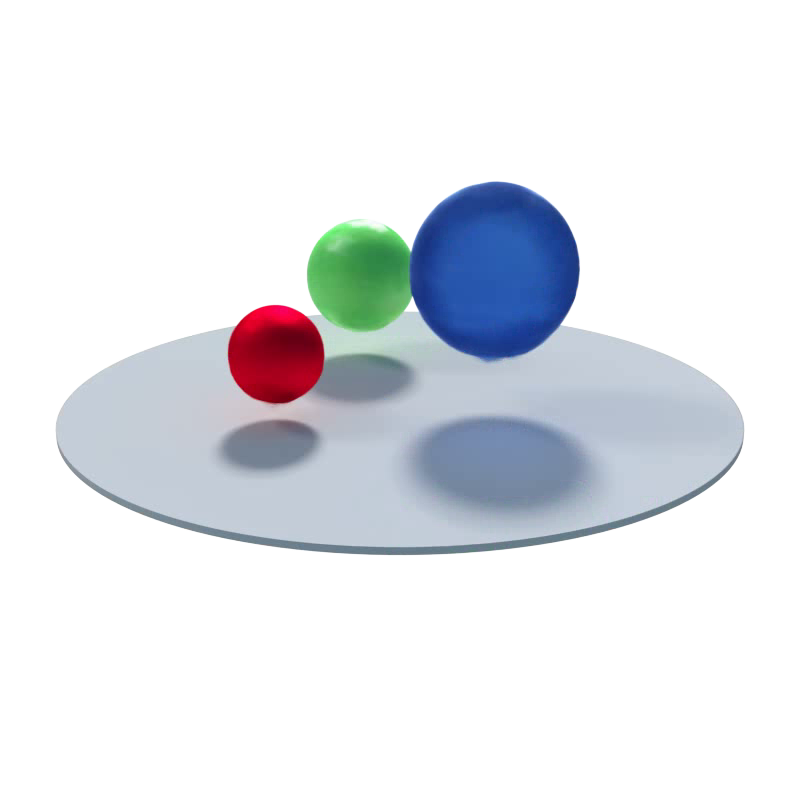} &
            \includegraphics[width=0.145\textwidth]{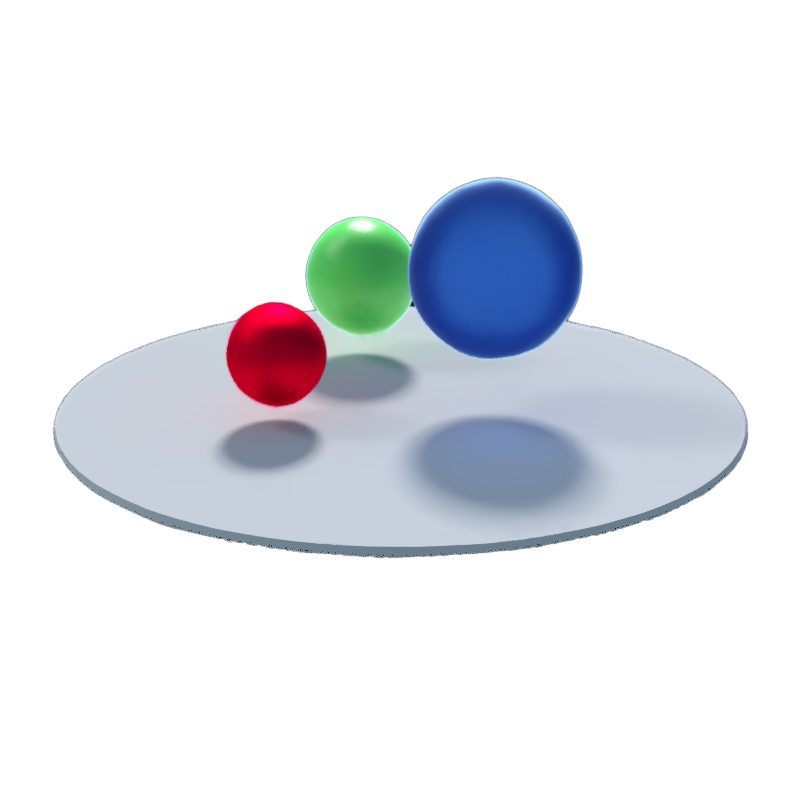} &
            \includegraphics[width=0.145\textwidth]{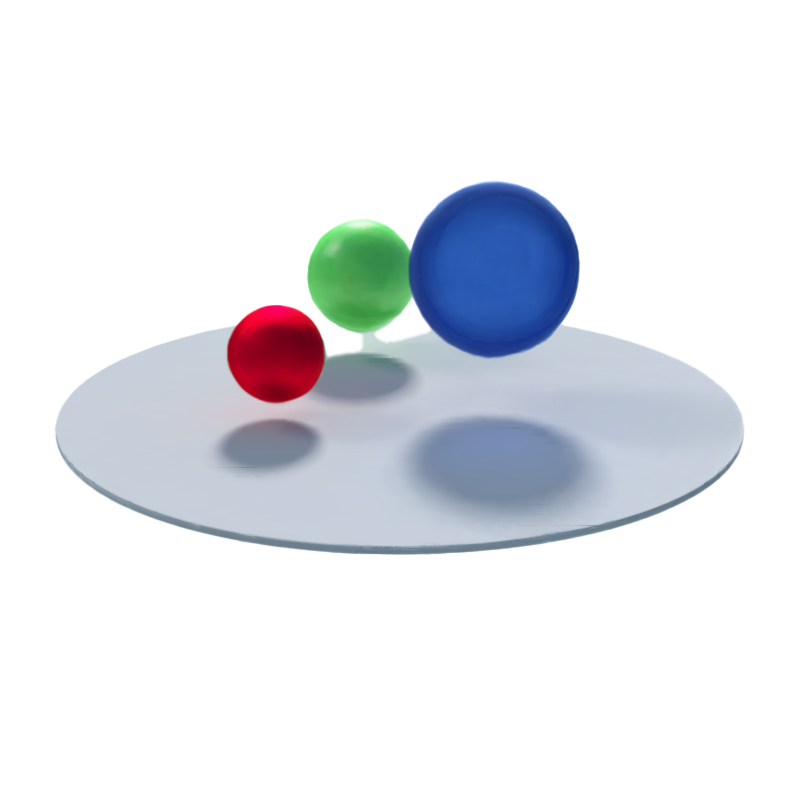}

        \end{tabular}
    }
\vspace{-5pt}
	\caption{\textbf{Qualitative comparisons of baselines and our method on monocular synthetic dataset.} We visualize each scene using baselines and our method.
Experimental results indicate that our approach recovers more details when rendering novel viewpoints and can reconstruct more delicate structures over time, such as hands or skeletons. The efficacy of Deformable-GS can be attributed to its capability to equally back-propagate the gradient to both the Deformation Field and the 3D Gaussians. Larger gradients in the dynamic portion of the Deformation Field can further assist the 3D Gaussians in achieving better densification in dynamic regions.
}\label{fig:syn-quality}
\end{figure*}

\begin{figure*}[ht] 
  \centering
  \includegraphics[width=0.92\textwidth]{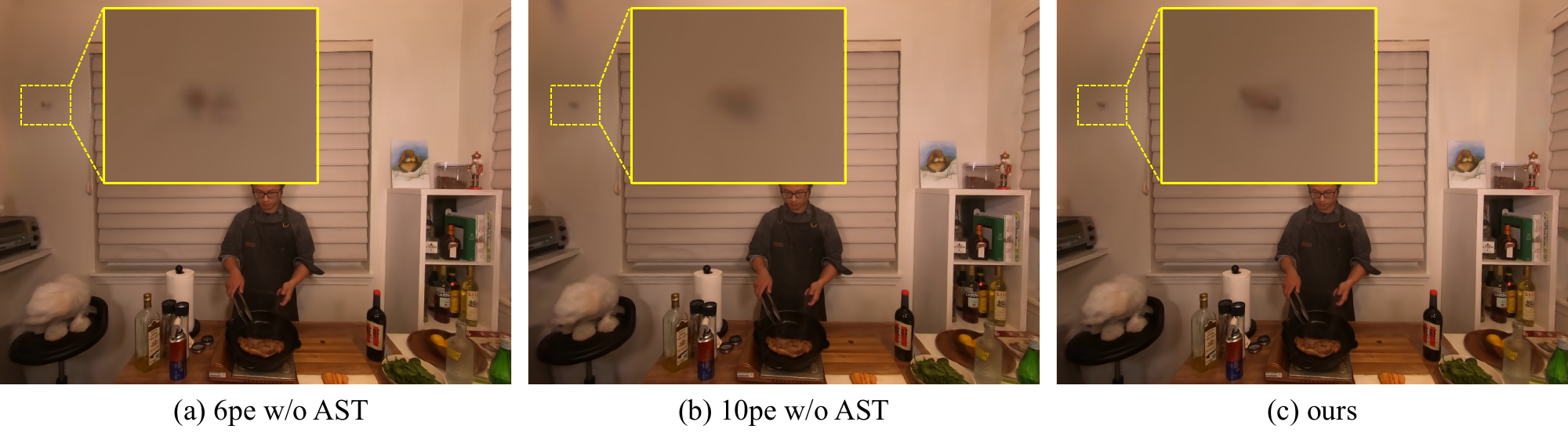}
    \caption{\textbf{Ablation study.} We conduct ablation studies focusing on the annealing smooth training scheme within real-world datasets, wherein \emph{pe} signifies the positional encoding over time. Compared with the reduced order (a) and the original order (b) of positional encoding over time, it becomes evident that the annealing smooth training strategy (c) effectively preserves high-frequency information. Simultaneously, it mitigates the temporal overfitting challenges instigated by imprecise pose estimations.}\label{fig:ablation}
\end{figure*}

\begin{figure*}
    \centering
    \addtolength{\tabcolsep}{-6.5pt}
    \footnotesize{
        \setlength{\tabcolsep}{1pt} 
        \begin{tabular}{p{8.2pt}cccccccc}
            & GT & Ours & TiNeuVox & HyperNeRF & NeRF-DS & 3D-GS  \\
        \raisebox{18pt}{\rotatebox[origin=c]{90}{as}}&
         \includegraphics[width=0.155\textwidth]{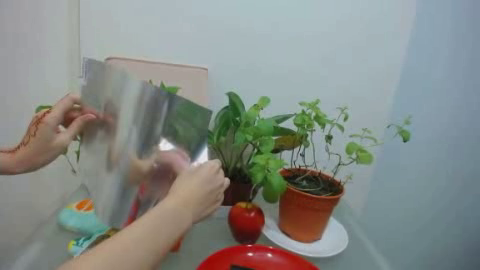} &
        \includegraphics[width=0.155\textwidth]{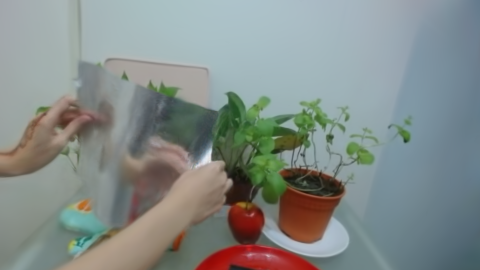} &
        \includegraphics[width=0.155\textwidth]{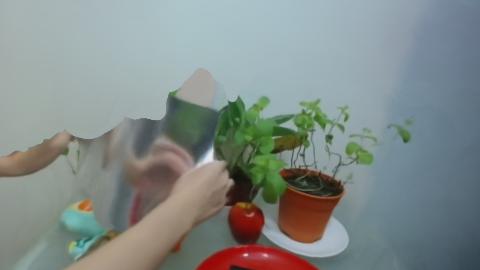} &
        \includegraphics[width=0.155\textwidth]{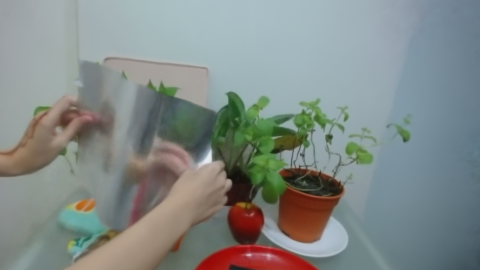} &
        \includegraphics[width=0.155\textwidth]{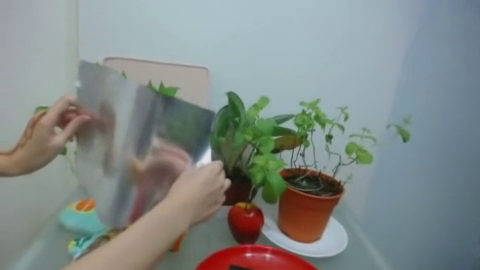} &
        \includegraphics[width=0.155\textwidth]{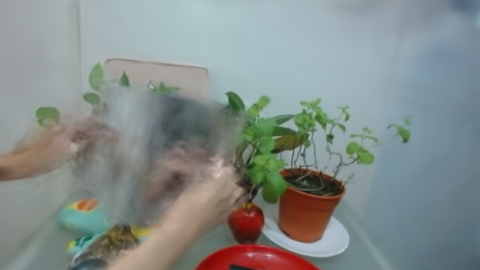}
         \\

        \raisebox{20pt}{\rotatebox[origin=c]{90}{bell}}&
         \includegraphics[width=0.155\textwidth]{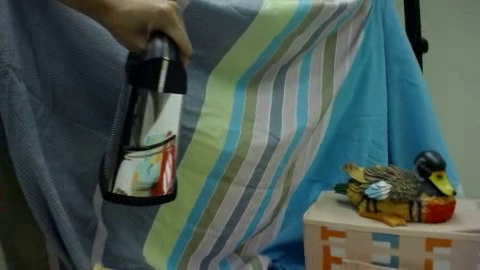} &
        \includegraphics[width=0.155\textwidth]{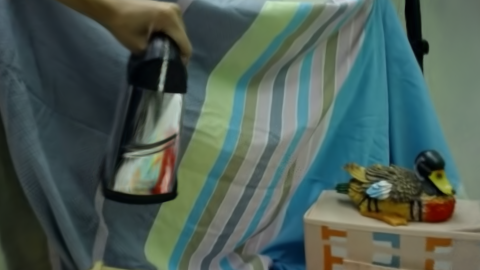} & 
        \includegraphics[width=0.155\textwidth]{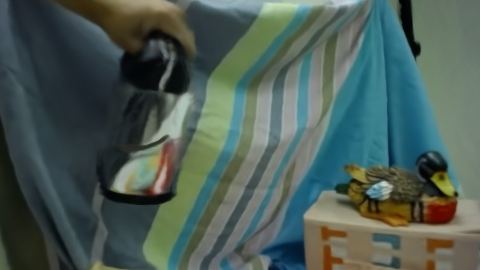} &
        \includegraphics[width=0.155\textwidth]{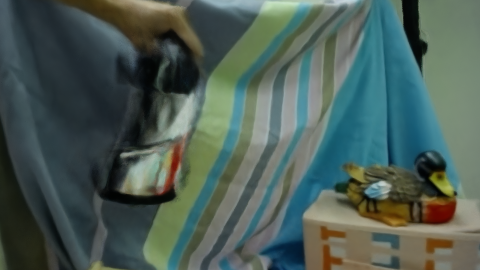} &
        \includegraphics[width=0.155\textwidth]{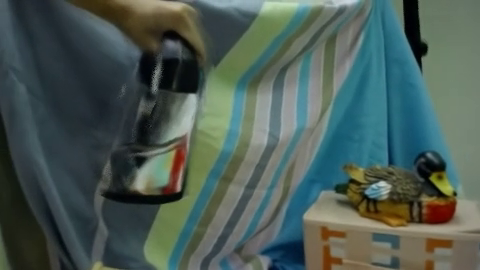} &
        \includegraphics[width=0.155\textwidth]{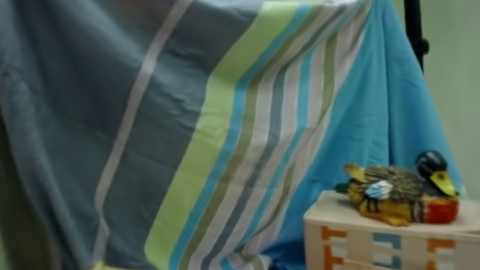}
        \\

        \raisebox{20pt}{\rotatebox[origin=c]{90}{sieve}}&
         \includegraphics[width=0.155\textwidth]{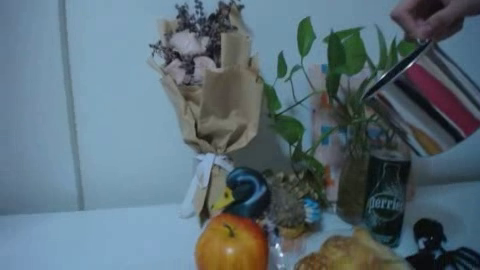} &
        \includegraphics[width=0.155\textwidth]{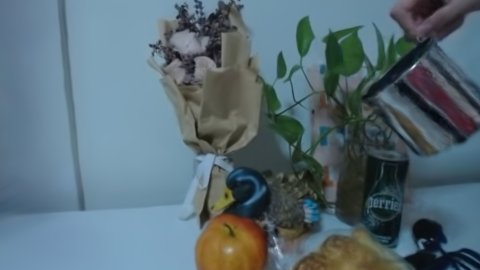} & 
        \includegraphics[width=0.155\textwidth]{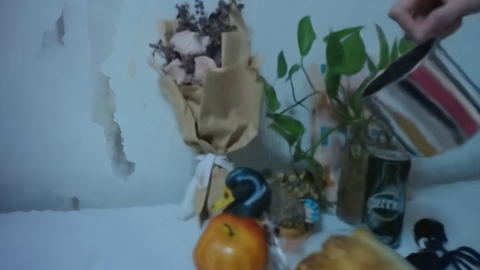} &
        \includegraphics[width=0.155\textwidth]{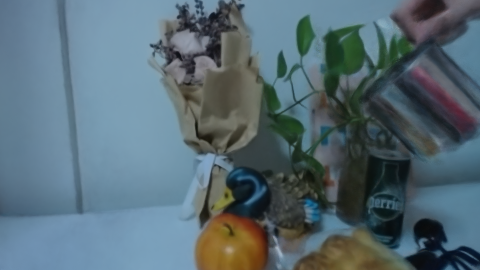} &
        \includegraphics[width=0.155\textwidth]{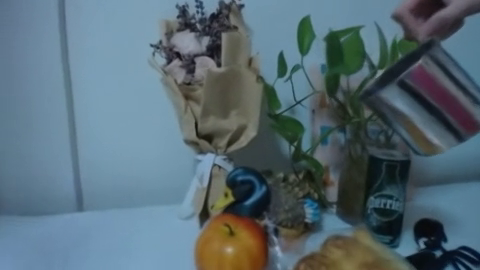} &
        \includegraphics[width=0.155\textwidth]{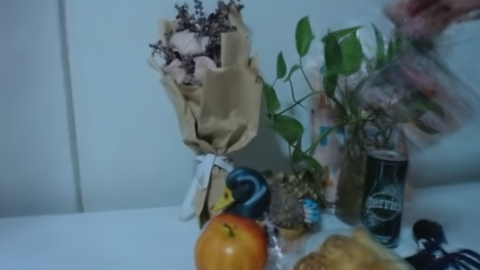}
        \\
            
        \end{tabular}
    }
\vspace{-5pt}
	\caption{\textbf{Qualitative comparisons of baselines and our method on NeRF-DS real-world dataset.} Experimental results indicate that our method can achieve superior rendering quality on real-world datasets where the pose is not absolutely precise.
}\label{fig:nerfds-quality}
\end{figure*}
\section{Experiment} \label{sec:exp}
\par In this section, we present the experimental evaluation of our method. To give proof of effectiveness, we evaluate our approach on the benchmark which consists of the synthetic dataset from D-NeRF \cite{pumarola2021d} and real-world datasets sourced from HyperNeRF \cite{park2021hypernerf} and NeRF-DS \cite{yan2023nerf}. The division on training and testing part, as well as the image resolution, aligns perfectly with the original paper. 


\subsection{Implementation Details}
\par We implement our framework using PyTorch \cite{paszke2019pytorch} and modify the differentiable Gaussian rasterization by incorporating depth visualization. For training, we conducted training for a total of 40k iterations. During the initial 3k iterations, we solely trained the 3D Gaussians to attain relatively stable positions and shapes. Subsequently, we jointly train the 3D Gaussians and the deformation field. For optimization, a single Adam optimizer \cite{KingBa15} is used but with a different learning rate for each component: the learning rate of 3D Gaussians is exactly the same as the official implementation, while the learning rate of the deformation network undergoes exponential decay, ranging from 8e-4 to 1.6e-6. Adam's $\beta$ value range is set to (0.9, 0.999). Experiments with synthetic datasets were all conducted against a black background and at a full resolution of 800x800.
All the experiments were done on an NVIDIA RTX 3090. 

\subsection{Results and Comparisons} \label{sec:comparisons}
\paragraph{Comparisons on synthetic dataset.} In our experiments, we benchmarked our method against several baselines using the monocular synthetic dataset introduced by D-NeRF \cite{pumarola2021d}. The quantitative evaluation, presented in Tab.~\ref{exp:dnerf}, offers compelling evidence of the superior performance of our approach over the current state-of-the-art. Notably, metrics pertinent to structural consistency, such as LPIPS and SSIM, demonstrate our method's pronounced superiority.

\par For a more visual assessment, we provide qualitative results in Fig. \ref{fig:syn-quality}. These visual comparisons underscore the capability of our method in delivering high-fidelity dynamic scene modeling. It's evident from the results that our approach ensures enhanced consistency and captures intricate rendering details in novel-view renderings.

\paragraph{Comparisons on real-world dataset.} 
\par We compare our method with the baselines using the monocular real-world dataset from NeRF-DS \cite{yan2023nerf} and HyperNeRF \cite{park2021hypernerf}. It should be noted that the camera poses for some of the HyperNeRF datasets are very inaccurate. Given that metrics like PSNR, designed to assess image rendering quality, are inclined to penalize slight deviations more than blurring, we have refrained from incorporating HyperNeRF in our quantitative analysis. For a qualitative analysis of HyperNeRF, please refer to the supplementary materials. The quantitative and qualitative evaluations for the NeRF-DS dataset are detailed in Tab. \ref{exp:nerfds-quant} and Fig. \ref{fig:nerfds-quality}, respectively. These results attest to the robustness of our method when applied to real-world scenes, even when the associated poses are not perfectly accurate.

\paragraph{Rendering Efficiency.} \label{sec:real-time}
\par The rendering speed is correlated with the quantity of 3D Gaussians. Overall, when the number of 3D Gaussians is below 250k, our method can achieve real-time rendering over 30 FPS on an NVIDIA RTX 3090. Detailed results can be found in the supplementary material.


\paragraph{Depth Visualization.}
\par
\begin{figure}
    \centering
    \footnotesize{
        \setlength{\tabcolsep}{1pt} 
        \begin{tabular}{cccc}
        \includegraphics[width=0.11\textwidth]{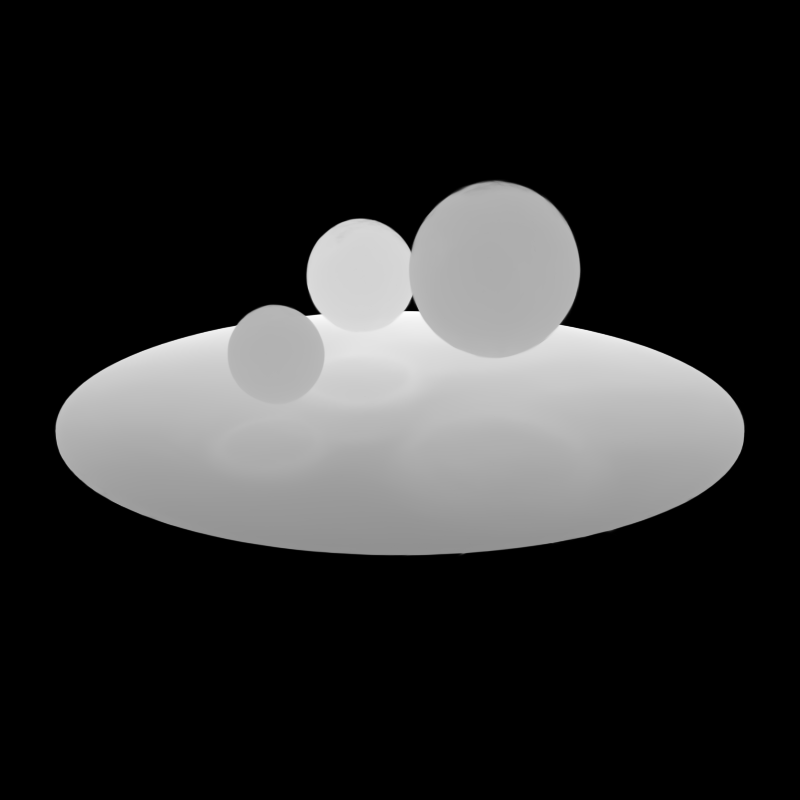} &
        \includegraphics[width=0.11\textwidth]{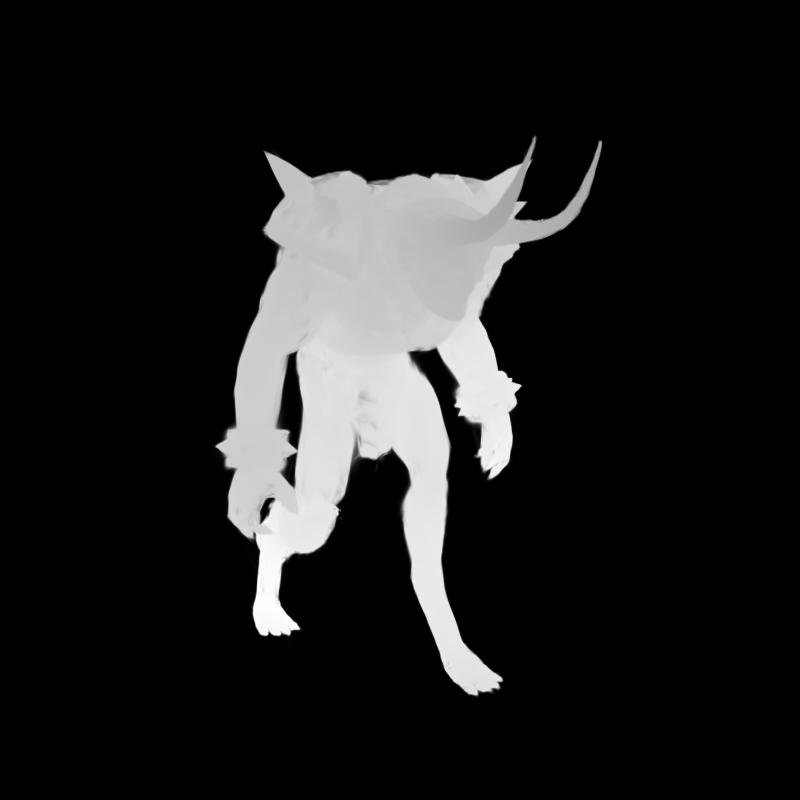} &
        \includegraphics[width=0.11\textwidth]{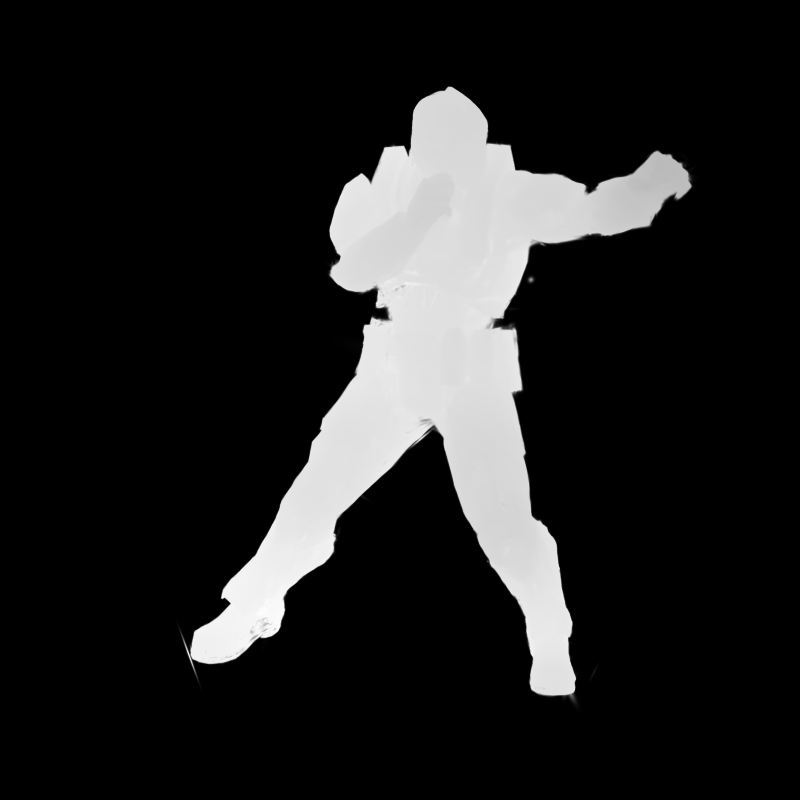} &
        \includegraphics[width=0.11\textwidth]{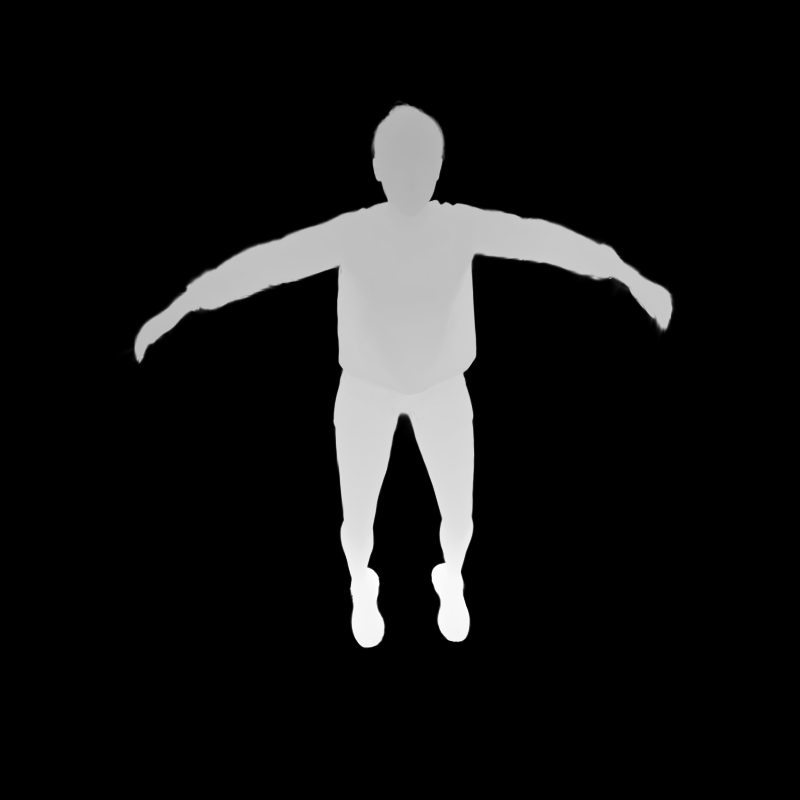} \\
        \includegraphics[width=0.11\textwidth]{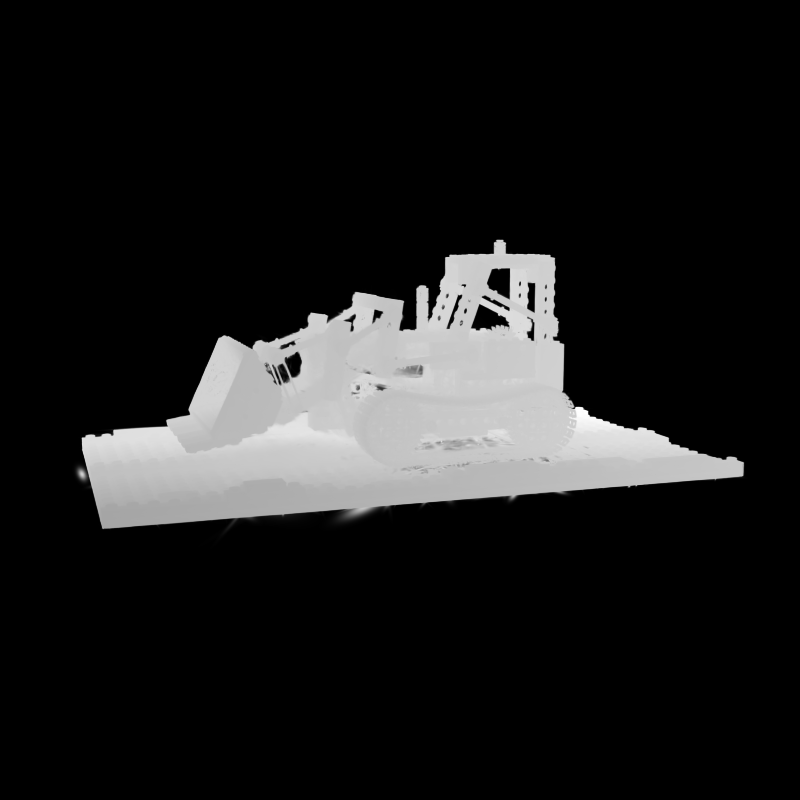} &
        \includegraphics[width=0.11\textwidth]{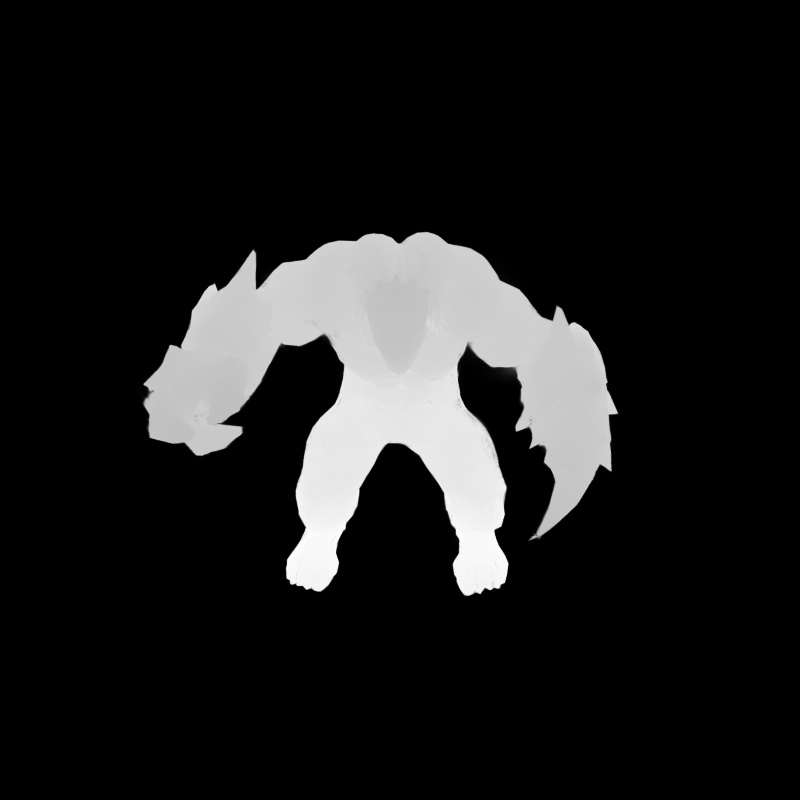} &
        \includegraphics[width=0.11\textwidth]{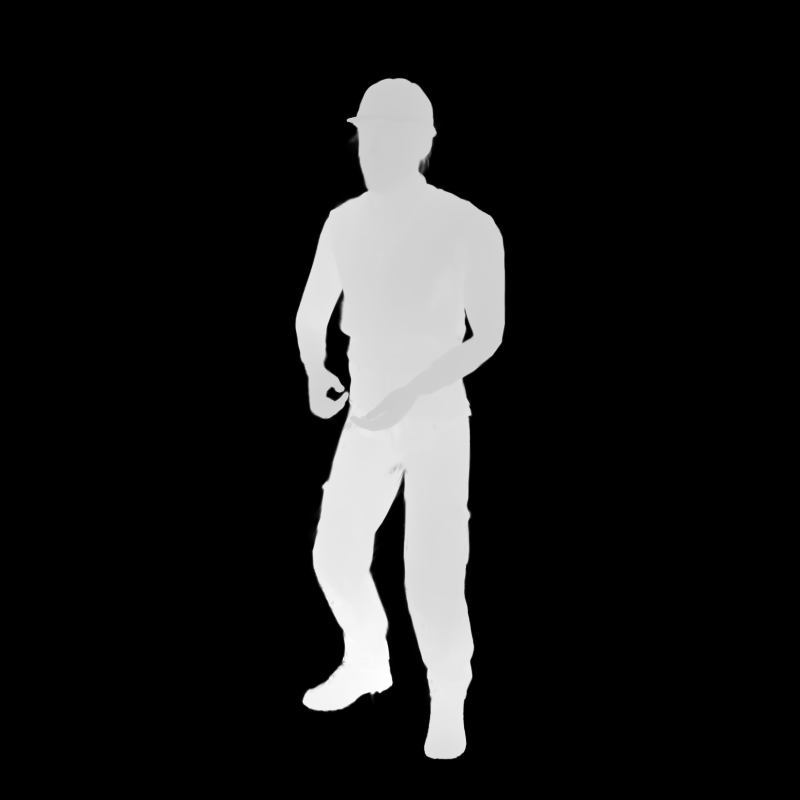} &
        \includegraphics[width=0.11\textwidth]{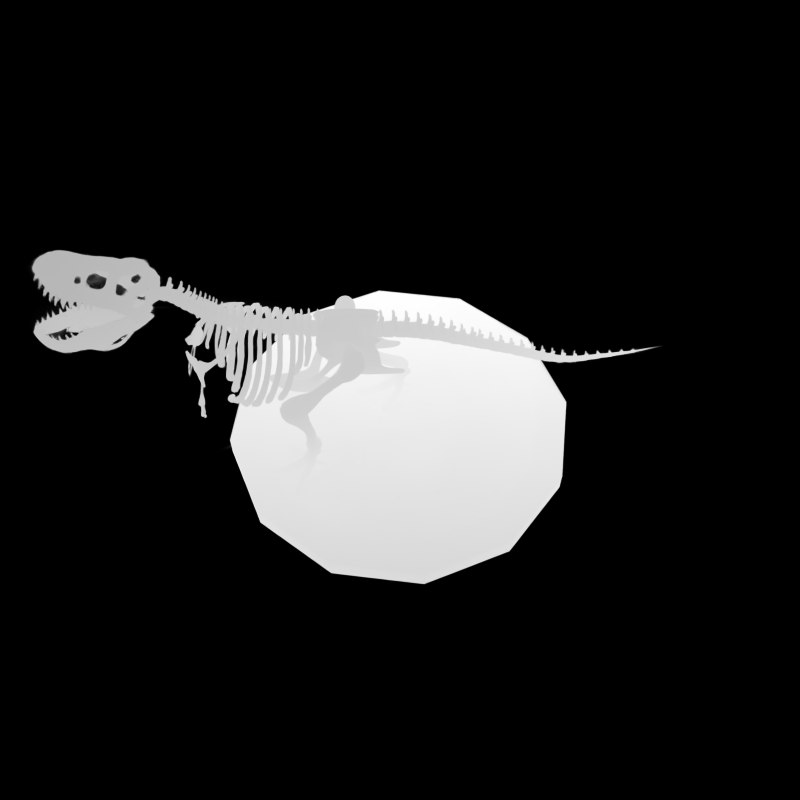} \\
        \end{tabular}
    }
    \caption{\textbf{Depth Visualization.} We visualized the depth map of the D-NeRF dataset. The first row includes bouncing-balls, hell-warrior, hook, and jumping-jacks, while the second row includes lego, mutant, standup, and trex.
    }\label{fig:depth}
\end{figure}

 We visualized the depth of synthetic dataset scenes in Fig. \ref{fig:depth} to demonstrate that our deformation network is well fitted to produce temporal transformation rather than relying on color-based hard-coding. The precise depth underscores the accuracy of our geometric reconstruction, proving highly advantageous for the novel-view synthesis task.

\subsection{Ablation Study}
\paragraph{Annealing smooth training. } 
As illustrated in Fig. \ref{fig:ablation} and Tab.~\ref{exp:nerfds-quant}, this mechanism fosters improved convergence towards intricate regions, effectively mitigating the overfitting tendencies in real-world datasets. Furthermore, it is unequivocally clear from our observations that this strategy significantly bolsters the temporal smoothness of the deformation field. See more ablations in supplementary materials.


\section{Limitations}
\par 
Through our experimental evaluations, we observed that the convergence of 3D Gaussians is profoundly influenced by the diversity of perspectives. As a result, datasets characterized by sparse viewpoints and limited viewpoint coverage may lead our method to encounter overfitting challenges. Additionally, the efficacy of our approach is contingent upon the accuracy of pose estimations. This dependency was evident when our method did not achieve optimal PSNR values on the Nerfies/HyperNeRF dataset, attributable to deviations in pose estimation via COLMAP. Furthermore, the temporal complexity of our approach is directly proportional to the quantity of 3D Gaussians. In scenarios with an extensive array of 3D Gaussians, there is a potential escalation in both training duration and memory consumption. Lastly, our evaluations have predominantly revolved around scenes with moderate motion dynamics. The method's adeptness at handling intricate human motions, such as nuanced facial expressions, remains an open question. We perceive these constraints as promising directions for subsequent research endeavors.

\begin{table}[]
\centering
\begin{tabular}{l||lll}
\hline
    & PSNR $\uparrow$ & SSIM $\uparrow$ & LPIPS $\downarrow$ \\ \hline
    3D-GS      & 20.29 & 0.7816 & 0.2920 \\
    TiNeuVox   & 21.61 & 0.8234 & 0.2766 \\
    HyperNeRF  & 23.45 & \cellcolor{yzythird}0.8488 & \cellcolor{yzythird}0.1990 \\ 
    NeRF-DS    & \cellcolor{yzythird}23.60 & \cellcolor{yzysecond}0.8494 & \cellcolor{yzysecond}0.1816 \\ \hline
    Ours (w/o AST) & \cellcolor{yzysecond}23.97 & 0.8346 & 0.2037 \\ 
    Ours       & \cellcolor{yzybest}24.11 & \cellcolor{yzybest}0.8525 & \cellcolor{yzybest}0.1769 \\ \hline
\end{tabular}
\caption{\textbf{Metrics on NeRF-DS dataset}. We computed the mean of the metrics across all seven scenes. Cells are highlighted as follows: \colorbox{yzybest}{best}, \colorbox{yzysecond}{second best}, and \colorbox{yzythird}{third best}. For individual metrics about each scene, please refer to the supplementary material.}
\label{exp:nerfds-quant}
\end{table}

\section{Conclusions}
\par 
\par We introduce a novel deformable 3D Gaussian splatting method, specifically designed for monocular dynamic scene modeling, which surpasses existing methods in both quality and speed. By learning the 3D Gaussians in canonical space, we enhance the versatility of the 3D-GS differentiable rendering pipeline for dynamically captured monocular scenes. It's crucial to recognize that point-based methods, in comparison to implicit representations, are more editable and better suited for post-production tasks. Additionally, our method incorporates an annealing smooth training strategy, aimed at reducing overfitting associated with time encoding while maintaining intricate scene details, without adding any extra training overhead. Experimental results demonstrate that our method not only achieves superior rendering effects but is also capable of real-time rendering.
{\small
\bibliographystyle{ieee_fullname}
\bibliography{ref}
}
\appendix


\section{Overview}
The appendix provides some implementation details and further results that accompany the paper. 

\begin{itemize}
    \item Section~\ref{sec:detail} introduces the implementation details of the network architecture in our approach.
    \item Section~\ref{sec: add-results} provides additional results, including more visualizations, rendering efficiency, more comparisons, and more ablations.
    \item Section~\ref{sec:failure} discusses the failure cases of our method.

\end{itemize}

\section{Implementation Details} \label{sec:detail}
\subsection{Network Architecture of the Deformation Field}
\par
\begin{figure}[ht] 
  \centering
  \includegraphics[width=0.4\textwidth]{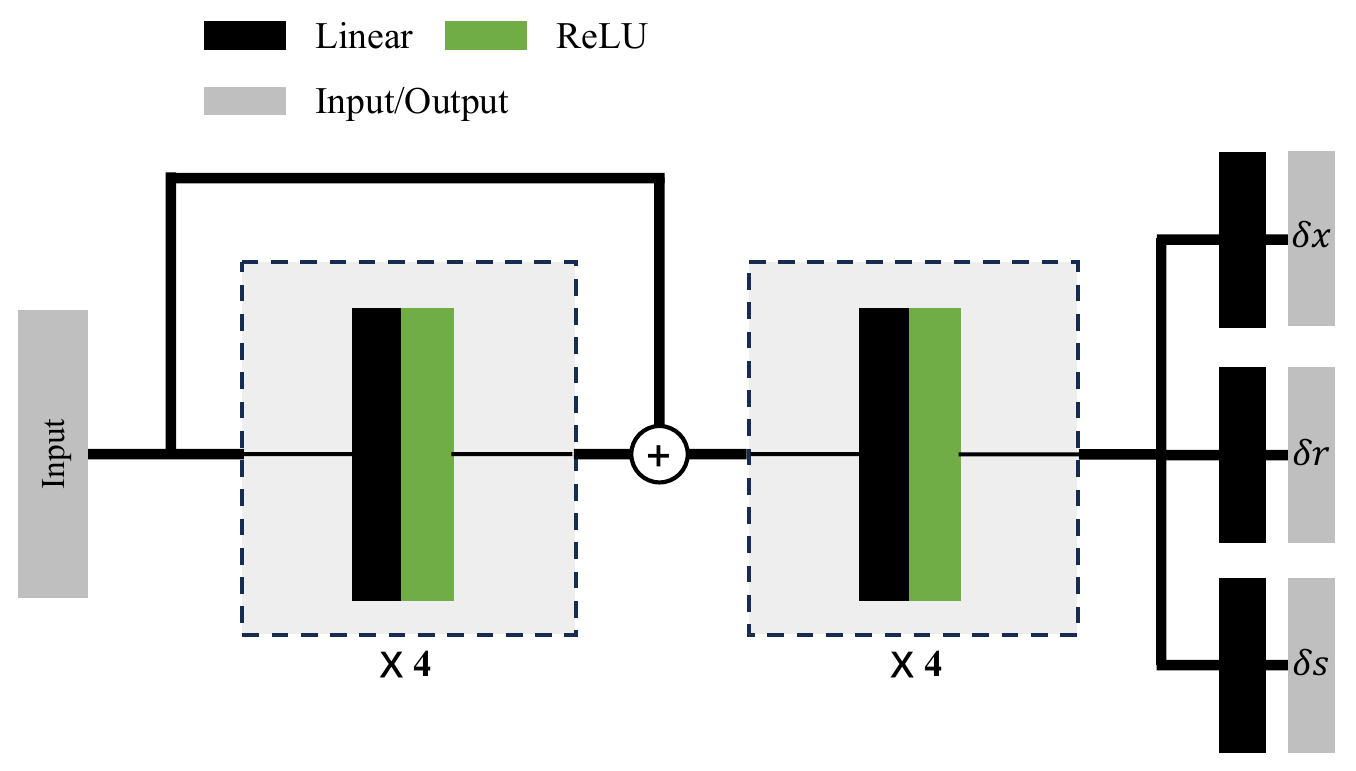}
  \caption{\textbf{The architecture of our deformation MLP.} 
  } \label{fig:network-arch}
\end{figure}

 We learn the deformation field with an MLP network $\mathcal{F_{\theta}}:(\gamma(\mathbf{x}), \gamma(t)) \rightarrow(\delta x, \delta r, \delta s)$, which maps from each coordinate of 3D Gaussians and time to their corresponding deviations in position, rotation, and scaling. The weights $\theta$ of the MLP are optimized through this mapping. As shown in Fig. \ref{fig:network-arch}, our MLP $\mathcal{F_{\theta}}$ initially processes the input through eight fully connected layers that employ ReLU activations and feature 256-dimensional hidden layers, and outputs a 256-dimensional feature vector. This vector is subsequently passed through three additional fully connected layers (\textbf{without activation}) to separately output the offsets over time for position, rotation, and scaling. It should be noted that similar to NeRF, we concatenate the feature vector and the input in the fourth layer. Due to the compact structure of MLP, our additional storage compared to 3D Gaussians is only 2MB.

\begin{figure}
    \centering
    \addtolength{\tabcolsep}{-6.5pt}
    \footnotesize{
        \setlength{\tabcolsep}{1pt} 
        \begin{tabular}{p{8.2pt}cccc}
        \raisebox{40pt}{\rotatebox[origin=c]{90}{broom}}&
             \includegraphics[width=0.1\textwidth]{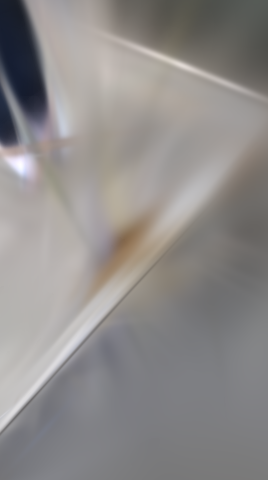} &
            \includegraphics[width=0.1\textwidth]{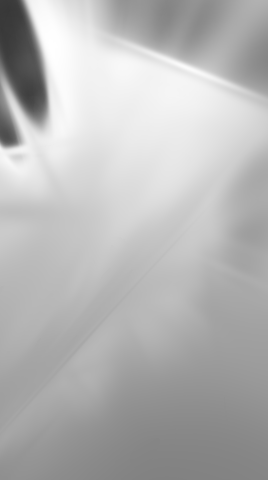} &
            \includegraphics[width=0.1\textwidth]{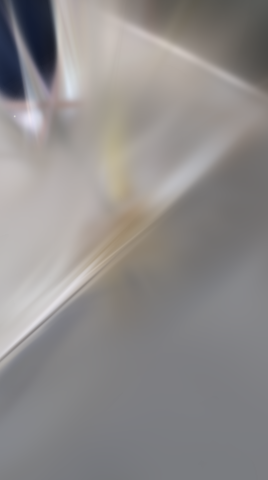} &
            \includegraphics[width=0.1\textwidth]{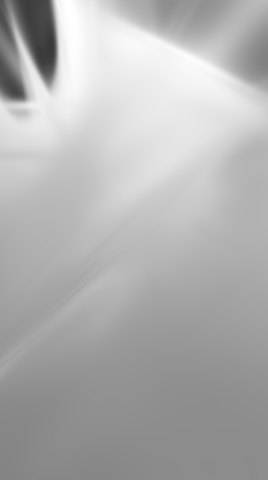} \\

        & train & train-depth & test & test-depth
        
        \end{tabular}
    }
\vspace{-5pt}
	\caption{\textbf{Failure case on inaccurate pose.} Excessively inaccurate poses can lead to the failure of the convergence on the training set.}\label{fig:failure-pose}
\end{figure}

\begin {table*}[ht]
\centering
\resizebox{\textwidth}{!}{
\begin{tabular}{ccccccccccccccccc}
\hline \multirow[b]{2}{*}{ Method } & \multicolumn{3}{c}{ Sieve } & \multicolumn{3}{c}{ Plate } & \multicolumn{3}{c}{ Bell } & \multicolumn{3}{c}{ Press } \\
& PSNR$\uparrow$ & SSIM$\uparrow$ & LPIPS$\downarrow$ & PSNR$\uparrow$ & SSIM$\uparrow$ & LPIPS$\downarrow$ & PSNR$\uparrow$ & SSIM$\uparrow$ & LPIPS$\downarrow$ & PSNR$\uparrow$ & SSIM $\uparrow$ & LPIPS$\downarrow$ \\

\hline 3D-GS & 23.16 & 0.8203 & 0.2247 & 16.14 & 0.6970 & 0.4093 & 21.01 & 0.7885 & 0.2503 & 22.89 & 0.8163 & 0.2904 \\

TiNeuVox & 21.49 & 0.8265 & 0.3176 & \cellcolor{yzybest}20.58 & \cellcolor{yzythird}0.8027 & 0.3317 & 23.08 & \cellcolor{yzythird}0.8242 & 0.2568 & 24.47 & 0.8613 & 0.3001 \\

HyperNeRF & \cellcolor{yzythird}25.43 & \cellcolor{yzysecond}0.8798 & 0.1645 & 18.93 & 0.7709 & \cellcolor{yzythird}0.2940 & 23.06 & 0.8097 & 0.2052 & \cellcolor{yzybest}26.15 & \cellcolor{yzybest}0.8897 & \cellcolor{yzythird}0.1959 \\

NeRF-DS & \cellcolor{yzybest}25.78 & \cellcolor{yzybest}0.8900 & \cellcolor{yzybest}0.1472 & \cellcolor{yzysecond}20.54 & \cellcolor{yzysecond}0.8042 & \cellcolor{yzybest}0.1996 & \cellcolor{yzythird}23.19 & 0.8212 & \cellcolor{yzythird}0.1867 & 25.72 & \cellcolor{yzythird}0.8618 & 0.2047 \\

Ours (w/o AST) & 25.33 & 0.8620 & \cellcolor{yzythird}0.1594 & 20.32 & 0.7173 & 0.3914 & \cellcolor{yzysecond}25.62 & \cellcolor{yzysecond}0.8498 & \cellcolor{yzysecond}0.1540 & \cellcolor{yzythird}25.78 & 0.8613 & \cellcolor{yzysecond}0.1919 \\

Ours & \cellcolor{yzysecond}25.70 & \cellcolor{yzythird}0.8715 & \cellcolor{yzysecond}0.1504 & \cellcolor{yzythird}20.48 & \cellcolor{yzybest}0.8124 & \cellcolor{yzysecond}0.2224 & \cellcolor{yzybest}25.74 & \cellcolor{yzybest}0.8503 & \cellcolor{yzybest}0.1537 & \cellcolor{yzysecond}26.01 & \cellcolor{yzysecond}0.8646 & \cellcolor{yzybest}0.1905 \\

\hline & \multicolumn{3}{c}{ Cup } & \multicolumn{3}{c}{ As } & \multicolumn{3}{c}{ Basin } & \multicolumn{3}{c}{ Mean } \\

Method & PSNR$\uparrow$ & SSIM$\uparrow$ & LPIPS$\downarrow$ & PSNR$\uparrow$ & SSIM$\uparrow$ & LPIPS$\downarrow$ & PSNR$\uparrow$ & SSIM$\uparrow$ & LPIPS$\downarrow$ & PSNR$\uparrow$ & SSIM $\uparrow$ & LPIPS$\downarrow$ \\

\hline 3D-GS & 21.71 & 0.8304 & 0.2548 & 22.69 & 0.8017 & 0.2994 & 18.42 & 0.7170 & 0.3153 & 20.29 & 0.7816 & 0.2920 \\

TiNeuVox & 19.71 & 0.8109 & 0.3643 & 21.26 & 0.8289 & 0.3967 & \cellcolor{yzybest}20.66 & \cellcolor{yzythird}0.8145 & 0.2690 & 21.61 & 0.8234 & 0.2766 \\

HyperNeRF & 24.59 & \cellcolor{yzythird}0.8770 & \cellcolor{yzythird}0.1650 & \cellcolor{yzythird}25.58 & \cellcolor{yzybest}0.8949 & \cellcolor{yzysecond}0.1777 & \cellcolor{yzysecond}20.41 & \cellcolor{yzybest}0.8199 & 0.1911 & 23.45 & \cellcolor{yzythird}0.8488 & \cellcolor{yzythird}0.1990 \\

NeRF-DS & \cellcolor{yzybest}24.91 & 0.8741 & 0.1737 & 25.13 & 0.8778 & \cellcolor{yzybest}0.1741 & \cellcolor{yzythird}19.96 & \cellcolor{yzysecond}0.8166 & \cellcolor{yzybest}0.1855 & \cellcolor{yzythird}23.60 & \cellcolor{yzysecond}0.8494 & \cellcolor{yzysecond}0.1816 \\

Ours (w/o AST) & \cellcolor{yzythird}24.80 & \cellcolor{yzysecond}0.8848 & \cellcolor{yzysecond}0.1571 & \cellcolor{yzysecond}26.29 & \cellcolor{yzythird}0.8800 & 0.1830 & 19.68 & 0.7869 & \cellcolor{yzysecond}0.1888 & \cellcolor{yzysecond}23.97 & 0.8346 & 0.2037 \\

Ours & \cellcolor{yzysecond}24.86 & \cellcolor{yzybest}0.8908 & \cellcolor{yzybest}0.1532 & \cellcolor{yzybest}26.31 & \cellcolor{yzysecond}0.8842 & \cellcolor{yzythird}0.1783 & 19.67 & 0.7934 & \cellcolor{yzythird}0.1901 & \cellcolor{yzybest}24.11 & \cellcolor{yzybest}0.8524 & \cellcolor{yzybest}0.1769 \\

\hline
\end{tabular}}
\vspace{-5pt}
\caption{\textbf{Quantitative comparison on NeRF-DS dataset per-scene}. We color each cell as \colorbox{yzybest}{best}, \colorbox{yzysecond}{second best}, and \colorbox{yzythird}{third best}. Our method, overall, achieves the best rendering quality and robust convergence in the majority of scenes. It is worth noting that the metrics we used are the same as those in the main text, with LPIPS using the VGG network. The slight differences in our measurement metrics, compared to those used in NeRF-DS and HyperNeRF, are due to their papers employing MS-SSIM and LPIPS with the AlexNet.}
\label{tab: nerfds-per}
\end{table*}

\begin{table*}[ht]
\centering
\begin{tabular}{@{}lcc|lcc|lcc@{}}
\toprule
\multicolumn{3}{c|}{\textbf{D-NeRF Dataset}} & \multicolumn{3}{c|}{\textbf{NeRF-DS Dataset}} & \multicolumn{3}{c}{\textbf{HyperNeRF Dataset}} \\ \midrule
\textbf{Scene} & \textbf{FPS} & \textbf{Num (k)} & \textbf{Scene} & \textbf{FPS} & \textbf{Num (k)} & \textbf{Scene} & \textbf{FPS} & \textbf{Num (k)} \\
\midrule
Lego       & 24   & 300   & AS         & 48   & 185   & Espresso    & 15   & 620   \\
Jump       & 85   & 90    & Basin      & 29   & 250   & Americano   & 6    & 1300  \\
Bouncing   & 38   & 170   & Bell       & 18   & 400   & Cookie      & 9    & 1080  \\
T-Rex      & 30   & 220   & Cup        & 35   & 200   & Chicken & 10  & 740   \\
Mutant     & 40   & 170   & Plate      & 31   & 230   & Torchocolate & 8   & 1030  \\
Warrior    & 172  & 40    & Press      & 48   & 185   & Lemon       & 23   & 420   \\
Standup    & 93   & 80    & Sieve      & 35   & 200   & Hand        & 6    & 1750  \\
Hook       & 45   & 150   &            &      &       & Printer     & 12   & 650   \\
\bottomrule
\end{tabular}
\caption{\textbf{Experiments on FPS with respect to the number of 3D Gaussians.} The results of the experiments demonstrate that our method is capable of real-time rendering on a 3090 GPU when the number of point clouds is less than 250k. The excessively high number of 3D Gaussians in HyperNeRF reflects the critical importance of the camera pose accuracy for the convergence of our method.}
\label{tab:fps}
\end{table*}

\begin{figure*}
    \centering
    \addtolength{\tabcolsep}{-6.5pt}
    \footnotesize{
        \setlength{\tabcolsep}{1pt} 
        \begin{tabular}{p{8.2pt}ccccccc}
            & GT & Ours & Tineuvox & HyperNeRF & 3D-GS  \\
        \raisebox{35pt}{\rotatebox[origin=c]{90}{torchocolate}}&
         \includegraphics[width=0.18\textwidth]{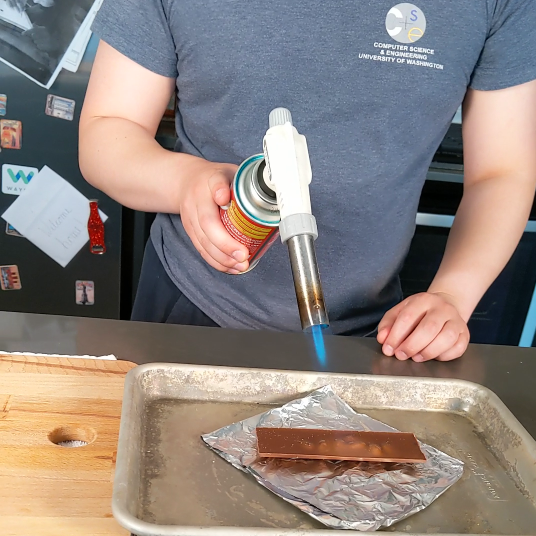} &
        \includegraphics[width=0.18\textwidth]{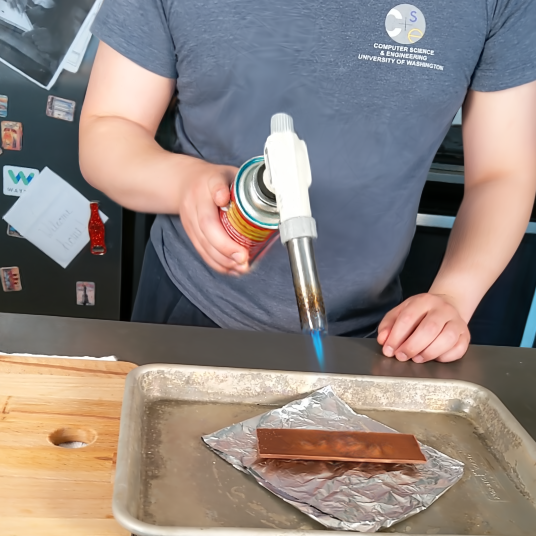} &
        \includegraphics[width=0.18\textwidth]{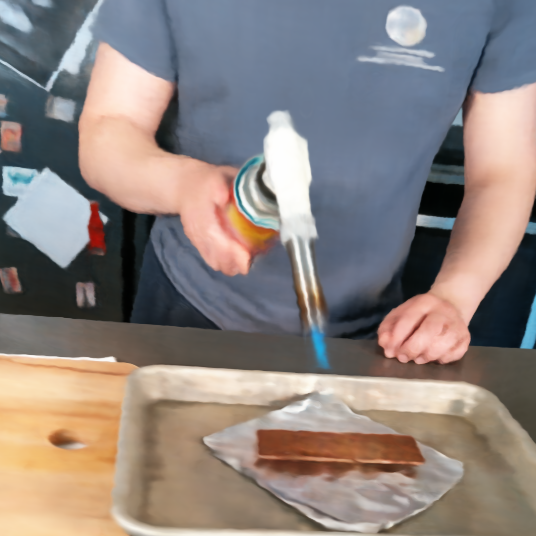} &
        \includegraphics[width=0.18\textwidth]{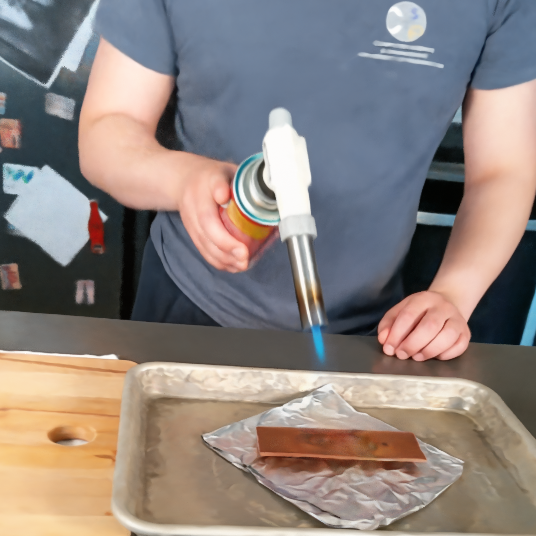} &
        \includegraphics[width=0.18\textwidth]{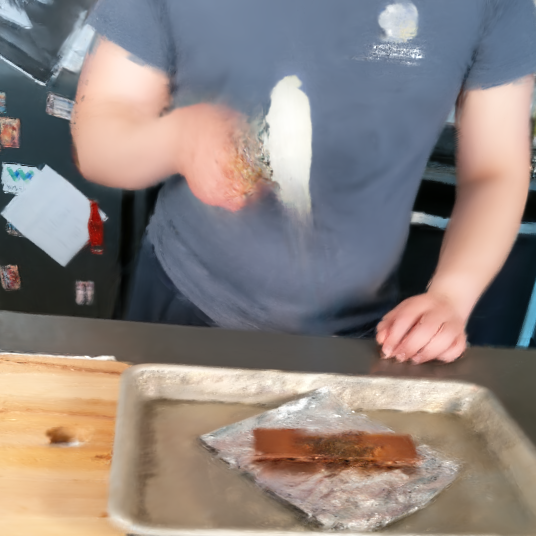}
         \\

        \raisebox{35pt}{\rotatebox[origin=c]{90}{hand}}&
         \includegraphics[width=0.18\textwidth]{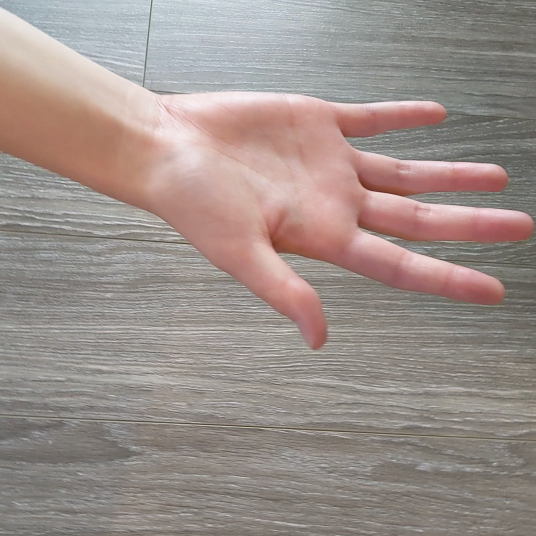} &
        \includegraphics[width=0.18\textwidth]{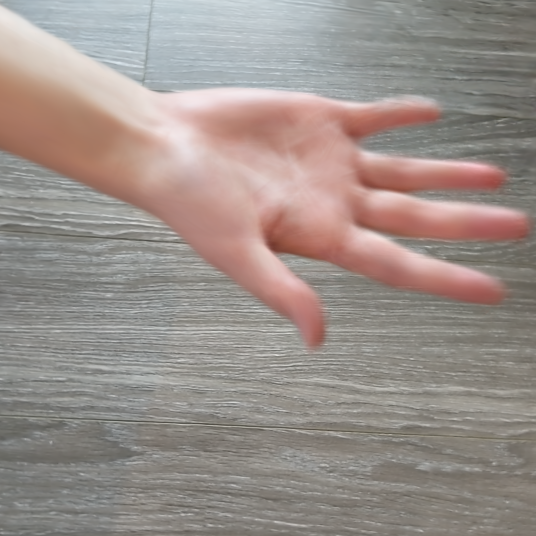} & 
        \includegraphics[width=0.18\textwidth]{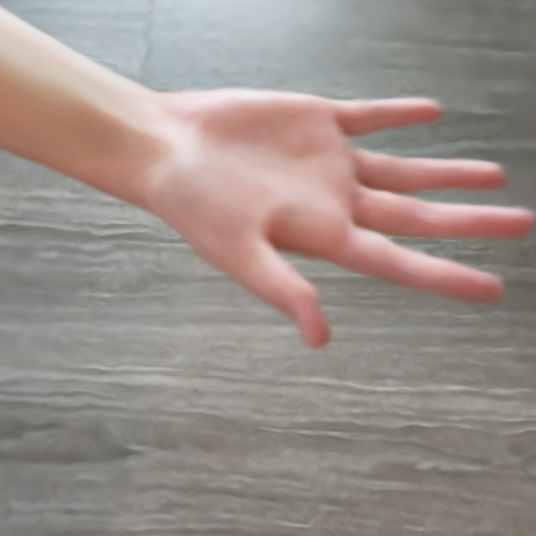} &
        \includegraphics[width=0.18\textwidth]{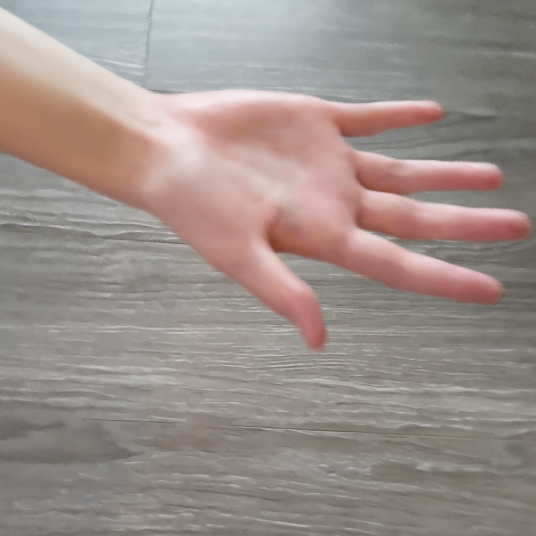} &
        \includegraphics[width=0.18\textwidth]{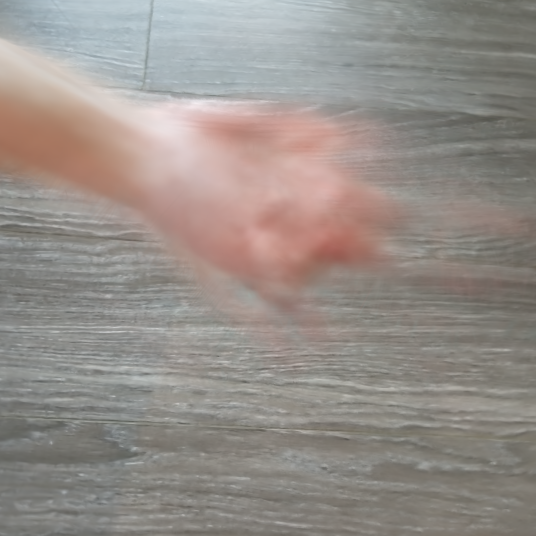}
        \\
        
        \raisebox{10pt}{\rotatebox[origin=c]{90}{lemon}}&
         \includegraphics[width=0.18\textwidth]{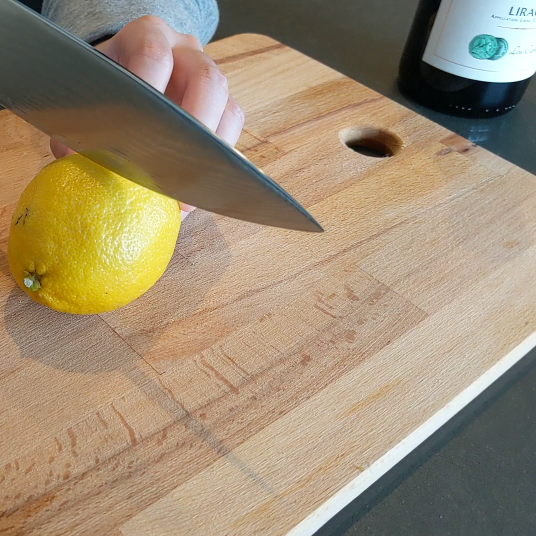} &
        \includegraphics[width=0.18\textwidth]{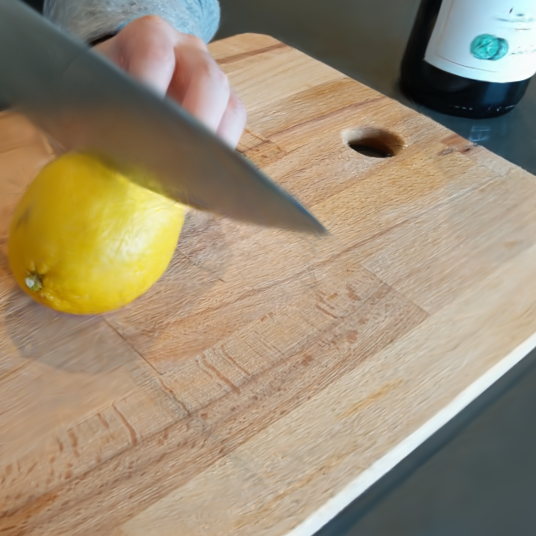} &
        \includegraphics[width=0.18\textwidth]{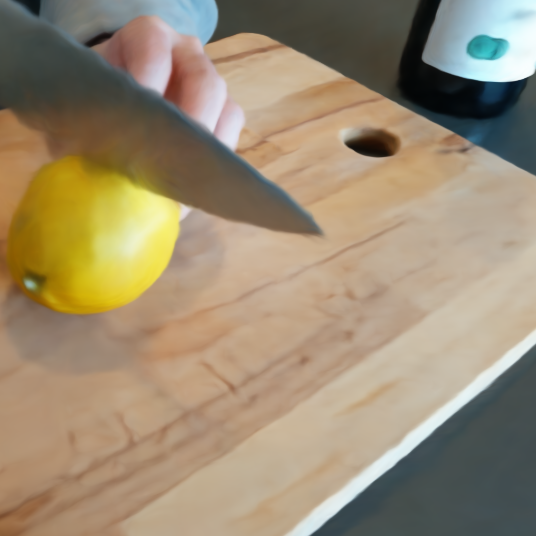} &
        \includegraphics[width=0.18\textwidth]{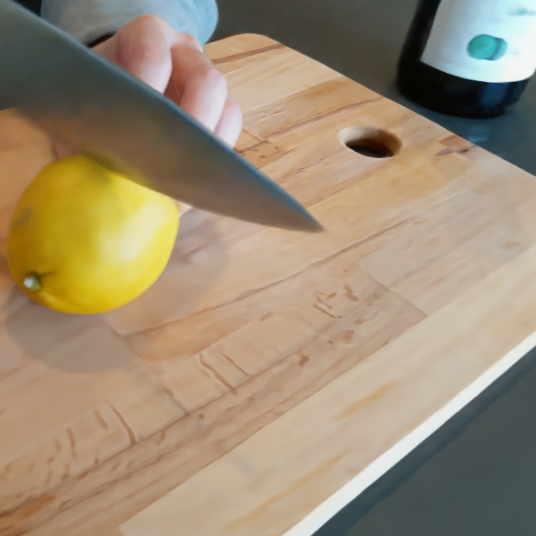} &
        \includegraphics[width=0.18\textwidth]{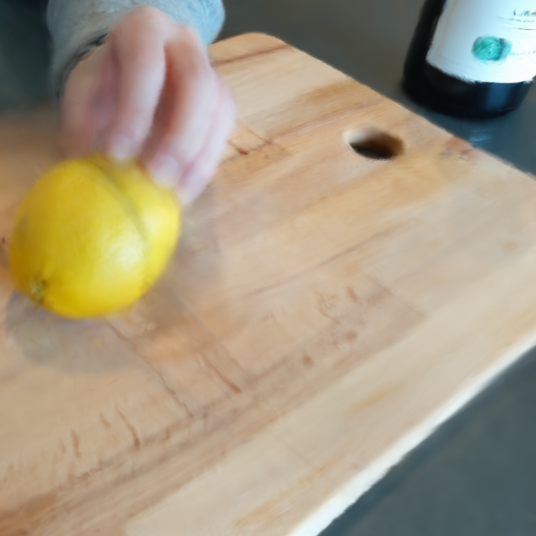}
        \\

        \raisebox{35pt}{\rotatebox[origin=c]{90}{printer}}&
         \includegraphics[width=0.18\textwidth]{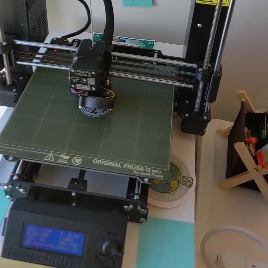} &
        \includegraphics[width=0.18\textwidth]{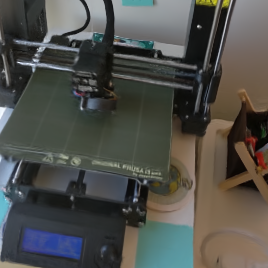} &
        \includegraphics[width=0.18\textwidth]{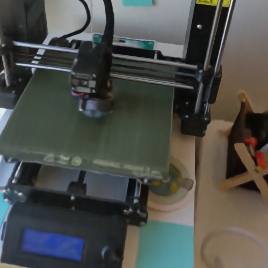} &
        \includegraphics[width=0.18\textwidth]{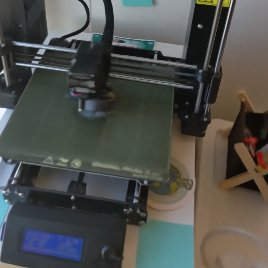} &
        \includegraphics[width=0.18\textwidth]{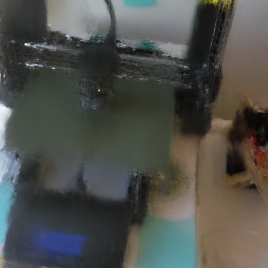}
        \\

        \raisebox{35pt}{\rotatebox[origin=c]{90}{chicken}}&
         \includegraphics[width=0.18\textwidth]{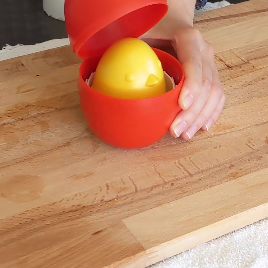} &
        \includegraphics[width=0.18\textwidth]{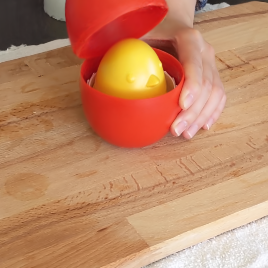} &
        \includegraphics[width=0.18\textwidth]{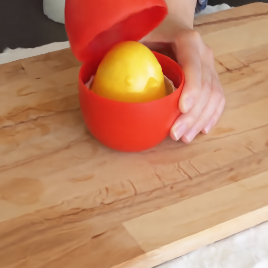} &
        \includegraphics[width=0.18\textwidth]{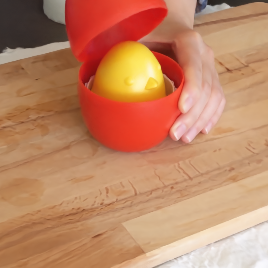} &
        \includegraphics[width=0.18\textwidth]{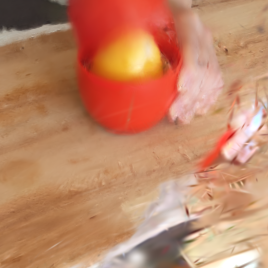}
        \\
            
        \end{tabular}
    }
\vspace{-5pt}
	\caption{\textbf{Qualitative comparisons of baselines and our method on HyperNeRF dataset.} The first three rows present the results of the time interpolation task, while the last two rows depict the outcomes of the novel viewpoint synthesis task. Experimental results indicate that our method can achieve superior rendering quality on real datasets where the pose is not absolutely precise.
}\label{fig:hyper-quality}
\end{figure*}

\begin{figure*}
    \centering
    \addtolength{\tabcolsep}{-6.5pt}
    \footnotesize{
        \setlength{\tabcolsep}{1pt} 
        \begin{tabular}{p{8.2pt}cccccc}
        \raisebox{33pt}{\rotatebox[origin=c]{90}{kuka}}&
             \includegraphics[width=0.145\textwidth]{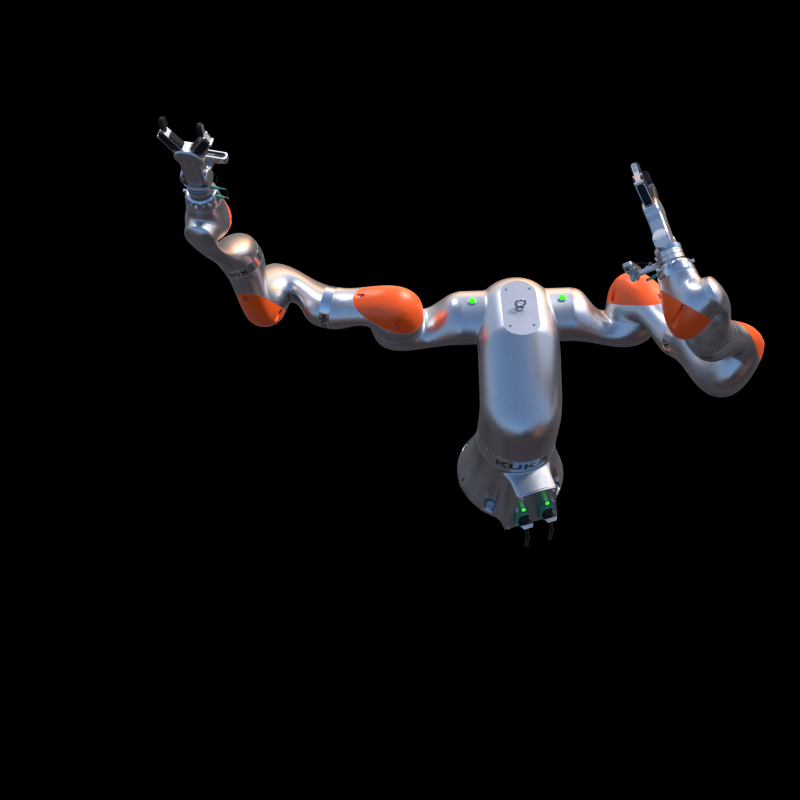} &
            \includegraphics[width=0.145\textwidth]{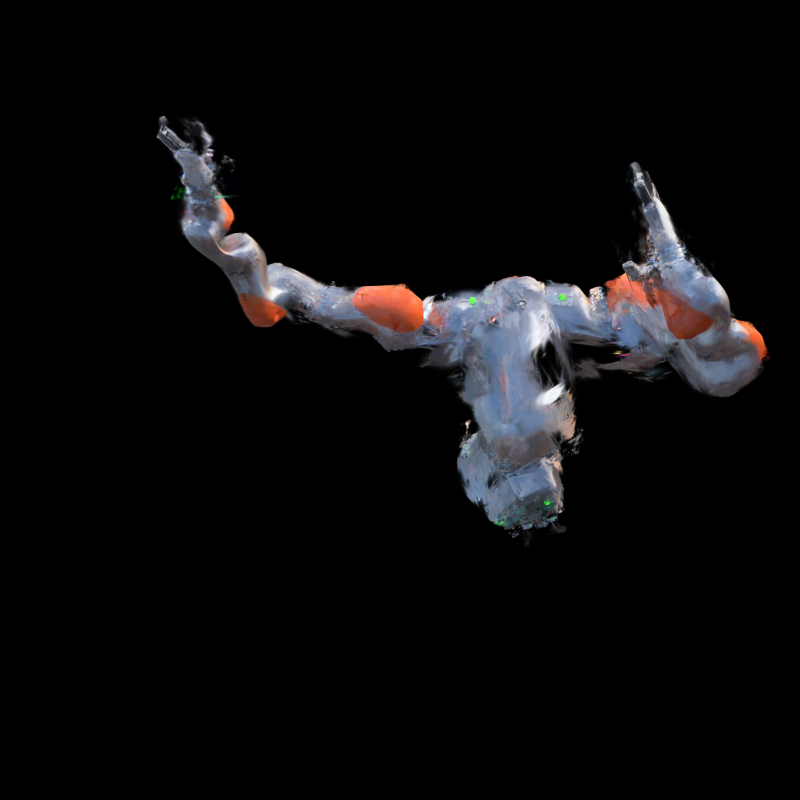} &
            \includegraphics[width=0.145\textwidth]{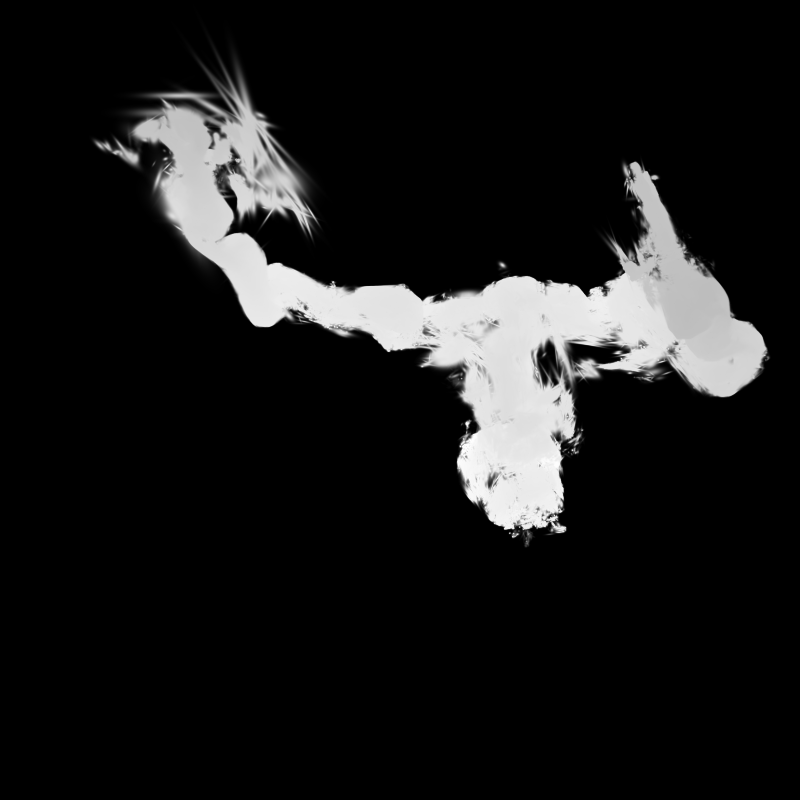} &
            \includegraphics[width=0.145\textwidth]{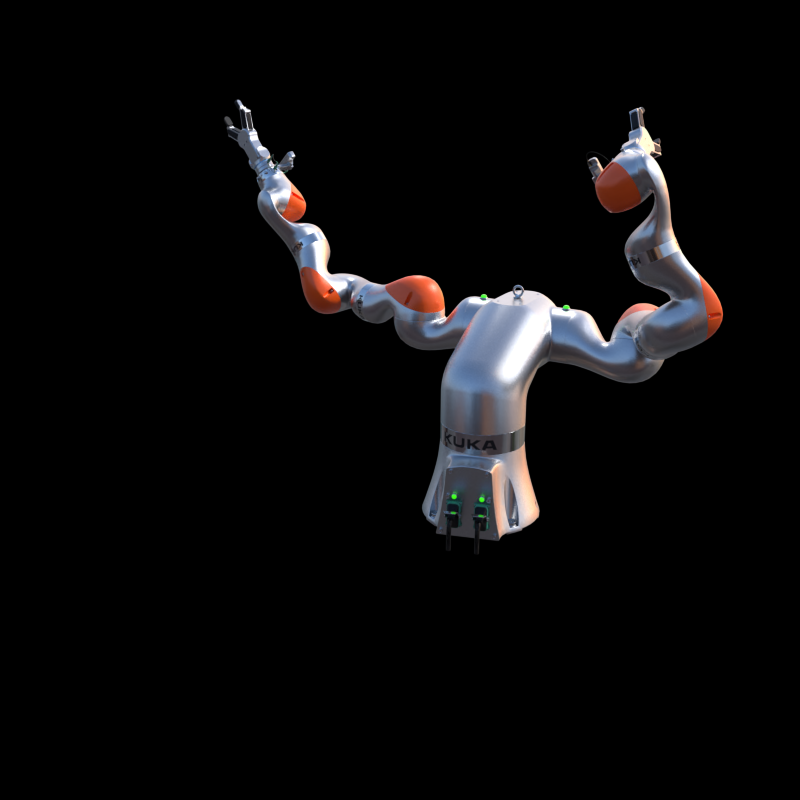} &
            \includegraphics[width=0.145\textwidth]{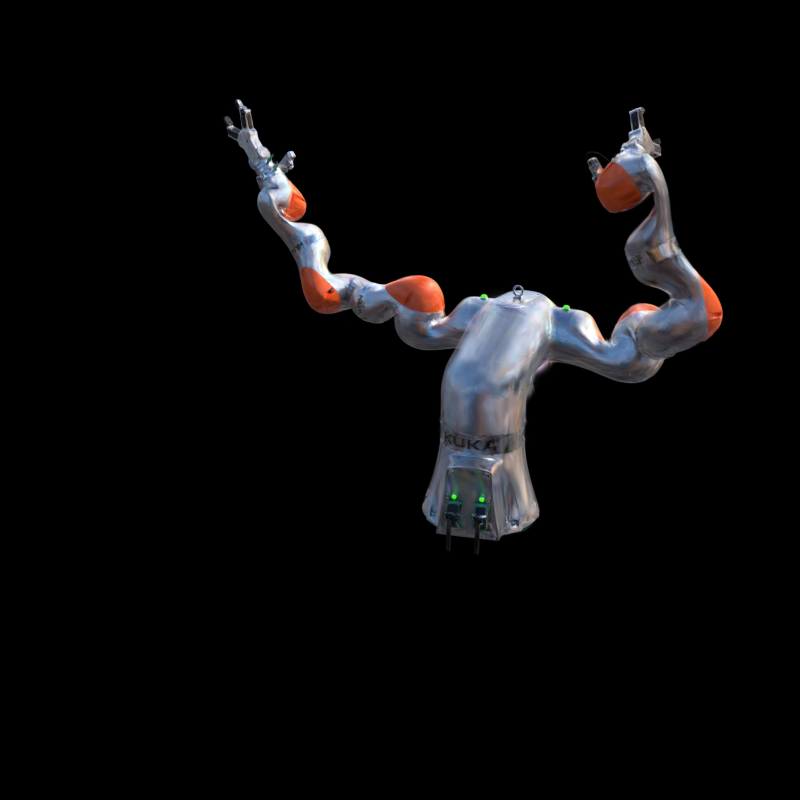} &
            \includegraphics[width=0.145\textwidth]{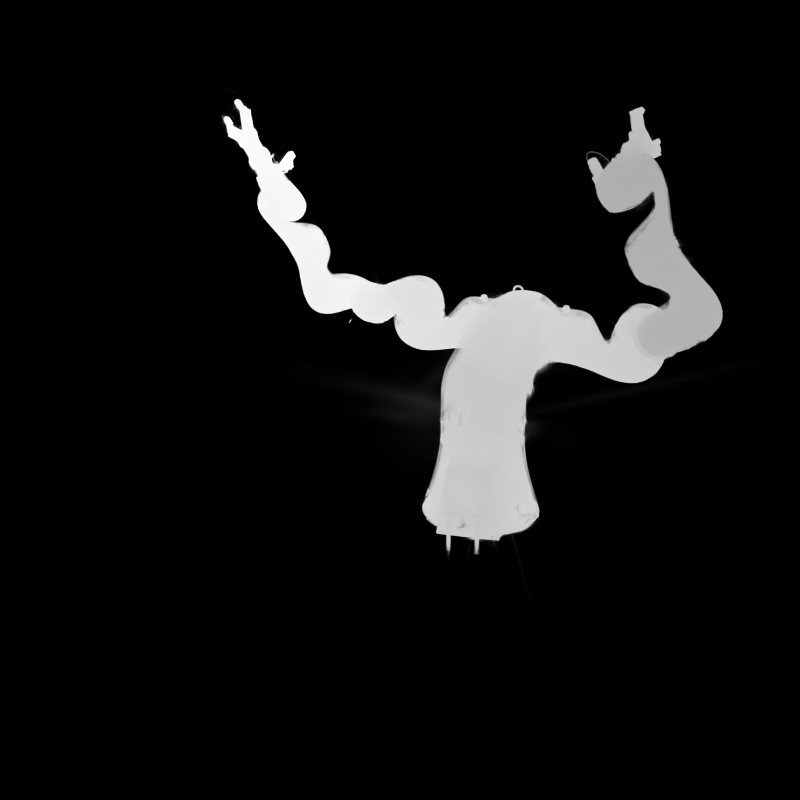} \\

        \raisebox{35pt}{\rotatebox[origin=c]{90}{robot}}&
             \includegraphics[width=0.145\textwidth]{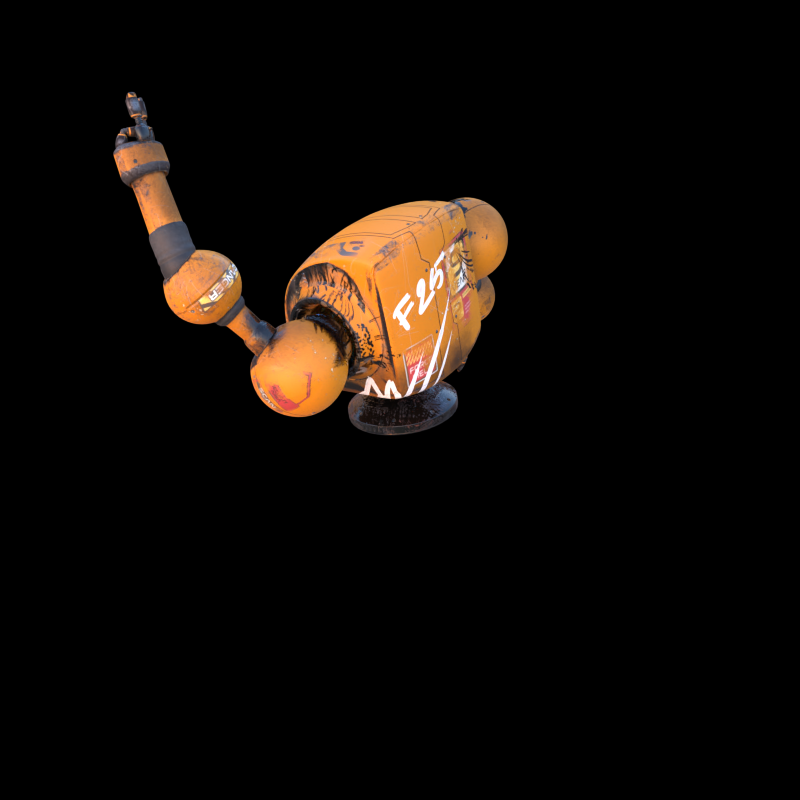} &
            \includegraphics[width=0.145\textwidth]{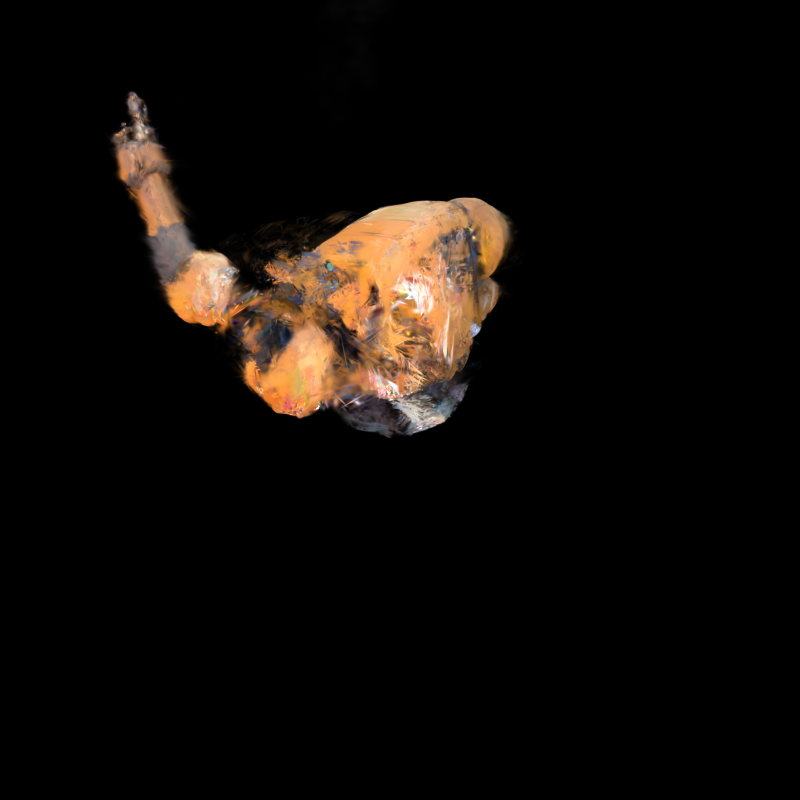} &
            \includegraphics[width=0.145\textwidth]{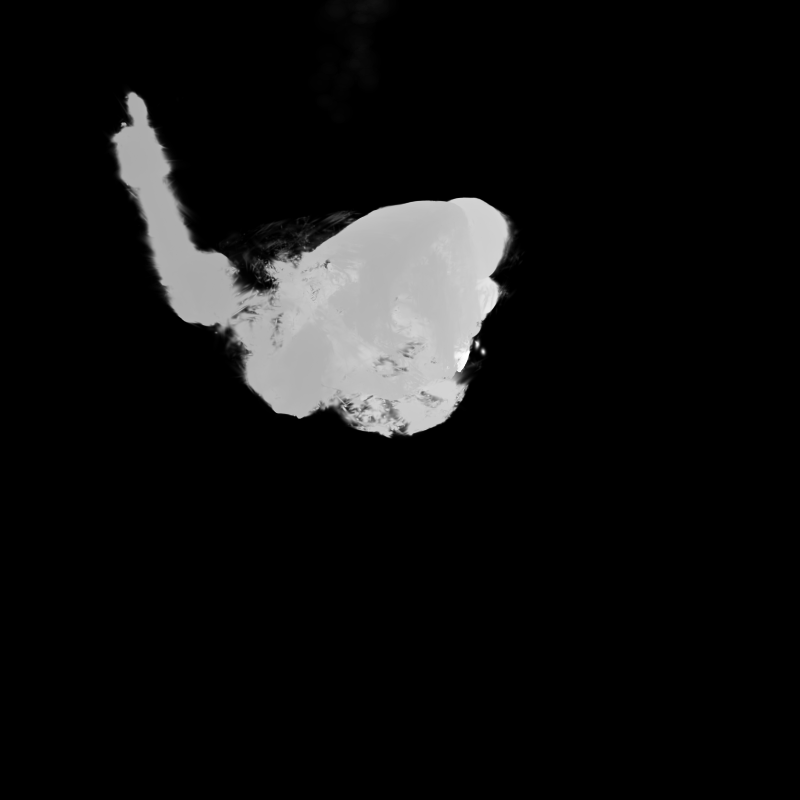} &
            \includegraphics[width=0.145\textwidth]{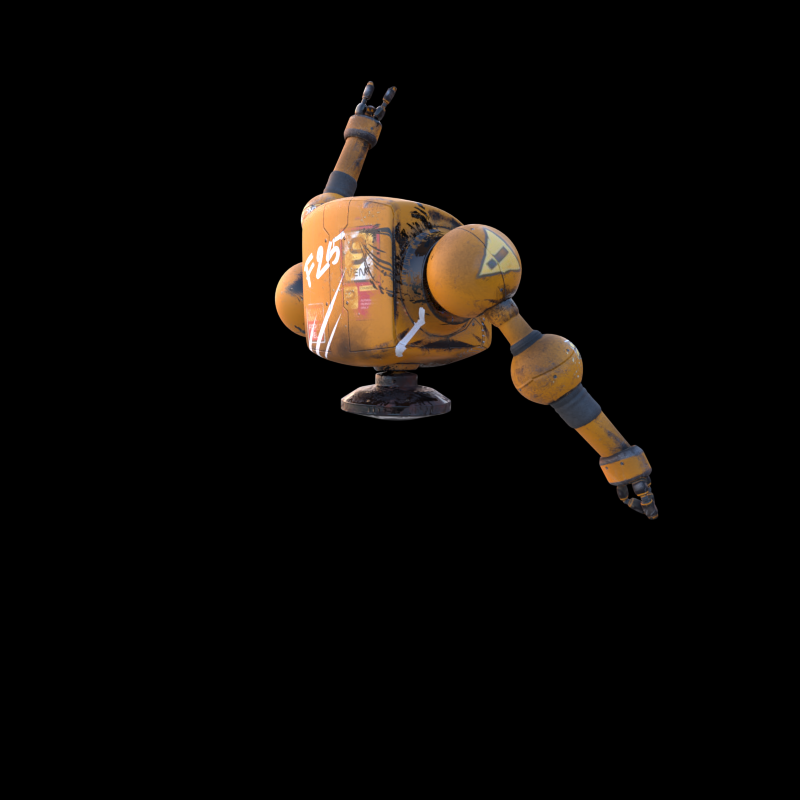} &
            \includegraphics[width=0.145\textwidth]{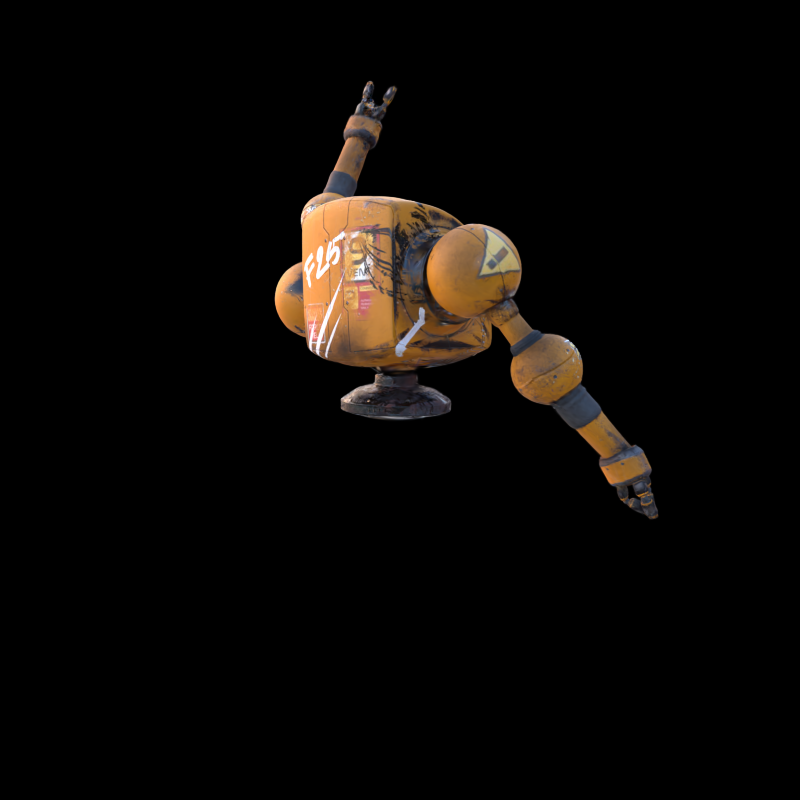} &
            \includegraphics[width=0.145\textwidth]{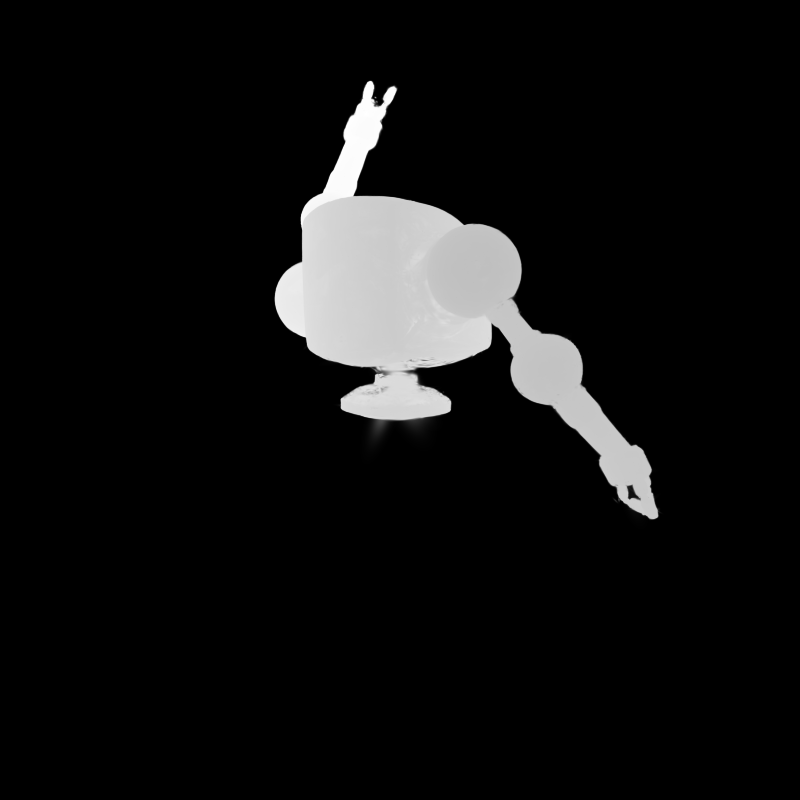} \\

        \raisebox{35pt}{\rotatebox[origin=c]{90}{glove}}&
             \includegraphics[width=0.145\textwidth]{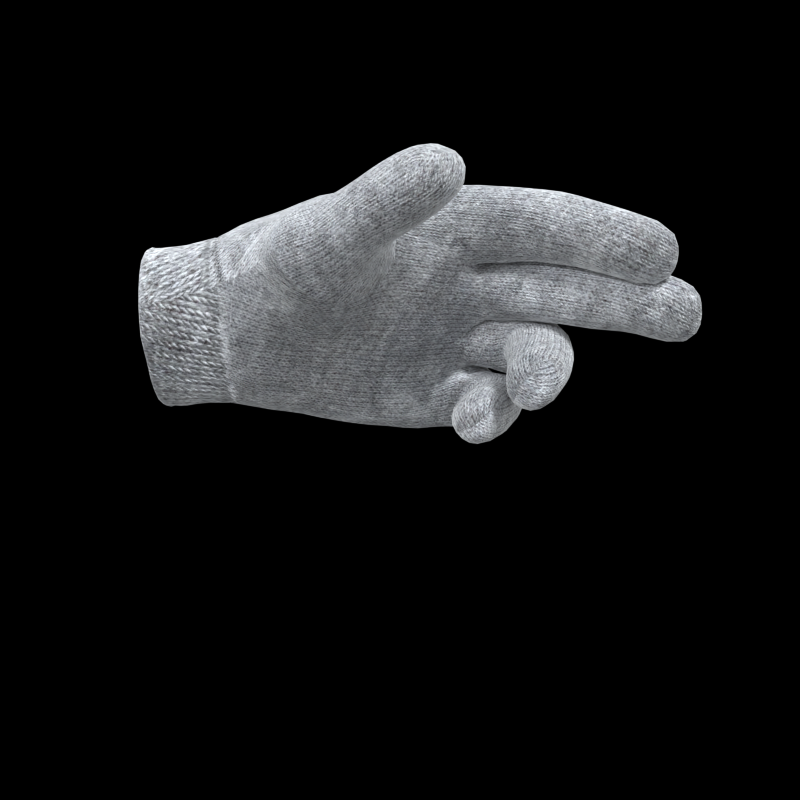} &
            \includegraphics[width=0.145\textwidth]{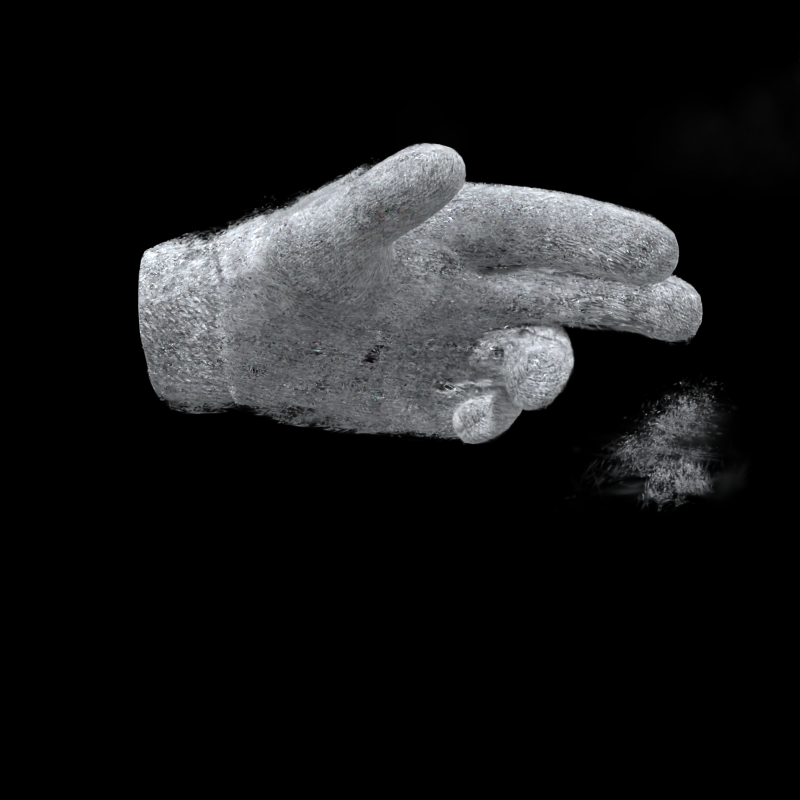} &
            \includegraphics[width=0.145\textwidth]{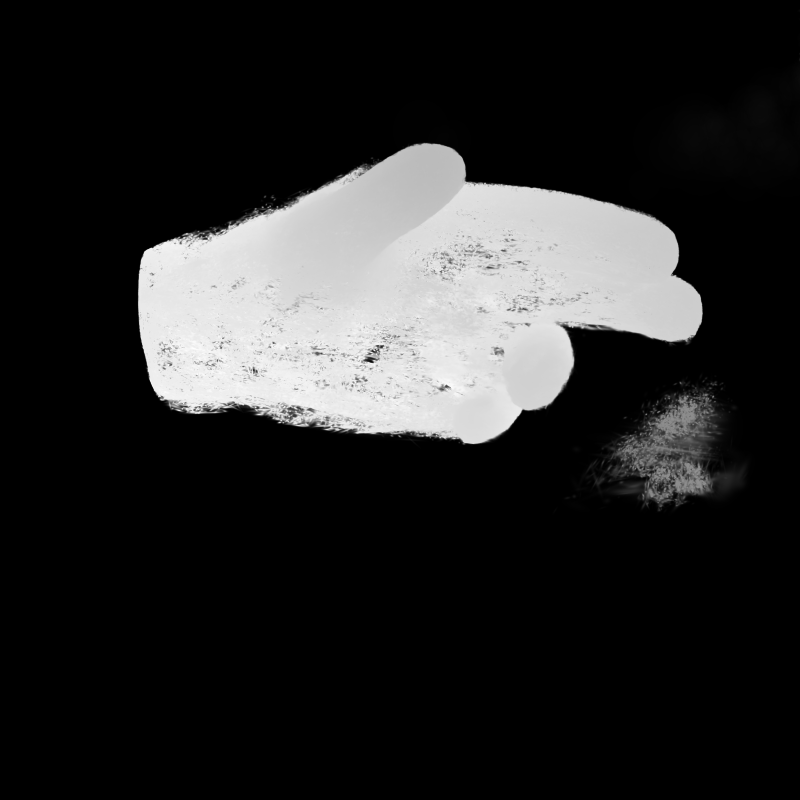} &
            \includegraphics[width=0.145\textwidth]{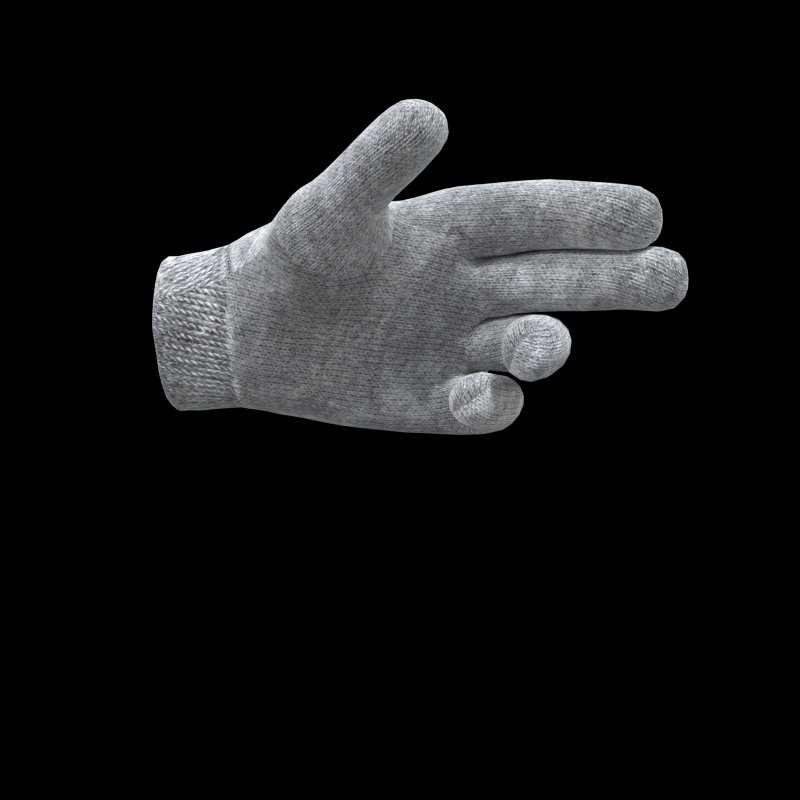} &
            \includegraphics[width=0.145\textwidth]{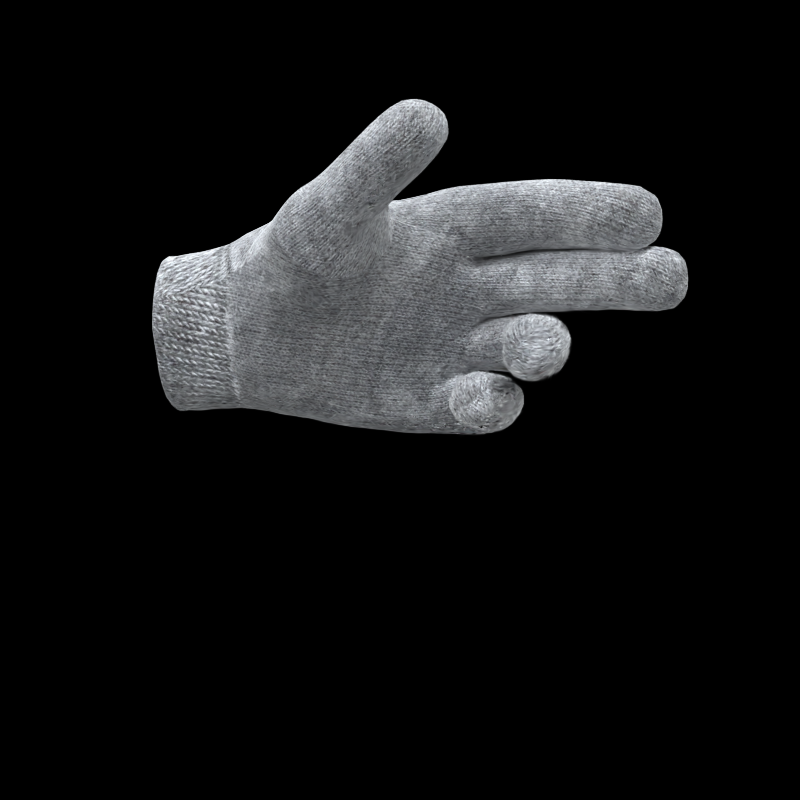} &
            \includegraphics[width=0.145\textwidth]{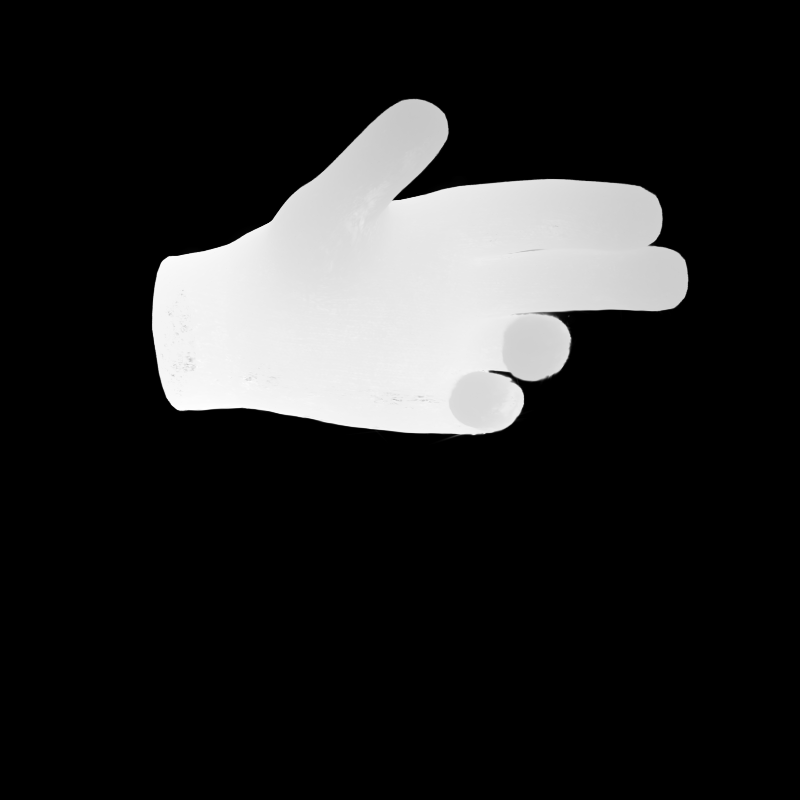} \\

        & gt & render & depth & gt-swap & render-swap & depth-swap
        
        \end{tabular}
    }
\vspace{-5pt}
	\caption{\textbf{Failure case on few training viewpoints.} The first three columns represent the original dataset configurations. The term \textbf{swap} indicates the exchange of training and test sets, thereby ensuring that the model's inputs possess a sufficiently diverse array of viewpoints}\label{fig:failure-viewpoints}
\end{figure*}

\begin{table*}[]
\centering
\begin{tabular}{lccccccccc}
\hline
& lego           & jump           & boungcing      & trex           & mutant         & warrior        & standup        & hook           & mean           \\
\hline
ours       & \textbf{33.07} & \textbf{37.72} & 41.01          & 38.10          & 42.63          & 41.54          & 44.62          & 37.42          & 39.51          \\
ours-SE(3) & 32.91          & 37.60          & \textbf{41.05} & \textbf{38.29} & \textbf{42.83} & \textbf{41.73} & \textbf{44.68} & \textbf{37.60} & \textbf{39.58} \\
\hline
\end{tabular}
\caption{\textbf{Comparison with SE(3) deformation field on the D-NeRF dataset.}}
\label{tab:se3-dnerf}
\end{table*}

\begin{table*}[]
\centering
\begin{tabular}{lcccccccc}
\hline
& as           & basin           & bell      & cup           & plate         & press        & sieve          & mean           \\
\hline
ours       & 26.31 & \textbf{19.67} & \textbf{25.74}          & \textbf{24.86}          & \textbf{20.48}          & \textbf{26.01}          & \textbf{25.70}          & \textbf{24.11}          \\
ours-SE(3) & \textbf{26.37}          & 19.64          & 25.43 & 24.83 & 20.28 & 25.63 & 25.46 &  23.95 \\
\hline
\end{tabular}
\caption{\textbf{Comparison with SE(3) deformation field on the NeRF-DS dataset.}}
\label{tab:se3-nerfds}
\end{table*}

\begin {table*}[ht]
\centering
\begin{tabular}{cccccccccccccc}
\hline \multirow[b]{2}{*}{ } & \multicolumn{3}{c}{ Hell Warrior } & \multicolumn{3}{c}{ Mutant } & \multicolumn{3}{c}{ Hook }\\
& PSNR$\uparrow$ & SSIM$\uparrow$ & LPIPS$\downarrow$ & PSNR$\uparrow$ & SSIM$\uparrow$ & LPIPS$\downarrow$ & PSNR$\uparrow$ & SSIM$\uparrow$ & LPIPS$\downarrow$ \\

\hline 
w/o $\delta s$ & \cellcolor{yzybest}41.55 & \cellcolor{yzybest}0.9878 & \cellcolor{yzybest}0.0223 & 42.15 & 0.9949 & \cellcolor{yzysecond}0.0053 & \cellcolor{yzysecond}37.01 & \cellcolor{yzysecond}0.9859 & \cellcolor{yzysecond}0.0153 \\

w/o $\delta r$ & 41.17 & 0.9866 & 0.0256 & \cellcolor{yzysecond}42.51 & \cellcolor{yzysecond}0.9950 & 0.0054 & 36.82 & 0.9852 & 0.0167 \\

w r\&s & 40.39 & 0.9833 & 0.0323 & 41.30 & 0.9934 & 0.0075 & 36.15 & 0.9818 & 0.0214 \\

ours & \cellcolor{yzysecond}41.54 & \cellcolor{yzysecond}0.9873 & \cellcolor{yzysecond}0.0234 & \cellcolor{yzybest}42.63 & \cellcolor{yzybest}0.9951 & \cellcolor{yzybest}0.0052 & \cellcolor{yzybest}37.42 & \cellcolor{yzybest}0.9867 & \cellcolor{yzybest}0.0144 \\

\hline & \multicolumn{3}{c}{ Bouncing Balls } & \multicolumn{3}{c}{ Lego } & \multicolumn{3}{c}{ T-Rex } \\

 & PSNR$\uparrow$ & SSIM$\uparrow$ & LPIPS$\downarrow$ & PSNR$\uparrow$ & SSIM$\uparrow$ & LPIPS$\downarrow$ & PSNR$\uparrow$ & SSIM$\uparrow$ & LPIPS$\downarrow$ \\

\hline 
w/o $\delta s$  & 40.82 & 0.9952 & 0.0095 & 31.30 & 0.9705 & 0.0260 & 37.39 & 0.9928 & 0.0105 \\

w/o $\delta r$ & \cellcolor{yzybest}41.11 & \cellcolor{yzybest}0.9953 & \cellcolor{yzybest}0.0092 & 32.87 & 0.9783 & 0.0192 & \cellcolor{yzysecond}37.99 & \cellcolor{yzysecond}0.9931 & \cellcolor{yzysecond}0.0101 \\

w r\&s  & 39.89 & 0.9945 & 0.0117 & \cellcolor{yzybest}33.71 & \cellcolor{yzybest}0.9798 & \cellcolor{yzybest}0.0181 & 37.06 & 0.9923 & 0.0113 \\

ours & \cellcolor{yzysecond}41.01 & \cellcolor{yzybest}0.9953 & \cellcolor{yzysecond}0.0093 & \cellcolor{yzysecond}33.07 & \cellcolor{yzysecond}0.9794 & \cellcolor{yzysecond}0.0183 & \cellcolor{yzybest}38.10 & \cellcolor{yzybest}0.9933 & \cellcolor{yzybest}0.0098 \\

\hline & \multicolumn{3}{c}{ Stand Up } & \multicolumn{3}{c}{ Jumping Jacks } & \multicolumn{3}{c}{ Mean } \\

& PSNR$\uparrow$ & SSIM$\uparrow$ & LPIPS$\downarrow$ & PSNR$\uparrow$ & SSIM$\uparrow$ & LPIPS$\downarrow$ & PSNR$\uparrow$ & SSIM$\uparrow$ & LPIPS$\downarrow$ \\

\hline
w/o $\delta s$ & 44.05 & \cellcolor{yzysecond}0.9946 & \cellcolor{yzysecond}0.0074 & \cellcolor{yzysecond}37.49 & \cellcolor{yzysecond}0.9895 & \cellcolor{yzysecond}0.0129 & 38.97 & 0.9889 & 0.0137 \\

w/o $\delta r$ & \cellcolor{yzysecond}44.18 & \cellcolor{yzysecond}0.9946 & 0.0075 & 37.48 & 0.9893 & 0.0138 & \cellcolor{yzysecond}39.27 & \cellcolor{yzysecond}0.9897 & \cellcolor{yzysecond}0.0134 \\

w r\&s  & 42.88 & 0.9932 & 0.0097 & 37.00 & 0.9878 & 0.0164 & 38.55 & 0.9883 & 0.0160 \\

ours & \cellcolor{yzybest}44.62 & \cellcolor{yzybest}0.9951 & \cellcolor{yzybest}0.0063 & \cellcolor{yzybest}37.72 & \cellcolor{yzybest}0.9897 & \cellcolor{yzybest}0.0126 & \cellcolor{yzybest}39.51 & \cellcolor{yzybest}0.9902 & \cellcolor{yzybest}0.0124 \\

\hline
\end{tabular}
\vspace{-5pt}
\caption{\textbf{Ablations on network architecture.} We color each cell as \colorbox{yzybest}{best} and \colorbox{yzysecond}{second best}. $\delta r$ and $\delta s$ denote the output components of the MLP model. The term \textbf{w r\&s} signifies that the model inputs include not only time and the position of the 3D Gaussians but also the 3D Gaussians' rotation and scaling. The experimental outcomes affirm that our network architecture is the most advantageous.}
\label{tab: ablation-arch}
\end{table*}

\begin{table*}[]
\centering
\begin{tabular}{lccccccccc}
\hline
& lego           & jump           & bouncing      & trex           & mutant         & warrior        & standup        & hook           & mean           \\
\hline
ours       & \textbf{33.07} & \textbf{37.72} & 41.01          & 38.10          & \textbf{42.63}          & \textbf{41.54}          & \textbf{44.62}          & \textbf{37.42 }         & \textbf{39.51}          \\
ours-white & 32.03          & 36.87          & \textbf{43.52} & \textbf{38.57} & 42.11 & 32.75 & 42.40 & 36.60 & 38.10 \\ 
ours-best & 33.07          & 37.72         & 43.52 & 38.57 & 42.63 & 41.54 & 44.62 & 37.42 & 39.89 \\
\hline
\end{tabular}
\caption{\textbf{Comparison with different background colors on the D-NeRF dataset.}We explored the impact of different background colors on rendering metrics using the D-NeRF dataset. The experimental results showed that overall, a black background yielded higher metrics, while bouncing and trex scenes performed better with a white background, and the warrior scene had higher metrics with a black background. To ensure \textbf{experimental consistency}, we uniformly used a black background in the main text. If one wishes to pursue the best metrics for a specific scene, one can refer to this table to adjust the background color.}
\label{tab:bg-color-dnerf}
\end{table*}

\par Our deformation field does not employ any grid/plane-based structures which have been demonstrated to be superior in static scenes because these structures are predicated on a \textbf{low-rank tensor assumption} \cite{Chen2022ECCV}. Dynamic scenes possess a higher rank compared to static scenes, and explicit point-based rendering exacerbates the rank of the scene.

\subsection{Optimization Loss}
\par During the training of our deformable Gaussians, we deform the 3D Gaussians at each timestep into the canonical space. We then optimize both the deformation network and the 3D Gaussians using a combination of $\mathcal L_1$ loss and D-SSIM loss \cite{kerbl3Dgaussians}:

\begin{equation}
\mathcal{L} = (1-\lambda) \mathcal{L}_1 + \lambda \mathcal{L}_{\text{D-SSIM}},
\end{equation}
where $\lambda=0.2$ is used in all our experiments.

\section{Additional Results} \label{sec: add-results}

\subsection{Per-Scene Results on the NeRF-DS Dataset}

In Tab. \ref{tab: nerfds-per}, we provide the results for individual scenes associated with Sec. 4 of the main paper. It can be observed that our method achieved superior metrics in almost every scene compared to those without AST, underscoring the generalizability of AST on real datasets where the pose is not perfectly accurate. Overall, our method outperforms baselines on the NeRF-DS Dataset.

\subsection{Results on the HyperNeRF Dataset}

We visualize the results of the HyperNeRF dataset in Fig. \ref{fig:hyper-quality}. Notably, metrics designed to assess image rendering quality, such as PSNR, tend to penalize minor offsets more heavily than blurring. Therefore, for datasets with less accurate camera poses, like HyperNeRF, our method's quantitative metrics might not consistently outperform those of methods yielding blurred outputs when faced with imprecise camera poses. Despite this, our rendered images often exhibit fewer artifacts and greater clarity. This phenomenon aligns with observations reported in Nerfies \cite{park2021nerfies} and HyperNeRF \cite{park2021hypernerf}.

\subsection{Results on Rendering Efficiency}
\par In our research, we present comprehensive Frames Per Second (FPS) testing results in Tab. \ref{tab:fps}. Tests were conducted on one NVIDIA RTX 3090. It is observed that when the number of point clouds remains below $\sim$250k, our method can achieve real-time rendering at rates greater than 30 FPS. A point of note is that the point cloud count reconstructed from the HyperNeRF dataset significantly exceeds that of other datasets, reaching a level of 1,000k. This excessive count is attributed to the highly inaccurate camera poses within the HyperNeRF dataset. In contrast, the NeRF-DS dataset, while also being derived from the real world, exhibits more accurate poses, resulting in a reconstructed point cloud count within a reasonable range. This issue of an overabundant point cloud count occurs not only in scenes with inaccurate poses but also in those with sparse viewpoints, as evidenced in scenes like the D-NeRF's Lego scene, which was trained on merely 50 images.

\subsection{More Ablations}
\paragraph{Network architecture.}

\par We present ablation experiments on the architecture of our purely implicit network, as shown in Tab. \ref{tab: ablation-arch}. The results of these experiments suggest that the structure within our pipeline is optimal. Notably, we did not adopt Grid/Plane-based structures because dynamic scenes do not conform to the \textbf{low-rank assumption}. Furthermore, the explicit point-based rendering of 3D-GS exacerbates the rank of dynamic scenes. Our early experimental validations have corroborated this assertion.

\subsection{Background color}

\par In the research of Neural Rendering, it's common to use a black or white background for rendering scenes without a background. In our experiments, we found that the background color has an impact on certain scenes in the D-NeRF dataset. The experimental results are shown in Tab. \ref{tab:bg-color-dnerf}. Overall, a black background yields better rendering results. For the sake of consistency in our experiments, we uniformly used a black background in our main text experiments. However, for the bouncing and trex scenes, using a white background can produce better results.

\subsection{Deformation using SE(3) Field}
\par Drawing inspiration from Nerfies \cite{park2021nerfies}, we applied a 6-DOF SE(3) field that accounts for rotation to the transformation of 3D Gaussian positions. The experimental results, presented in Tab. \ref{tab:se3-dnerf} and Tab. \ref{tab:se3-nerfds}, indicate that this constraint offers a minor improvement on the D-NeRF dataset. However, it appears to diminish the quality on the more complex real-world NeRF-DS dataset. Moreover, the additional computational overhead introduced by the SE(3) Field approximately increases 50 \% of the training time and results in about a 20\% decrease in FPS during rendering. Consequently, we opted to utilize a direct addition without imposing SE(3) constraints on the transformation of position.

\section{Failure Cases} \label{sec:failure}
\paragraph{Inaccurate poses. }In our research, we find that inaccurate poses can lead to the failure of the convergence of deformable-gs, as illustrated in Fig. \ref{fig:failure-pose}. For implicit representations, their inherent smoothness can maintain robustness in the face of minor deviations in pose. However, for the explicit point-based rendering, such inaccuracies are particularly detrimental, resulting in inconsistencies in the scene at different moments.

\paragraph{Few training viewpoints. } In our study, a notable scarcity of training views presents a dual challenge: both few-shot learning and a limited number of viewpoints. Either aspect can lead to overfitting in deformable-gs and even in 3D-GS on the training set. As demonstrated in Fig. \ref{fig:failure-viewpoints}, significant overfitting is evident in the DeVRF \cite{liu2022devrf} dataset. The training set for this scene contains 100 images, but the viewpoints for training are limited to only four. However, by swapping the training and test sets, where the test set contained an equal number of 100 images and viewpoints, we obtained markedly better results.

\end{document}